\documentclass{article}
\usepackage{natbib}
\usepackage[colorlinks=true, citecolor=blue]{hyperref}
\usepackage[preprint]{neurips_2025}
\usepackage[utf8]{inputenc} % allow utf-8 input
\usepackage[T1]{fontenc}    % use 8-bit T1 fonts
\usepackage{hyperref}       % hyperlinks
\usepackage{url}            % simple URL typesetting
\usepackage{booktabs}       % professional-quality tables
\usepackage{amsfonts}       % blackboard math symbols
\usepackage{nicefrac}       % compact symbols for 1/2, etc.
\usepackage{microtype}      % microtypography
\usepackage[table]{xcolor}         % colors
\usepackage{graphicx}
\usepackage{amsmath} 
\usepackage{amsthm}
\usepackage{soul}
\usepackage{array}
\usepackage{multirow}
\usepackage{booktabs}

\newtheorem{theorem}{Theorem}[section]
\newtheorem{proposition}[theorem]{Proposition}

\newtheorem{corollary}[theorem]{Corollary}

\theoremstyle{definition}
\newtheorem{definition}[theorem]{Definition}

\theoremstyle{remark}
\newtheorem{remark}[theorem]{Remark}

\newcommand{\revblue}[1]{#1}  % For final camera-ready (uncomment this line and comment above)
\title{FlowMixer: A \revblue{Depth-Agnostic} Neural Architecture for Interpretable Spatiotemporal Forecasting}
\definecolor{wordblue}{RGB}{8,88,188}
\definecolor{wordorange}{RGB}{188,88,8} 
\newcommand{\bv}[1]{\textcolor{wordblue}{\textbf{#1}}}
\newcommand{\bsv}[1]{\textcolor{wordorange}{\text{#1}}}
\usepackage{hyperref}
\author{
  Fares B. Mehouachi\\
  New York University in Abu Dhabi\\
  Abu Dhabi, UAE\\
  \texttt{fm2620@nyu.edu}\\
  \And
  Saif Eddin Jabari\\
  New York University\\
  Abu Dhabi, UAE \& Brooklyn, USA\\
  \texttt{sej7@nyu.edu}
}

\begin{document}

\makeatletter
\renewcommand{\@notice}{\footnotetext{\hspace{-1.8em}Accepted (main track) at the 39th Conference on Neural Information Processing Systems (NeurIPS 2025).}}
\makeatother

\maketitle

\begin{abstract}
We introduce FlowMixer, a \revblue{single-layer} neural architecture that leverages constrained matrix operations to model structured spatiotemporal patterns \revblue{with enhanced interpretability}. FlowMixer incorporates non-negative matrix mixing layers within a reversible mapping framework\textemdash applying transforms before mixing and their inverses afterward. This shape-preserving design enables a Kronecker-Koopman eigenmodes framework that bridges statistical learning with dynamical systems theory, providing interpretable spatiotemporal patterns and facilitating direct algebraic manipulation of prediction horizons without retraining. \revblue{The architecture's semi-group property enables this single layer to mathematically represent any depth through composition, eliminating depth search entirely.} Extensive experiments across diverse domains demonstrate FlowMixer's long-horizon forecasting capabilities while effectively modeling physical phenomena such as chaotic attractors and turbulent flows. \revblue{Our results achieve performance matching state-of-the-art methods while offering superior interpretability through directly extractable eigenmodes.} This work suggests that architectural constraints can simultaneously \revblue{maintain competitive performance and enhance} mathematical interpretability in neural forecasting systems.
\end{abstract}

\section{Introduction}
\label{sec:intro}

The ability to predict complex spatiotemporal patterns remains a fundamental challenge across scientific disciplines, including climate modeling, biological systems, and fluid dynamics \cite{pathak2018model,vinuesa2022enhancing, wang2024interpretable}. Statistical methods have traditionally dominated time series forecasting \cite{box2015time}, while mechanistic models form the backbone of dynamical systems analysis \cite{brunton2019data}. Recent advances in Machine Learning have demonstrated that these approaches can be complemented and extended through data-driven methods \cite{brunton2024promising,gauthier2021next}. For example, reservoir computing, originally designed for chaos prediction \cite{pathak2018model,gauthier2021next,li2024higher}, has shown promise in statistical forecasting \cite{du2017reservoir,bianchi2020reservoir}, while neural architectures for time series forecasting have successfully captured chaotic dynamics \cite{gilpin2021chaos}. This convergence hints at the possibility of a unified framework for spatiotemporal pattern learning \cite{vlachas2020backpropagation,yan2024emerging}.

Achieving this integration presents several challenges. Time series data often exhibit non-stationarity, where statistical properties evolve over time, creating distribution shifts between training and testing \cite{sugiyama2012machine}. Dynamical systems, although low-dimensional, may exhibit chaotic behavior characterized by extreme sensitivity to initial conditions \cite{lorenz1963deterministic,rossler1976equation,aizawa1994chaos,strogatz2018nonlinear}. Additionally, the differing operational requirements of these tasks\textemdash typically multi-step predictions for time series and iterative one-step predictions for dynamical systems\textemdash demand flexible and efficient temporal modeling capabilities.

Recent neural architectures have made significant progress in addressing these challenges. Transformer-based models like PatchTST \cite{patchtst2023} effectively capture long-range dependencies through attention mechanisms and patch tokenization. MLP-based architectures such as TimeMixer++ \cite{timemixerplusplus2024} and WPMixer \cite{wpmixer2025} achieve impressive efficiency through multi-scale decomposition and structured mixing operations. State space models, including Chimera \cite{chimera2024}, offer promising capabilities for modeling multivariate time series with coupled dynamics. Meanwhile, Gilpin \cite{gilpin2023model} has shown that large-scale statistical models can effectively capture the behavior of chaotic systems, while foundation models \cite{time_llm2024,gilpin2024zeroshot} demonstrate emerging capabilities for few-shot and zero-shot forecasting. \revblue{These architectures typically require extensive hyperparameter search to determine optimal network depth, which varies significantly across datasets and domains.}
\vspace{-0.25cm}
\begin{figure*}[!ht]
\centering
\includegraphics[width=0.96\textwidth]{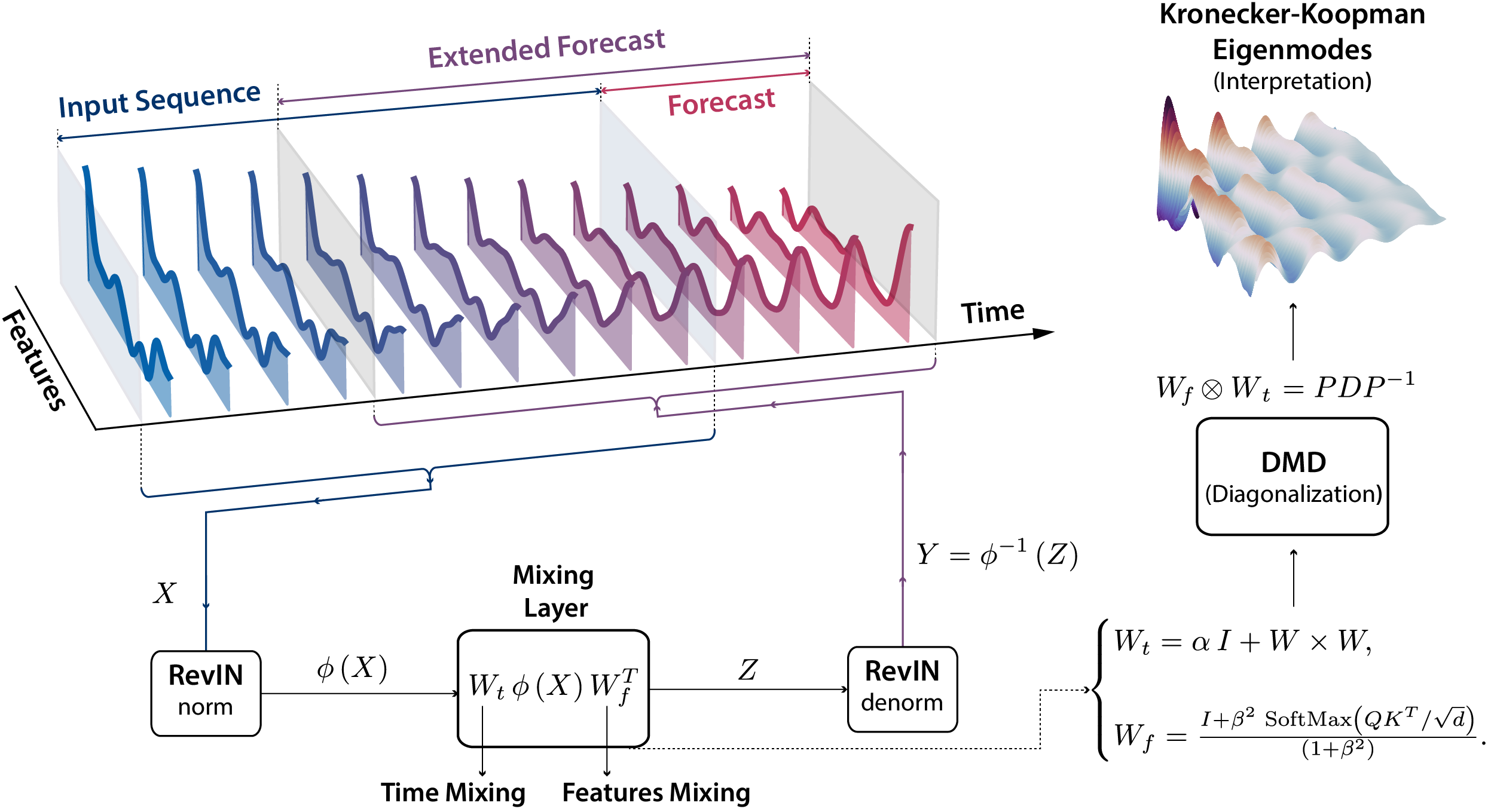}
\caption{Overview of FlowMixer's \revblue{single-layer} constrained architecture. The core architecture (bottom) processes input sequences using three key components: (1) a reversible mapping, mainly reversible instance normalization (RevIN) to handle distribution shifts, (2) a constrained mixing layer with \revblue{non-negative} time mixing ($W_t$) and stochastic feature mixing ($W_f$) matrices, and (3) adaptive skip connections embedded within the mixing matrices. The spatiotemporal evolution (top) demonstrates how data is padded to create square mixing matrices, enabling eigendecomposition and efficient temporal modeling and interpretation. The Kronecker-Koopman eigenmodes (right), derived from the architecture's mathematical structure ($W_f \otimes W_t = PDP^{-1}$), provide a space-time decomposition of the mixer input as a weighted sum of space-time eigenmodes.}
\label{fig:flowmixer}
\vspace{-0.25cm}
\end{figure*}

Here, we introduce FlowMixer (Fig.~\ref{fig:flowmixer}), a neural architecture designed to address these challenges through constrained design. Inspired by non-negative matrix factorization \cite{lee1999learning} and Dynamic Mode Decomposition (DMD) \cite{schmid2010dynamic}, FlowMixer aims to draw connections between statistical learning approaches and dynamical systems theory while providing a mathematically interpretable framework. 
\revblue{FlowMixer explores a different point in the design space: rather than optimizing architectures for individual domains, we demonstrate that careful constraints enable a single architecture to achieve competitive performance across time series forecasting, with additional capabilities in chaos prediction and turbulent flow modeling. This cross-domain versatility, combined with interpretable eigenmodes and elimination of depth search, offers practical advantages for applications requiring consistent behavior across varied spatiotemporal phenomena.}

Our contributions include:

\begin{itemize}
\item \revblue{A constrained architecture whose semi-group property eliminates depth as a hyperparameter: a single layer is all you need.}
\item \revblue{A shape-preserving design that enables direct extraction of Kronecker-Koopman eigenmodes for interpretability and allows algebraic prediction horizon change without retraining.}
\item \revblue{A simplified chaos prediction approach through Semi-Orthogonal Basic Reservoir (SOBR) that matches specialized frameworks while maintaining architectural simplicity.}
\item \revblue{Empirical validation across time series forecasting, chaotic systems, and turbulent flows demonstrating competitive cross-domain performance.}
\end{itemize}
%These innovations suggest that carefully designed architectural constraints can improve both predictive capabilities and mathematical interpretability, potentially offering new directions for statistical and physical modeling approaches.

\section{Related Work}
\label{sec:related}

\subsection{Time Series Forecasting Architectures}
\revblue{
Recent advances in time-series forecasting have focused on improving accuracy through architectural sophistication. TSMixer \cite{chen2023tsmixer} employs all-MLP time and feature mixing but requires empirical determination of optimal layer count. TimeMixer++ \cite{timemixerplusplus2024} employs hierarchical multi-scale decomposition with bidirectional mixing for seasonal and trend patterns. Chimera \cite{chimera2024} extends state space models to multivariate time series through a 2D SSM architecture with three-headed processing along time and variate dimensions. While these methods achieve strong performance, they compound the deployment challenge: each new dataset potentially requires extensive architecture search from scratch.}

\subsection{Koopman-Based Neural Methods}
\revblue{
The integration of Koopman operator theory with neural networks has produced sophisticated forecasting frameworks~\cite{liu2023koopa,yu2023koopman,yang2025urban}. Koopa \cite{liu2023koopa} employs multi-layer architectures with eigenvalue-based stability checking and explosion prevention mechanisms, while KNF \cite{yu2023koopman} utilizes a three-term loss function across various network components. Both approaches treat the Koopman semi-group property $\mathcal{K}_t \circ \mathcal{K}_s = \mathcal{K}_{t+s}$ as an optimization target. Notably, their eigenmodes remain deeply embedded within neural network layers\textemdash inaccessible for direct analysis.}

\subsection{Cross-Domain Spatiotemporal Methods}
\revblue{
Few architectures attempt unified modeling across statistical and physical domains. Reservoir computing \cite{pathak2018model,gauthier2021next} bridges chaos prediction and time series but requires careful tuning of reservoir size, spectral radius, and input scaling. N-BEATS \cite{oreshkin2020n}, though designed for time series, demonstrates exceptional performance across chaotic systems \cite{gilpin2021chaos}, suggesting untapped cross-domain potential.}

\revblue{FlowMixer sits at the intersection of these research directions: it adopts the mixing paradigm from time series architectures, guarantees Koopman's mathematical properties by construction, and achieves cross-domain applicability through a single constrained design.}

\section{FlowMixer Architecture}
\label{sec:architecture}

FlowMixer builds upon TSMixer \cite{chen2023tsmixer} by addressing the challenge of layer count optimization. Instead of empirical layer stacking, we developed a single mixing block with semi-group properties where successive applications combine algebraically (see Section~\ref{sec:semi_group}). This approach eliminates \revblue{depth as} a hyperparameter while enabling mathematical manipulation of prediction horizons (see Section~\ref{sec:extrap}). The various components of this architecture are presented hereafter.

\subsection{Formal Definition}
Let $X \in \mathbb{R}^{n_t \times n_f}$ be an input tensor, where $n_t$ denotes the number of time steps and $n_f$ the number of features. FlowMixer defines a transformation $\mathcal{F}: \mathbb{R}^{n_t \times n_f} \rightarrow \mathbb{R}^{n_t \times n_f}$ as:
\begin{equation}
\mathcal{F}(X,W_t,W_f,\phi) = \phi^{-1}(W_t\phi(X)W_f^T)
\end{equation}
where $\phi: \mathbb{R}^{n_t \times n_f} \rightarrow \mathbb{R}^{n_t \times n_f}$ is a reversible mapping, $W_t \in \mathbb{R}^{n_t \times n_t}$ is a time mixing matrix, and $W_f \in \mathbb{R}^{n_f \times n_f}$ is a feature mixing matrix. \revblue{FlowMixer preserves tensor dimensions: the output $\mathcal{F}(X) \in \mathbb{R}^{n_t \times n_f}$ approximates a target $Y$ of the same shape, where the last $h$ rows of $Y$ contain the forecast and earlier rows contain historical data. Padding with historical data, ensures square mixing matrices, enabling eigendecomposition and semi-group composition.} The name "FlowMixer" reflects this reversible flow of data around the central mixing operations.

\subsection{Core Components}
\label{sec:components}
\paragraph{Time-Dependent Reversible Instance Normalization:} We employ mainly RevIN \cite{kim2022reversible} as $\phi$ for feature-wise normalization, extending it to Time-Dependent RevIN (TD-RevIN):
\begin{equation}
\text{RevIN}(x_{t})=a\frac{x_{t}-\mathbb{E}[x_{t}]}{\sqrt{\text{Var}(x_{t})+\epsilon}}+b\to\text{TD-RevIN}(x_{t})=a_{t}\times\frac{x_{t}-\mathbb{E}[x_{t}]}{\sqrt{\text{Var}(x_{t})+\epsilon}}+b_{t}
\end{equation}
where $a_t,b_t$ are time-varying parameters, and $\times$ denotes the Hadamard product. This allows different normalization parameters for each timestep, enhancing the model's ability to handle non-stationary patterns. This TD extension is possible thanks to the shape-preserving property of FlowMixer.

\paragraph{Constrained Mixing:} The core functionality integrates principles from non-negative matrix factorization and attention mechanisms~\cite{lee1999learning,vaswani2017attention}. For an input tensor $X$, our architecture applies two complementary transformations:
\begin{equation} \text{TimeMix}(X) = W_tX, \quad \text{FeatureMix}(X) = XW_f^T \end{equation}

The time mixing matrix $W_t$ aligns with Koopman theory through a scaled matrix exponential:
\begin{equation} W_t = \alpha e^{(W_0 \times W_0)} \approx \alpha I + \alpha W_0 \times W_0 \end{equation} 
\revblue{where $\times$ denotes Hadamard (element-wise) product. $\alpha$ is a scaling learnable parameter initialized at one that controls memory persistence.  In practice, we use this first-order approximation with a learnable matrix $W$ that absorbs the scaling, yielding $W_t = \alpha I + W \times W$.} This design creates an analog to fractional differencing (FARIMA)~\cite{granger1980introduction} that adapts to varying temporal dependencies (see Appendix~\ref{app:FARIMA}). The nonnegative quadratic term forces the model to capture temporal dependencies through positive interactions only, reducing spurious correlations while introducing nonlinear interactions.

For feature mixing, we implement a static attention mechanism:
\begin{equation}
W_f = \frac{I + \beta^2\text{SoftMax}(QK^T/\sqrt{d_k})}{1 + \beta^2}
\end{equation}
where $K, Q \in \mathbb{R}^{n_f \times d_k}$ are learnable key/query matrices, $d_k$ is the multiplication dimension of $K, Q$, and $\beta$ modulates attention strength. This formulation has a graph-theoretical interpretation as transition probabilities in a weighted directed graph, analogous to a Markov process.

\paragraph{Seasonalities:} For time series with inherent periodicities, we enhance the time mixing operation by leveraging a Kronecker structure:
\begin{equation}
W_t = \sum_p W_{r(p)} \otimes W_p, \quad p \times r\left(p\right)=n_t.
\end{equation}
where $W_p \in \mathbb{R}^{p \times p}$ captures patterns at the period level $p$, $W_{r(p)} \in \mathbb{R}^{r \times r}$ models relationships between different period blocks, and $r \times p = n_t$. This formulation enables efficient modeling of multiple seasonal patterns while preserving the simplicity of the architecture.

\paragraph{SOBR: Semi-Orthogonal Basic Reservoir.} For chaotic systems, we introduce SOBR\textemdash a simplified reservoir computing approach that uses random, non-trainable, semi-orthogonal matrices. SOBR replaces $\phi$ with $S \circ \phi$ where $S\left(X\right) = \sigma(U_tXU_f^T)$ and $\sigma$ is an invertible activation (e.g., Leaky ReLU). These matrices are constructed to satisfy:
\begin{equation}
U_fU_f^T = I_{d_f}, \quad U_tU_t^T = I_{d_t}
\end{equation}
This expansion provides a higher-dimensional representation space appropriate for modeling chaotic systems (discussed in Section~\ref{sec:theory}).

\section{Theoretical Analysis}
\label{sec:theory}

The architectural constraints of FlowMixer yield unique theoretical properties that bridge concepts from statistical learning with dynamical systems theory \cite{brunton2019data, kutz2016dynamic}. The key insights emerge from the architecture's semi-group stability.

\subsection{Semi-Group Structure}
\label{sec:semi_group}
Square mixing matrices and invertible standardization enable semi-group stability\textemdash compositions of FlowMixers with identical standardization $\phi$ remain a FlowMixer:
\begin{equation}
\mathcal{F}(\mathcal{F}(x, \theta, \phi), \theta', \phi) = \mathcal{F}(x, \theta'', \phi)
\end{equation}

\revblue{This property fundamentally distinguishes FlowMixer from all existing forecasting architectures: while traditional architectures require depth optimization and alter capacity with each layer, any $n$-layer FlowMixer composition has an equivalent single-layer representation\footnote{For SOBR models, semi-group property holds in lifted space.}. This mathematical guarantee eliminates architecture search entirely and} enables flexible horizon adjustments without retraining.

\subsection{Eigen Decomposition and Kronecker-Koopman Framework}
In vectorized form, FlowMixer's mixing operation reveals a Kronecker product:
\begin{equation}
\text{vec}(W_tXW_f^T) = (W_f \otimes W_t)\text{vec}(X)
\end{equation}

This enables the extension of classical Dynamic Mode Decomposition (DMD)~\cite{schmid2010dynamic} to coupled spatiotemporal patterns in a novel principled way without Hankelization~\cite{arbabi2017ergodic,brunton2017chaos}. Given eigendecompositions $W_f = PDP^{-1}$ and $W_t = QEQ^{-1}$, where $D = \text{diag}(\lambda_i)$ and $E = \text{diag}(\mu_j)$, any input matrix $X$ can be expressed as:
\begin{equation}
X = \sum_{i,j} a_{i,j}(q_ip_j^T)
\label{eq:kkdecomp}
\end{equation}
where $q_i$, $p_j$ are columns of $Q$ and $P$, and $a_{i,j} = (Q^{-1}XP^{-T})_{i,j}$ are projection coefficients. 

\revblue{Unlike Koopman neural methods where eigenmodes remain embedded~\cite{liu2023koopa,yu2023koopman}, these modes are directly extractable through standard matrix decomposition.} Both time and feature eigenvectors form bases in their respective spaces. These components form the Kronecker-Koopman (KK) spatiotemporal eigenmodes:
\begin{equation}
\Phi_{i,j} = q_i \otimes p_j, \quad (i,j) \in [1,n_t] \times [1,n_f]
\end{equation}

The Kronecker product preserves \revblue{completeness}, yielding $n_t \times n_f$ basis eigenmodes spanning all possible spatiotemporal patterns. These modes are analogous to Koopman modes \cite{koopman1931hamiltonian, brunton2022modern}, but naturally couple spatial and temporal patterns. Section~\ref{sec:KK} provides an example visualization of these eigenmodes.

\subsection{Continuous Algebraic Horizon Modification}
\label{sec:extrap}
The KK structure enables direct algebraic modification of prediction horizons. For a horizon change from $h_0$ to $h_1$ and a ratio $t = h_1/h_0$, we can express the modified prediction as:
\begin{equation}
X\left(t\right)\leftarrow\sum_{i,j}a_{i,j}\underset{\text{KK Eigenmodes}}{\underbrace{\left(q_{i}p_{j}^{T}\right)}}\exp\left(t\log\underset{\text{KK Eigenvalues}}{\underbrace{\left(\lambda_{i}\mu_{j}\right)}}\right).
\label{eq:extrap}
\end{equation}

This enables continuous horizon adjustment through interpolation ($t < 1$) or extrapolation ($t > 1$) (See Appendix~\ref{app:extrap}). The framework extends naturally to derivatives. For $\alpha>0$:
\begin{equation}
\frac{\partial^{\alpha}X}{\partial t^{\alpha}}=\sum_{i,j}a_{i,j}\left(q_{i}p_{j}^{T}\right)\left(\lambda_{i}\mu_{j}\right)^{t}\left(\log\left(\lambda_{i}\mu_{j}\right)\right)^{\alpha}
\end{equation}

The eigenmodes structure remains invariant during these modifications, with temporal evolution controlled mainly through eigenvalue scaling. Stability \revblue{requires} that all eigenvalues are inside the unit circle and can be implemented through unitary constraints on time mixing matrices \cite{baddoo2023physics}. The feature matrix is already stochastic and verifies this property.

%%%%%%%%%%%%%%%%%%%%%%%%%%%%%%%%%%%%%%%%%%%%%%%%%%%%%%%%%%%%%%%%%%%%%%%%%%%%%%%
%%%%%%%%%%%%%%%%%%%%%%%%%%%%%%%%%%%%%%%%%%%%%%%%%%%%%%%%%%%%%%%%%%%%%%%%%%%%%%%
% Experimental Validation
%%%%%%%%%%%%%%%%%%%%%%%%%%%%%%%%%%%%%%%%%%%%%%%%%%%%%%%%%%%%%%%%%%%%%%%%%%%%%%%
%%%%%%%%%%%%%%%%%%%%%%%%%%%%%%%%%%%%%%%%%%%%%%%%%%%%%%%%%%%%%%%%%%%%%%%%%%%%%%%

\section{Experimental Validation}
\label{sec:experiments}

\subsection{Long-Horizon Time-Series Forecasting}

Time series forecasting presents persistent challenges, particularly for extended prediction horizons where error accumulation compounds modeling difficulties. We evaluate FlowMixer on benchmark datasets widely used in the forecasting literature, including ETT (electricity transformer temperature), Weather, Electricity, and Traffic datasets \cite{lai2018modeling}. Our comparisons include recent architectures such as TimeMixer++ \cite{timemixerplusplus2024}, Chimera \cite{chimera2024}, MSHyper \cite{mshyper2024, ada_mshyper2024}, CycleNet \cite{lin2024cyclenet}, and TSMixer \cite{chen2023tsmixer}. \revblue{FlowMixer deliberately forgoes the architectural complexity of these methods: no hierarchical decomposition, no multi-scale processing, no specialized seasonal modules. While the semi-group property guarantees composition e.g. $\mathcal{F}_{(96)} \circ \mathcal{F}_{(96)} = \mathcal{F}_{(192)}$, enabling algebraic horizon modification, we train each horizon independently for fair comparison (see Appendix~\ref{app:horizon}). The practical implication: practitioners can directly transfer FlowMixer between projects without architectural modifications.} Implementation details and hyperparameters are provided in Appendices~\ref{app:exp} and~\ref{app:hyper}, respectively.

\vspace{-0.125cm}
\begin{table}[!h]
\caption{Performance comparison of time series forecasting models on multiple datasets across different prediction horizons. Results show Mean Squared Error (MSE) and Mean Absolute Error (MAE), with lower values indicating better performance. \bv{Bold blue} is best, \bsv{orange} is second best. \revblue{FlowMixer achieves comparable performance using a single operational layer, eliminating the depth hyperparameter search required by all baseline methods. Statistical validation via Wilcoxon signed-rank tests confirms significance (p<0.05, see Appendix~\ref{app:wilcoxon}).}}

\label{tab:model-comparison}
\setlength{\tabcolsep}{1.0pt}
\small
\begin{tabular*}{\textwidth}{@{\extracolsep{\fill}}c@{}ccccccccccccccc}
\toprule
& & \multicolumn{2}{c}{\scriptsize\textbf{FlowMixer}} & \multicolumn{2}{c}{\scriptsize\textbf{Chimera}} & \multicolumn{2}{c}{\scriptsize\textbf{TimeMixer++}} & \multicolumn{2}{c}{\scriptsize\textbf{iTransformer}} & \multicolumn{2}{c}{\scriptsize\textbf{Ada-MSHyper}} & \multicolumn{2}{c}{\scriptsize\textbf{CycleNet}} & \multicolumn{2}{c}{\scriptsize\textbf{TSMixer}} \\
& & \multicolumn{2}{c}{\scriptsize(Ours)} & \multicolumn{2}{c}{\scriptsize(2024)} & \multicolumn{2}{c}{\scriptsize(2024)} & \multicolumn{2}{c}{\scriptsize(2024)} & \multicolumn{2}{c}{\scriptsize(2024)} & \multicolumn{2}{c}{\scriptsize(2024)} & \multicolumn{2}{c}{\scriptsize(2023)} \\
\cmidrule(lr){3-4}\cmidrule(lr){5-6}\cmidrule(lr){7-8}\cmidrule(lr){9-10}\cmidrule(lr){11-12}\cmidrule(lr){13-14}\cmidrule(lr){15-16}
& h & MSE & MAE & MSE & MAE & MSE & MAE & MSE & MAE & MSE & MAE & MSE & MAE & MSE & MAE \\
\midrule
\multirow{4}{*}{\rotatebox[origin=c]{90}{ETTh1}}
& 96  & \bv{0.355} & \bv{0.387} & 0.366 & \bsv{0.392} & \bsv{0.361} & 0.403 & 0.386 & 0.405 & 0.372 & 0.393 & 0.375 & 0.395 & \bsv{0.361} & \bsv{0.392} \\
& 192 & \bv{0.394} & \bv{0.409} & \bsv{0.402} & \bsv{0.414} & 0.416 & 0.441 & 0.441 & 0.436 & 0.433 & 0.417 & 0.436 & 0.428 & 0.404 & 0.418 \\
& 336 & \bsv{0.415} & \bsv{0.423} & \bv{0.406} & \bv{0.419} & 0.430 & 0.434 & 0.487 & 0.458 & 0.422 & 0.433 & 0.496 & 0.455 & 0.420 & 0.431 \\
& 720 & \bv{0.433} & \bsv{0.454} & 0.458 & 0.477 & 0.467 & \bv{0.451} & 0.503 & 0.491 & \bsv{0.445} & 0.459 & 0.520 & 0.484 & 0.463 & 0.472 \\
\midrule
\multirow{4}{*}{\rotatebox[origin=c]{90}{ETTh2}}
& 96  & \bsv{0.264} & 0.330 & \bv{0.262} & \bv{0.327} & 0.276 & \bsv{0.328} & 0.297 & 0.349 & 0.283 & 0.332 & 0.298 & 0.344 & 0.274 & 0.341 \\
& 192 & \bv{0.320} & \bv{0.368} & \bv{0.320} & \bsv{0.372} & 0.342 & 0.379 & 0.380 & 0.400 & 0.358 & 0.374 & 0.372 & 0.396 & 0.339 & 0.385 \\
& 336 & \bsv{0.344} & \bsv{0.396} & \bv{0.316} & \bv{0.381} & 0.346 & 0.398 & 0.428 & 0.432 & 0.428 & 0.437 & 0.431 & 0.439 & 0.361 & 0.406 \\
& 720 & \bv{0.388} & 0.452 & \bsv{0.389} & \bsv{0.430} & 0.392 & \bv{0.415} & 0.427 & 0.445 & 0.413 & 0.432 & 0.450 & 0.458 & 0.445 & 0.470 \\
\midrule
\multirow{4}{*}{\rotatebox[origin=c]{90}{ETTm1}}
& 96  & \bsv{0.298} & 0.345 & 0.318 & 0.354 & 0.310 & \bv{0.334} & 0.334 & 0.368 & 0.301 & 0.354 & 0.319 & 0.360 & \bv{0.285} & \bsv{0.339} \\
& 192 & 0.333 & \bsv{0.365} & \bsv{0.331} & 0.369 & 0.348 & \bv{0.362} & 0.377 & 0.391 & 0.345 & 0.375 & 0.360 & 0.381 & \bv{0.327} & \bsv{0.365} \\
& 336 & \bsv{0.360} & \bsv{0.386} & 0.363 & 0.389 & 0.376 & 0.391 & 0.426 & 0.420 & 0.375 & 0.397 & 0.389 & 0.403 & \bv{0.356} & \bv{0.382} \\
& 720 & \bv{0.398} & \bv{0.410} & \bsv{0.409} & 0.415 & 0.440 & 0.423 & 0.491 & 0.459 & 0.437 & 0.435 & 0.447 & 0.441 & 0.419 & \bsv{0.414} \\
\midrule
\multirow{4}{*}{\rotatebox[origin=c]{90}{ETTm2}}
& 96  & \bv{0.159} & 0.252 & 0.169 & 0.265 & 0.170 & \bv{0.245} & 0.180 & 0.264 & 0.165 & 0.257 & \bsv{0.163} & \bsv{0.249} & \bsv{0.163} & 0.252 \\
& 192 & \bv{0.211} & \bv{0.290} & 0.221 & \bv{0.290} & 0.229 & 0.291 & 0.250 & 0.309 & 0.230 & 0.307 & 0.229 & \bv{0.290} & \bsv{0.216} & \bv{0.290} \\
& 336 & \bv{0.260} & \bv{0.324} & 0.279 & 0.339 & 0.303 & 0.343 & 0.311 & 0.348 & 0.282 & 0.328 & 0.284 & 0.327 & \bsv{0.268} & \bv{0.324} \\
& 720 & \bv{0.333} & \bsv{0.377} & \bsv{0.341} & \bv{0.376} & 0.373 & 0.399 & 0.412 & 0.407 & 0.375 & 0.396 & 0.389 & 0.391 & 0.420 & 0.422 \\
\midrule
\multirow{4}{*}{\rotatebox[origin=c]{90}{Weather}}
& 96  & \bv{0.143} & \bv{0.194} & 0.146 & 0.206 & 0.155 & 0.205 & 0.174 & 0.214 & 0.157 & \bsv{0.195} & 0.158 & 0.203 & \bsv{0.145} & 0.198 \\
& 192 & \bv{0.185} & \bv{0.235} & \bsv{0.189} & \bsv{0.239} & 0.201 & 0.245 & 0.221 & 0.254 & 0.218 & 0.259 & 0.207 & 0.247 & 0.191 & 0.242 \\
& 336 & \bv{0.235} & 0.276 & 0.244 & 0.281 & \bsv{0.237} & \bsv{0.265} & 0.278 & 0.296 & 0.251 & \bv{0.252} & 0.262 & 0.289 & 0.242 & 0.280 \\
& 720 & 0.305 & \bsv{0.326} & \bv{0.297} & \bv{0.309} & 0.312 & 0.334 & 0.358 & 0.347 & \bsv{0.304} & 0.328 & 0.344 & 0.344 & 0.320 & \bsv{0.326} \\
\midrule
\multirow{4}{*}{\rotatebox[origin=c]{90}{Electricity}}
& 96  & \bv{0.131} & \bsv{0.226} & 0.132 & 0.234 & 0.135 & \bv{0.222} & 0.148 & 0.240 & 0.135 & 0.238 & 0.136 & 0.229 & \bv{0.131} & 0.229 \\
& 192 & \bsv{0.146} & 0.239 & \bv{0.144} & \bv{0.223} & 0.147 & \bsv{0.235} & 0.162 & 0.253 & 0.152 & 0.239 & 0.152 & 0.244 & 0.151 & 0.246 \\
& 336 & \bsv{0.160} & \bsv{0.255} & \bv{0.156} & 0.259 & 0.164 & \bv{0.245} & 0.178 & 0.269 & 0.168 & 0.266 & 0.170 & 0.264 & 0.161 & 0.261 \\
& 720 & \bsv{0.195} & \bsv{0.289} & \bv{0.184} & \bv{0.280} & 0.212 & 0.310 & 0.225 & 0.317 & 0.212 & 0.293 & 0.212 & 0.299 & 0.197 & 0.293 \\
\midrule
\multirow{4}{*}{\rotatebox[origin=c]{90}{Traffic}}
& 96  & 0.377 & 0.264 & \bv{0.366} & \bv{0.248} & 0.392 & 0.253 & 0.395 & 0.268 & 0.384 & \bv{0.248} & 0.458 & 0.296 & \bsv{0.376} & 0.264 \\
& 192 & \bv{0.388} & 0.268 & \bsv{0.394} & 0.292 & 0.402 & \bv{0.258} & 0.417 & 0.276 & 0.401 & \bv{0.258} & 0.457 & 0.294 & 0.397 & 0.277 \\
& 336 & \bv{0.401} & 0.275 & \bsv{0.409} & 0.311 & 0.428 & \bsv{0.263} & 0.433 & 0.283 & 0.423 & \bv{0.261} & 0.470 & 0.299 & 0.413 & 0.290 \\
& 720 & \bv{0.434} & 0.291 & 0.443 & 0.294 & \bsv{0.441} & \bv{0.282} & 0.467 & 0.302 & 0.453 & \bv{0.282} & 0.502 & 0.314 & 0.444 & 0.306 \\
\bottomrule
\end{tabular*}
\vspace{-0.2cm}
\end{table}
%\vspace{-0.2cm}
Table~\ref{tab:model-comparison} demonstrates FlowMixer's competitive performance across multiple datasets and horizons, with particularly strong results on ETTm2 and the 720-timestep horizon for Traffic and ETT forecasting tasks. What distinguishes FlowMixer is its architectural simplicity\textemdash unlike contemporary models with deep architectures and hierarchical decompositions, it achieves these results consistently using just a single operational layer with carefully designed constraints. The computational efficiency directly benefits from this simplicity, with FlowMixer processing most ETT datasets in under 2 seconds per epoch (see Appendix~\ref{app:hyper}). Statistical validation through Wilcoxon signed-rank tests confirms that these improvements are significant (p<0.05, see Appendix~\ref{app:wilcoxon}). Ablation studies confirm these architectural choices contribute to the model's forecasting capabilities (Appendix~\ref{app:abla}), demonstrating how carefully targeted constraints enhance multivariate time series prediction.

\subsection{Kronecker-Koopman Analysis of Spatiotemporal Patterns}
\label{sec:KK}
FlowMixer offers a novel perspective on spatiotemporal analysis beyond the traditional Dynamic Mode Decomposition (DMD) framework \cite{schmid2010dynamic}. Combining time and feature spectral elements yields the Kronecker-Koopman (KK) eigenmodes (see Section~\ref{sec:theory}). To illustrate these patterns, we compute KK eigenmodes from the traffic dataset~\cite{zhou2021informer} enabling effective visualization of spatial components.

\begin{figure}[!h]
\centering
\includegraphics[width=0.96\textwidth]{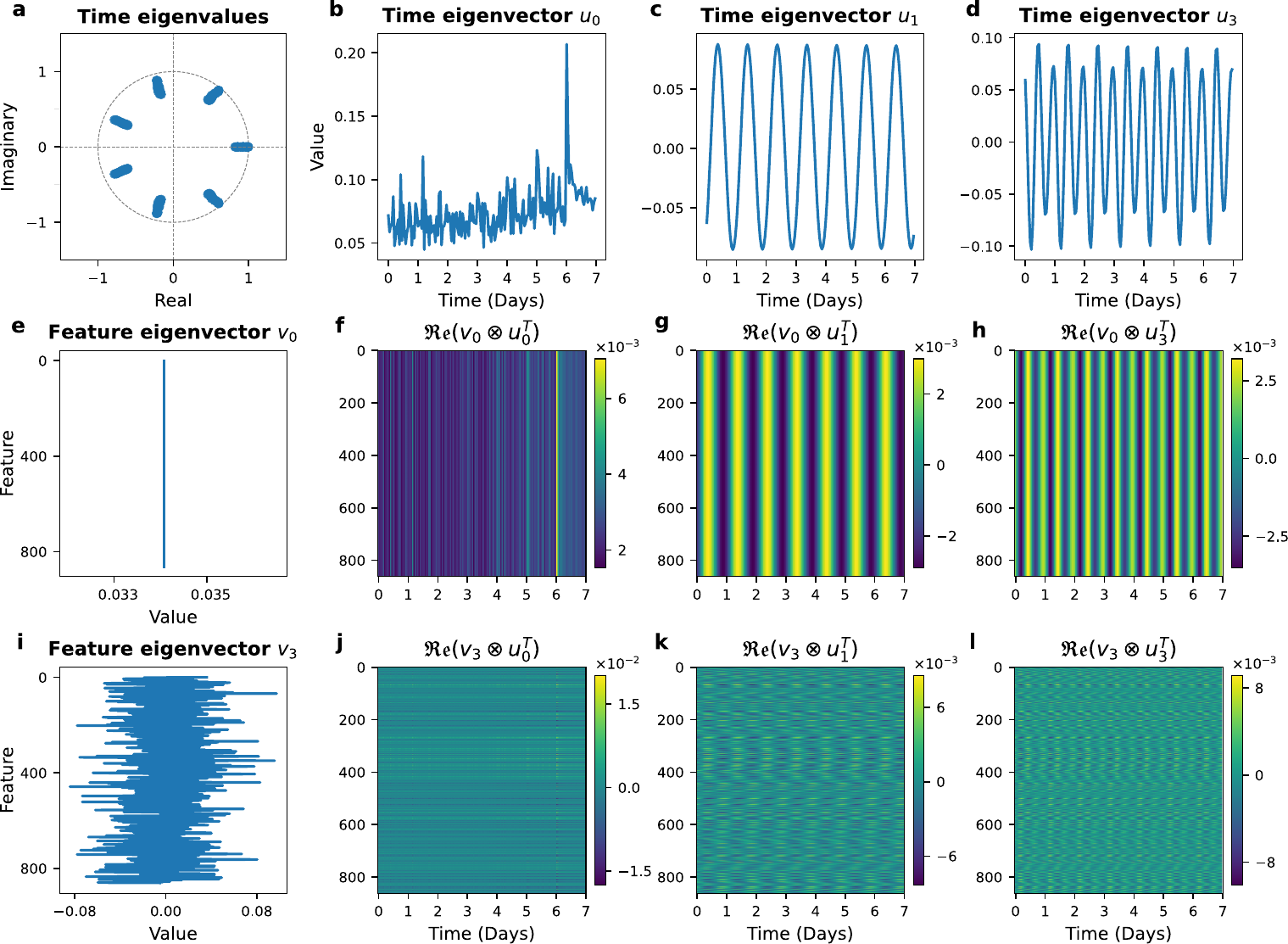}
\caption{Visualization of Kronecker-Koopman Eigenmodes for traffic dataset. (a) Time eigenvalues distribution, with aligned angular values indicating the periodic nature of the traffic datasets. (b-d) First three time eigenvectors (real part). (e,i) First and Third space eigenvectors, with the first showing constant values consistent with the stochastic space mixing matrix (representing a Markov transition process). (f-h, j-l) Real parts of Kronecker-Koopman Eigenmodes revealing space-time patterns. The periodic structures in time eigenvectors and coherent patterns in higher-mode products demonstrate how FlowMixer captures spatiotemporal dynamics.}
\label{fig:kro_koop}
\vspace{-0.2cm}
\end{figure}

When applied to traffic flow data, KK analysis yields structural insights. The time eigenvalues distribution (Figure~\ref{fig:kro_koop}a) reveals clear angular alignments corresponding to daily, weekly, and intra-day traffic periodicities. The primary time eigenvector (b) captures baseline (trend) traffic volume with weekend transitions, while subsequent eigenvectors (c-d) exhibit accelerating oscillatory behavior (seasonalities) reflecting 24-hour and 12-hour cycles. Feature eigenvectors (e-g) encode spatial relationships between traffic sensors from global to local scales, with the dominant eigenvector exhibiting the expected constant structure characteristic of left stochastic matrices. 
The resulting Kronecker-Koopman spectrum, especially higher-order modes (j-l), manifests coherent cross-dimensional patterns that reflect propagation phenomena consistent with congestion dynamics in traffic theory~\cite{treiber2013traffic, vanlint2002freeway}.

This framework provides spatiotemporal modes beyond classical DMD~\cite{schmid2010dynamic} and Hankelized-DMD~\cite{arbabi2017ergodic,brunton2017chaos} by separating temporal and spatial components instead of relying on Hankelized embeddings. Our approach captures the inherent structure of any system where spatiotemporal patterns govern dynamics, and expands the traditional time series dichotomy, from a classic trend/seasonality decomposition to a spatiotemporal framework:

\begin{equation}   \hat{y}_{t}=\hat{y}_{t}^{\text{trend}}+\hat{y}_{t}^{\text{seasonal}}\xrightarrow{\text{KK}}\left(\hat{y}_{t}^{\text{trend}}+\hat{y}_{t}^{\text{seasonal}}\right)\otimes\left(\hat{x}_{t}^{\text{global}}+\hat{x}_{t}^{\text{local}}\right)^T
\end{equation}

\subsection{Prediction of Chaotic Dynamical Systems}

The prediction of chaotic systems represents one of science's fundamental challenges, where exponential sensitivity to initial conditions traditionally limits long-term forecasting \cite{strogatz2018nonlinear}. Specialized architectures like reservoir computing have advanced this field significantly \cite{pathak2018model, vlachas2020backpropagation, sangiorgio2021forecasting, gauthier2021next, choi2024}, often focusing on recurrent frameworks for iterative single-step predictions. We explore how FlowMixer's constrained design principles can contribute to this domain through a complementary approach.

The key component is SOBR (Semi-Orthogonal Basic Reservoir)\textemdash an adaptation of reservoir computing principles that aligns with FlowMixer's constrained design philosophy (cf. Section~\ref{sec:components}). While Machine Learning approaches have traditionally addressed high-dimensional problems \cite{lecun2015deep}, chaotic systems present a different challenge: complex dynamics emerging from low dimensions (e.g., the three-dimensional Lorenz system). SOBR addresses this dimensional consideration through a targeted expansion: it projects data into higher-dimensional spaces using random semi-orthogonal matrices, maintaining full rank and unit spectral radius. This Koopman-inspired dimensional lifting enhances FlowMixer's pattern recognition capabilities.
FlowMixer's results for chaotic attractors without SOBR are provided in Appendix~\ref{app:noSOBR}.

SOBR integrates with FlowMixer through the reversible mapping $\phi$. In its simplest form it is expressed as $S\left(X\right) = \sigma(U_tXU_f^T)$, where semi-orthogonal matrices $U_f$ and $U_t$ expand feature and time dimensions respectively, complemented with an invertible activation $\sigma$ (typically Leaky ReLU). This design enhances FlowMixer's ability to capture chaotic attractor dynamics. \revblue{While this SOBR integration preserves the semi-group property only in the lifted space, not the original space (see Appendix~\ref{app:sobr_math})}, the benefits of dimensional lifting outweigh this tradeoff for chaotic systems, where accurate pattern recognition takes precedence over \revblue{direct} algebraic horizon manipulation.

\begin{figure*}[!h]
\centering
\includegraphics[width=0.98\textwidth]{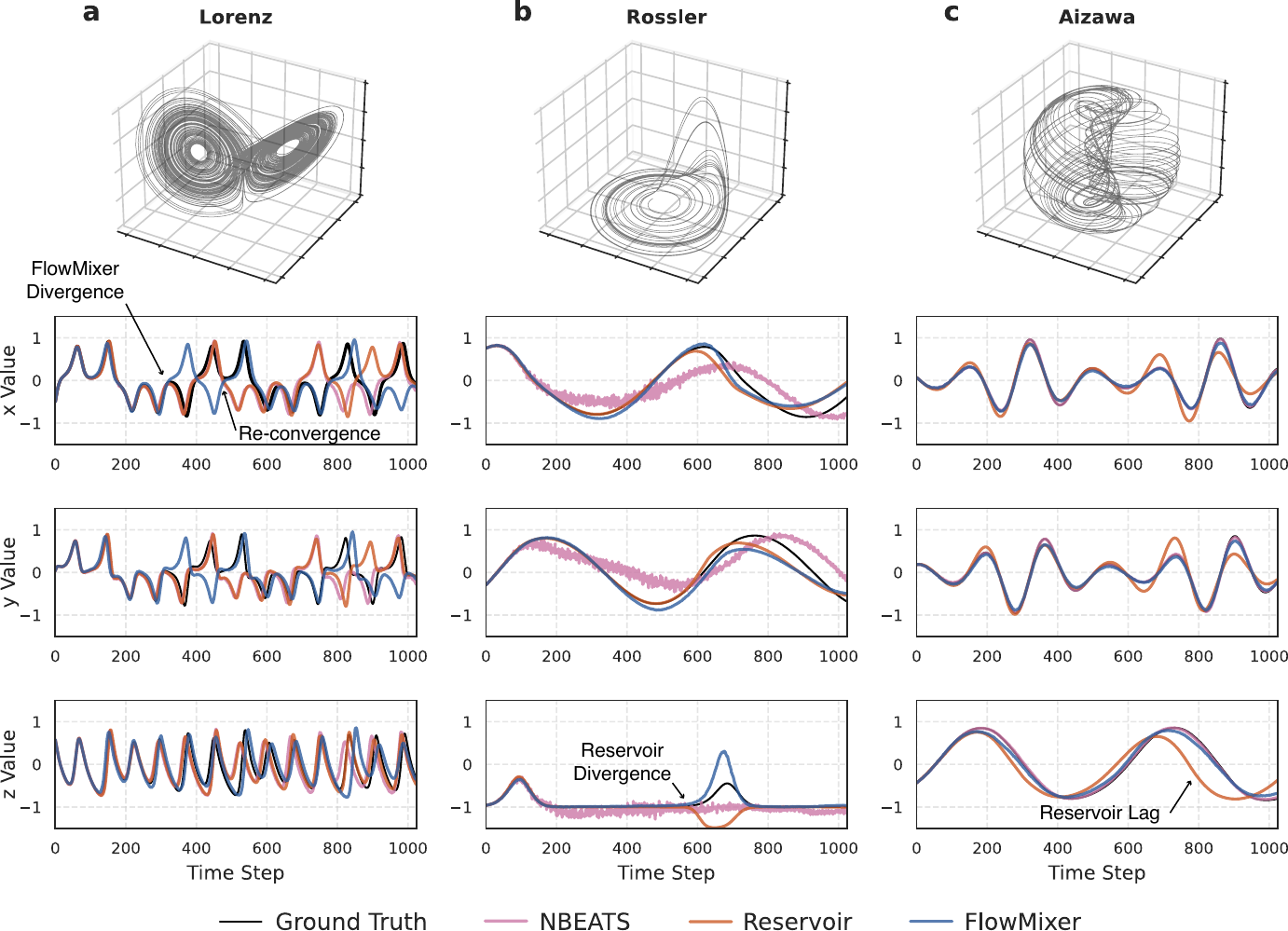}
\caption{Predictions of Lorenz (a), Rössler (b), and Aizawa (c) chaotic attractors (scaled [-1,1]). Each row shows the evolution of x, y, and z variables over time. Ground truth (black), FlowMixer (blue), Reservoir Computing (orange)~\cite{pathak2018model, platt2021robust,pathak2018hybrid}, and N-BEATS (purple)~\cite{oreshkin2020n} trajectories are compared. While all methods capture the Lorenz attractor's structure, N-BEATS shows a notable deviation in the Rössler system, and Reservoir Computing shows differences particularly in the $z$-component predictions. FlowMixer exhibits consistent performance across all three systems. Experimental settings are summarized in Appendix~\ref{app:hyper} and presented in detail in Appendix~\ref{app:extra_hyper_chaos}.}
\label{fig:chaotic_attractors}
%\vskip -0.25 cm
\end{figure*}

We evaluate this approach on three canonical chaotic systems: the Lorenz butterfly \cite{lorenz1963deterministic}, Rössler attractor \cite{rossler1976equation}, and Aizawa system \cite{aizawa1994chaos}. Rather than employing iterative single-step predictions, FlowMixer processes 32-step sequences to forecast the next 32 steps directly. By compounding these predictions, we obtain forecasts over 1024 steps\textemdash approximately 9-10 Lyapunov times for the Lorenz system. Figure \ref{fig:chaotic_attractors} compares FlowMixer's performance with Reservoir Computing (RC) \cite{pathak2018model, platt2021robust,pathak2018hybrid} and N-BEATS \cite{gilpin2021chaos,oreshkin2020n}. \revblue{While all methods capture the general attractor structure, they exhibit different failure modes: RC struggles with accurate $z$-component predictions particularly in the Lorenz system, N-BEATS diverges notably in the Rössler system, while FlowMixer maintains consistent accuracy across all three attractors.} The model captures characteristic features from the Lorenz butterfly to the Aizawa spiral dynamics, occasionally realigning with ground truth trajectories after divergence\textemdash suggesting effective pattern learning.
These results illustrate how architectural constraints can \revblue{maintain} predictive capabilities for chaotic systems.
FlowMixer with SOBR offers an alternative approach that complements existing specialized frameworks. \revblue{This same single-layer architecture that forecasts time series also captures chaotic dynamics}. This connection between statistical and dynamical modeling suggests broader implications for scientific computing, which we explore next through turbulent flow prediction.

\subsection{Predicting 2D Turbulent Flows}

The prediction of turbulent flows represents a crucial challenge at the intersection of computational physics and engineering \cite{brunton2020machine}. While direct numerical simulation of the Navier-Stokes equations provides high accuracy, its computational demands limit practical applications. Recent physics-informed neural networks (PINNs) \cite{raissi2019physics, karniadakis2021physics} have made valuable contributions by embedding physical constraints into learning architectures, though these approaches often involve balancing multiple loss terms and addressing convergence considerations \cite{wang2022when}.

We demonstrate how FlowMixer's constrained architecture offers a complementary data-driven approach on two canonical test cases: two-dimensional flow past a cylinder at Reynolds number $Re = 150$ and flow around a NACA airfoil at $Re = 1000$ with 15° angle of attack. For 2D fluid dynamics prediction, FlowMixer processes the vorticity field $\omega$ directly, with the core transformation applied as:

\begin{equation}
\omega_{t+1:t+h} = \phi^{-1}(W_t\phi(\omega_{t-h+1:t})W_f^T)
\end{equation}

where $\omega_{t-h+1:t}$ represents a sequence of $h$ consecutive vorticity snapshots, and $\omega_{t+1:t+h}$ is the predicted sequence for the next $h$ timesteps.

\begin{figure*}[h!]
\centering
\includegraphics[width=0.98\textwidth]{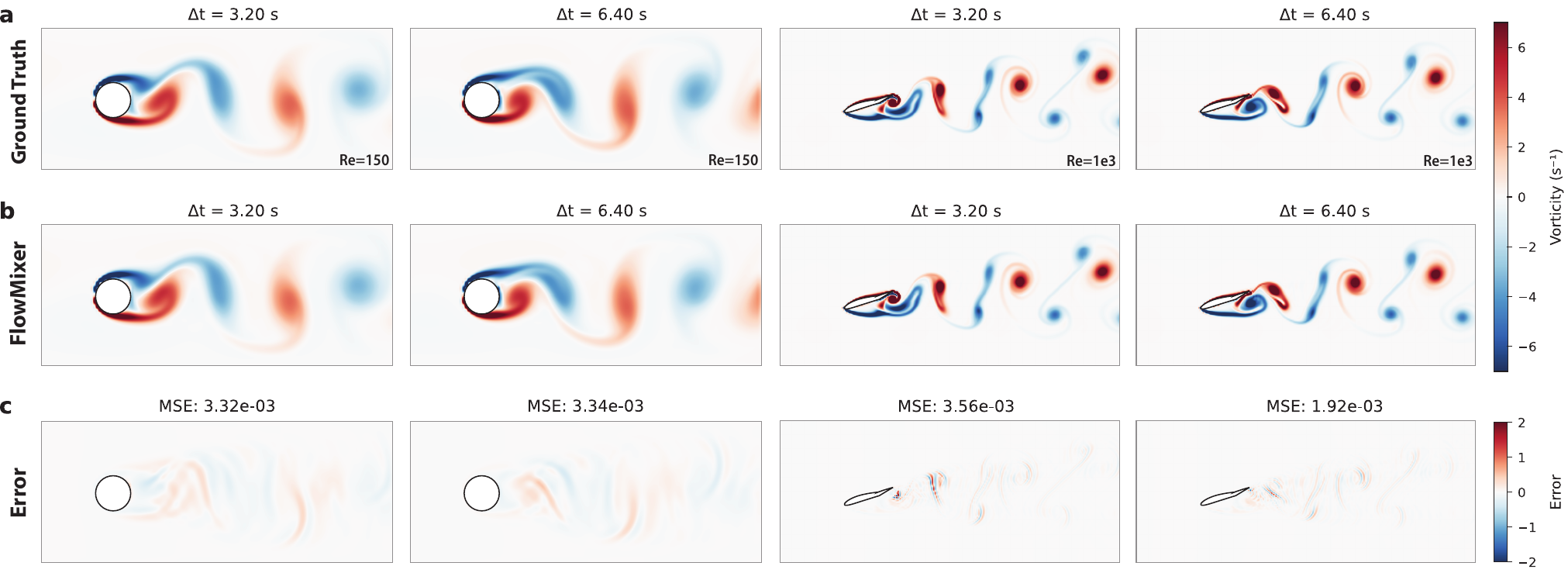}
\caption{Prediction of vorticity fields for flow past a cylinder at $Re=150$ (left columns) and a NACA airfoil at $Re=1000$ (right columns). (a) Ground truth vorticity fields showing the evolution of wake structures. (b) FlowMixer predictions demonstrate accurate capture of both near-body structures and downstream vortex evolution. (c) Error fields with corresponding Mean Squared Error (MSE) values show minimal discrepancy between prediction and ground truth across varying geometries and flow regimes. Experimental details are provided in Appendix~\ref{app:2dturb}. Additional flow visualizations and analyses are provided in Appendix~\ref{app:convLSTM} and Appendix~\ref{app:SGDvsADAM}.}
\label{fig:vorticity_predictions}
\end{figure*}

Our simulations capture the full dynamics across computational domains with $400 \times 160$ resolution. The cylinder flow is characterized by the von Kármán vortex street with a Strouhal number of approximately 0.2, while the airfoil case introduces asymmetric geometry, flow separation, and higher Reynolds number effects. For both cases, we train FlowMixer to process 64 consecutive flow field snapshots and predict the next 64 timesteps $(h=64)$\textemdash representing physical time horizons of 6.4 seconds for the cylinder and the airfoil. Figure \ref{fig:vorticity_predictions} compares the predicted vorticity fields with ground truth. For the cylinder case, FlowMixer accurately captures the alternating clockwise (blue) and counterclockwise (red) rotation patterns in the wake. In the more challenging airfoil case, the model successfully predicts the complex separation patterns, asymmetric wake structures, and vortex interactions despite the substantially higher Reynolds number.

Quantitatively, FlowMixer maintains MSE values of approximately 3.3e-3 for the cylinder case and 2e-3 to 8e-3 for the airfoil case, corresponding to approximately 4-5\% relative $l^2$ error. These results \revblue{are competitive} with existing approaches in the literature \cite{raissi2020hidden, wang2020towards, guemes2021coarse}, while using standard stochastic gradient descent with momentum for training rather than specialized optimization procedures (comparisons of optimization strategies are presented in Appendix~\ref{app:SGDvsADAM}).

FlowMixer adapts to increased flow complexity without architectural modifications. When transitioning from the regular vortex shedding of the cylinder at $Re=150$ to the higher Reynolds number airfoil case $Re=1000$ with complex separation and wake interactions, the model maintains comparable accuracy using the same architectural constraints. The error fields (Figure \ref{fig:vorticity_predictions}c) show relatively uniform distribution without significant localized spikes, suggesting the model effectively captures dominant flow patterns across different regimes. This behavior is noteworthy given that other approaches often require explicit conservation constraints to achieve stable predictions across varying flow regimes \cite{mohan2020embedding, wang2021learning, beucler2021enforcing}.

The successful prediction of complex flow patterns around both simple and complex geometries at different Reynolds numbers demonstrates the versatility of FlowMixer's architecture. The ability to capture these dynamics without specialized physical constraints points toward applications in aerodynamic design optimization, real-time flow monitoring, and reduced-order modeling for engineering systems where both computational efficiency and accuracy are essential requirements. %\revblue{The model's ability to handle both canonical test cases and complex separated flows without modification suggests that constrained architectural design can effectively capture fluid dynamics across diverse flow regimes.}

%%%%%%%%%%%%%%%%%%%%%%%%%%%%%%%%%%%%%%%%%%%%%%%%%%%%%%%%%%%%%%%%%%%%%%%%%%%%%%%
%%%%%%%%%%%%%%%%%%%%%%%%%%%%%%%%%%%%%%%%%%%%%%%%%%%%%%%%%%%%%%%%%%%%%%%%%%%%%%%
% Conclusion
%%%%%%%%%%%%%%%%%%%%%%%%%%%%%%%%%%%%%%%%%%%%%%%%%%%%%%%%%%%%%%%%%%%%%%%%%%%%%%%
%%%%%%%%%%%%%%%%%%%%%%%%%%%%%%%%%%%%%%%%%%%%%%%%%%%%%%%%%%%%%%%%%%%%%%%%%%%%%%%

\section*{Conclusion}
FlowMixer demonstrates how architectural constraints inspired by dynamical systems theory enhance both predictive capabilities and interpretability in neural forecasting systems. Through extensive validation across multiple domains, we have shown this constrained approach offers \revblue{competitive} performance in long-horizon forecasting while effectively modeling complex physical phenomena such as chaotic attractors and turbulent flows.

The architecture introduces three complementary innovations: (1) constrained non-negative matrix mixing operations, (2) a Kronecker-Koopman framework providing interpretable spatiotemporal patterns, and (3) a Semi-Orthogonal Basic Reservoir approach enabling stable prediction of chaotic systems. Together, these elements create a unified framework bridging statistical learning and dynamical systems modeling.

Our results highlight an important principle in neural architecture design: carefully chosen mathematical constraints can simultaneously maintain competitive performance while enhancing interpretability without necessitating increased model complexity. \revblue{Notably, the semi-group composition property eliminates depth optimization entirely: one layer is all you need. This frees practitioners from a critical deployment bottleneck that affects all existing architectures.}

We acknowledge current limitations, including cubic computational complexity in sequence length and the absence of explicit physical conservation guarantees. Future work will explore sparsification techniques to improve computational scaling and investigate integration with physics-informed approaches to enhance scientific applications.

As deep learning advances scientific computing, architectures like FlowMixer demonstrate the value of incorporating domain knowledge through mathematical design principles rather than explicit regularization\textemdash offering a promising direction for neural forecasting systems that are both effective and interpretable across scientific and engineering applications.

%%%%%%%%%%%%%%%%%%%%%%%%%%%%%%%%%%%%%%%%%%%%%%%%%%%%%%%%%%%%%%%%%%%%%%%%%%%%%%%
%%%%%%%%%%%%%%%%%%%%%%%%%%%%%%%%%%%%%%%%%%%%%%%%%%%%%%%%%%%%%%%%%%%%%%%%%%%%%%%
% BIBLIOGRAPHY
%%%%%%%%%%%%%%%%%%%%%%%%%%%%%%%%%%%%%%%%%%%%%%%%%%%%%%%%%%%%%%%%%%%%%%%%%%%%%%%
%%%%%%%%%%%%%%%%%%%%%%%%%%%%%%%%%%%%%%%%%%%%%%%%%%%%%%%%%%%%%%%%%%%%%%%%%%%%%%%

\newpage

\section*{Code Availability}
The code for FlowMixer is available at \url{https://github.com/FaresBMehouachi/FlowMixer}. The authors declare no competing interests.
\section*{Acknowledgment}
This work was supported by the NYUAD Center for Interacting Urban Networks (CITIES), funded by Tamkeen under the NYUAD Research Institute Award CG001. The views expressed in this article are those of the authors and do not reflect the opinions of CITIES or their funding agencies
%\bibliography{main}
%\bibliography{main_fixed}
\bibliography{main_verified}
\bibliographystyle{unsrt}

%%%%%%%%%%%%%%%%%%%%%%%%%%%%%%%%%%%%%%%%%%%%%%%%%%%%%%%%%%%%

%\appendix
%\include{appendix/app_v5}

%%%%%%%%%%%%%%%%%%%%%%%%%%%%%%%%%%%%%%%%%%%%%%%%%%%%%%%%%%%%

%%%%%%%%%%%%%%%%%%%%%%
%%%%%%%%%%%%%%%%%%%%%%
%%%%%%%%%%%%%%%%%%%%%%
%%% APPENDIX
%%%%%%%%%%%%%%%%%%%%%%
%%%%%%%%%%%%%%%%%%%%%%
%%%%%%%%%%%%%%%%%%%%%%
\appendix

%%%%%%%%%%%%%%%%%%%%%%%%%%%%%%%%%%%%%%%%%%%%%%%%%%%%%%%%%%%%%%%%%%%%%%%%%%%%%%%
%%%%%%%%%%%%%%%%%%%%%%%%%%%%%%%%%%%%%%%%%%%%%%%%%%%%%%%%%%%%%%%%%%%%%%%%%%%%%%%
% APPENDIX Kronecker-Koopman Eigenmodes
%%%%%%%%%%%%%%%%%%%%%%%%%%%%%%%%%%%%%%%%%%%%%%%%%%%%%%%%%%%%%%%%%%%%%%%%%%%%%%%
%%%%%%%%%%%%%%%%%%%%%%%%%%%%%%%%%%%%%%%%%%%%%%%%%%%%%%%%%%%%%%%%%%%%%%%%%%%%%%%

%%%%%%%%%%%%%%%%%%%%%%%%%%%%%%%%%%%%%%%%%%%%%%%%%%%%%%%%%%%%%%%%%%%%%%%%%%%%%%%
%%%%%%%%%%%%%%%%%%%%%%%%%%%%%%%%%%%%%%%%%%%%%%%%%%%%%%%%%%%%%%%%%%%%%%%%%%%%%%%
% APPENDIX Experimental Details
%%%%%%%%%%%%%%%%%%%%%%%%%%%%%%%%%%%%%%%%%%%%%%%%%%%%%%%%%%%%%%%%%%%%%%%%%%%%%%%
%%%%%%%%%%%%%%%%%%%%%%%%%%%%%%%%%%%%%%%%%%%%%%%%%%%%%%%%%%%%%%%%%%%%%%%%%%%%%%%
\newpage
\section{Appendix: Long-Horizon Time Series Forecasting}
\label{app:exp}
\subsection{Experimental Details}

We evaluated FlowMixer using seven benchmark datasets encompassing diverse domains with varying temporal resolutions and prediction challenges. The ETT datasets capture electrical transformer measurements over a two-year period, consisting of both hourly (ETTh) and 15-minute (ETTm) resolution variants. The Electricity dataset records hourly power consumption from 321 customers, while Traffic contains hourly road occupancy rates from California. The Weather dataset comprises 21 meteorological indicators sampled at 10-minute intervals throughout 2020.

Data preparation followed standard protocols, with chronological partitioning using train/validation/test ratios of 0.6/0.2/0.2 for ETT and 0.7/0.2/0.1 for others. Each feature underwent independent z-score normalization. For most datasets we use an input length of 1024 ($\geq$720) timesteps. For Electricity and Traffic, we utilize longer input sequences (up to 3072 timesteps) due to their increased feature dimensionality.

Model training was implemented in TensorFlow and executed on a single NVIDIA A100 GPU. The training protocol incorporated early stopping with patience of 10 epochs and learning rate reduction (factor 0.1, patience 5) based on validation loss. We limited training to 100 epochs with batch size 32 and systematically evaluated dropout rates between 0.0 and 0.7. The model's hyperparameters were optimized for each dataset and prediction horizon combination, with comprehensive configurations detailed in Table~\ref{tab:hyperparams}. While training hyperparameters were optimized per dataset, the core architecture remained fixed with a single mixing block\textemdash eliminating the depth search required by baseline methods.

\subsection{Extra comparison}

\begin{table}[!h]
\caption{Extra comparison of FlowMixer with N-HITS~\cite{challu2022n}, FEDformer~\cite{zhou2022fedformer}, Autoformer~\cite{wu2021autoformer}, Dlinear~\cite{zeng2023transformers}, and some Koopman based models: KooPA~\cite{liu2023koopa} and KNF~\cite{yu2023koopman}. The reported standard deviations for FlowMixer use five seeds. \bv{Bold blue} indicate best performance, demonstrating FlowMixer's competitive results.}\label{tab:extended-comparison}
\setlength{\tabcolsep}{0.9pt}
\scriptsize
\begin{tabular*}{\textwidth}{@{\extracolsep{\fill}}c@{}ccccccccccccccc}
\toprule
& & \multicolumn{2}{c}{\scriptsize\textbf{FlowMixer}} & \multicolumn{2}{c}{\scriptsize\textbf{N-HITS}} & \multicolumn{2}{c}{\scriptsize\textbf{FEDformer}} & \multicolumn{2}{c}{\scriptsize\textbf{Autoformer}} & \multicolumn{2}{c}{\scriptsize\textbf{DLinear}} & \multicolumn{2}{c}{\scriptsize\textbf{KooPA}} & \multicolumn{2}{c}{\scriptsize\textbf{KNF}} \\
& & \multicolumn{2}{c}{\scriptsize(Ours)} & \multicolumn{2}{c}{\scriptsize(2021)} & \multicolumn{2}{c}{\scriptsize(2022)} & \multicolumn{2}{c}{\scriptsize(2021)} & \multicolumn{2}{c}{\scriptsize(2023)} & \multicolumn{2}{c}{\scriptsize(2023)} & \multicolumn{2}{c}{\scriptsize(2023)} \\
\cmidrule(lr){3-4}\cmidrule(lr){5-6}\cmidrule(lr){7-8}\cmidrule(lr){9-10}\cmidrule(lr){11-12}\cmidrule(lr){13-14}\cmidrule(lr){15-16}
& h & MSE & MAE & MSE & MAE & MSE & MAE & MSE & MAE & MSE & MAE & MSE & MAE & MSE & MAE \\
\midrule
\multirow{4}{*}{\rotatebox[origin=c]{90}{ETTh1}}
& 96  & \bv{0.358$\pm$.001} & \bv{0.390$\pm$.001} & 0.475 & 0.498 & 0.376 & 0.415 & 0.435 & 0.446 & 0.386 & 0.400 & 0.371 & 0.405 & 0.975 & 0.744 \\
& 192 & \bv{0.394$\pm$.001} & \bv{0.412$\pm$.001} & 0.426 & 0.519 & 0.423 & 0.446 & 0.456 & 0.457 & 0.437 & 0.432 & 0.416 & 0.439 & 0.941 & 0.744 \\
& 336 & \bv{0.418$\pm$.001} & \bv{0.430$\pm$.001} & 0.550 & 0.564 & 0.444 & 0.462 & 0.486 & 0.487 & 0.481 & 0.459 & -- & -- & -- & -- \\
& 720 & \bv{0.465$\pm$.001} & \bv{0.475$\pm$.001} & 0.598 & 0.641 & 0.469 & 0.492 & 0.515 & 0.517 & 0.519 & 0.516 & -- & -- & -- & -- \\
\midrule
\multirow{4}{*}{\rotatebox[origin=c]{90}{ETTh2}}
& 96  & \bv{0.264$\pm$.001} & \bv{0.330$\pm$.001} & 0.328 & 0.364 & 0.332 & 0.374 & 0.332 & 0.368 & 0.333 & 0.387 & 0.297 & 0.349 & 0.433 & 0.446 \\
& 192 & \bv{0.320$\pm$.002} & \bv{0.368$\pm$.001} & 0.372 & 0.408 & 0.407 & 0.446 & 0.426 & 0.434 & 0.477 & 0.476 & 0.356 & 0.393 & 0.528 & 0.503 \\
& 336 & \bv{0.344$\pm$.002} & \bv{0.396$\pm$.001} & 0.397 & 0.421 & 0.400 & 0.447 & 0.477 & 0.479 & 0.594 & 0.541 & -- & -- & -- & -- \\
& 720 & \bv{0.420$\pm$.002} & \bv{0.452$\pm$.002} & 0.461 & 0.497 & 0.412 & 0.469 & 0.453 & 0.453 & 0.831 & 0.657 & -- & -- & -- & -- \\
\midrule
\multirow{4}{*}{\rotatebox[origin=c]{90}{ETTm1}}
& 96  & \bv{0.298$\pm$.002} & \bv{0.345$\pm$.002} & 0.370 & 0.468 & 0.326 & 0.390 & 0.510 & 0.492 & 0.345 & 0.372 & 0.294 & 0.345 & 0.957 & 0.782 \\
& 192 & \bv{0.333$\pm$.001} & \bv{0.365$\pm$.001} & 0.436 & 0.488 & 0.365 & 0.415 & 0.515 & 0.514 & 0.380 & 0.389 & 0.337 & 0.378 & 0.896 & 0.731 \\
& 336 & \bv{0.363$\pm$.001} & \bv{0.383$\pm$.001} & 0.483 & 0.510 & 0.392 & 0.425 & 0.510 & 0.492 & 0.413 & 0.413 & -- & -- & -- & -- \\
& 720 & \bv{0.404$\pm$.001} & \bv{0.410$\pm$.001} & 0.489 & 0.537 & 0.446 & 0.458 & 0.527 & 0.493 & 0.474 & 0.433 & -- & -- & -- & -- \\
\midrule
\multirow{4}{*}{\rotatebox[origin=c]{90}{ETTm2}}
& 96  & \bv{0.159$\pm$.001} & \bv{0.252$\pm$.001} & 0.176 & 0.255 & 0.180 & 0.271 & 0.205 & 0.293 & 0.192 & 0.282 & 0.171 & 0.254 & 1.535 & 1.012 \\
& 192 & \bv{0.211$\pm$.001} & \bv{0.290$\pm$.001} & 0.245 & 0.305 & 0.252 & 0.318 & 0.278 & 0.336 & 0.284 & 0.362 & 0.226 & 0.298 & 1.355 & 0.908 \\
& 336 & \bv{0.260$\pm$.001} & \bv{0.324$\pm$.001} & 0.295 & 0.346 & 0.324 & 0.364 & 0.343 & 0.379 & 0.369 & 0.427 & -- & -- & -- & -- \\
& 720 & \bv{0.333$\pm$.001} & \bv{0.377$\pm$.001} & 0.401 & 0.413 & 0.410 & 0.420 & 0.414 & 0.419 & 0.554 & 0.522 & -- & -- & -- & -- \\
\midrule
\multirow{4}{*}{\rotatebox[origin=c]{90}{Weather}}
& 96  & \bv{0.143$\pm$.002} & \bv{0.194$\pm$.002} & 0.158 & 0.195 & 0.238 & 0.314 & 0.249 & 0.329 & 0.193 & 0.292 & 0.154 & 0.205 & 0.295 & 0.308 \\
& 192 & \bv{0.185$\pm$.002} & \bv{0.235$\pm$.002} & 0.211 & 0.247 & 0.275 & 0.325 & 0.325 & 0.325 & 0.228 & 0.296 & 0.193 & 0.241 & 0.462 & 0.437 \\
& 336 & \bv{0.235$\pm$.002} & \bv{0.276$\pm$.001} & 0.274 & 0.300 & 0.339 & 0.377 & 0.397 & 0.397 & 0.283 & 0.335 & -- & -- & -- & -- \\
& 720 & \bv{0.305$\pm$.001} & \bv{0.326$\pm$.002} & 0.351 & 0.353 & 0.351 & 0.409 & 1.042 & 1.216 & 0.345 & 0.381 & -- & -- & -- & -- \\
\midrule
\multirow{4}{*}{\rotatebox[origin=c]{90}{Electricity}}
& 96  & \bv{0.131$\pm$.001} & \bv{0.226$\pm$.003} & 0.147 & 0.249 & 0.186 & 0.302 & 0.196 & 0.302 & 0.197 & 0.282 & 0.136 & 0.236 & 0.198 & 0.284 \\
& 192 & \bv{0.146$\pm$.002} & \bv{0.239$\pm$.003} & 0.167 & 0.249 & 0.197 & 0.311 & 0.211 & 0.321 & 0.196 & 0.285 & 0.156 & 0.254 & 0.245 & 0.321 \\
& 336 & \bv{0.160$\pm$.004} & \bv{0.255$\pm$.002} & 0.186 & 0.290 & 0.213 & 0.328 & 0.214 & 0.328 & 0.209 & 0.301 & -- & -- & -- & -- \\
& 720 & \bv{0.195$\pm$.003} & \bv{0.289$\pm$.004} & 0.243 & 0.340 & 0.236 & 0.344 & 0.236 & 0.342 & 0.245 & 0.333 & -- & -- & -- & -- \\
\midrule
\multirow{4}{*}{\rotatebox[origin=c]{90}{Traffic}}
& 96  & \bv{0.377$\pm$.002} & \bv{0.264$\pm$.001} & 0.402 & 0.282 & 0.576 & 0.359 & 0.597 & 0.371 & 0.650 & 0.396 & 0.401 & 0.275 & 0.645 & 0.376 \\
& 192 & \bv{0.388$\pm$.002} & \bv{0.268$\pm$.002} & 0.420 & 0.297 & 0.610 & 0.380 & 0.607 & 0.382 & 0.598 & 0.370 & 0.403 & 0.284 & 0.699 & 0.405 \\
& 336 & \bv{0.401$\pm$.002} & \bv{0.275$\pm$.004} & 0.448 & 0.313 & 0.608 & 0.375 & 0.623 & 0.387 & 0.605 & 0.373 & -- & -- & -- & -- \\
& 720 & \bv{0.434$\pm$.003} & \bv{0.291$\pm$.004} & 0.539 & 0.353 & 0.621 & 0.375 & 0.639 & 0.395 & 0.645 & 0.394 & -- & -- & -- & -- \\
\bottomrule
\end{tabular*}
\end{table}

\clearpage
\section{Periodicities Analysis}
\label{app:periods}
\begin{figure}[!h]
\centering
\includegraphics[width=1.0\textwidth]{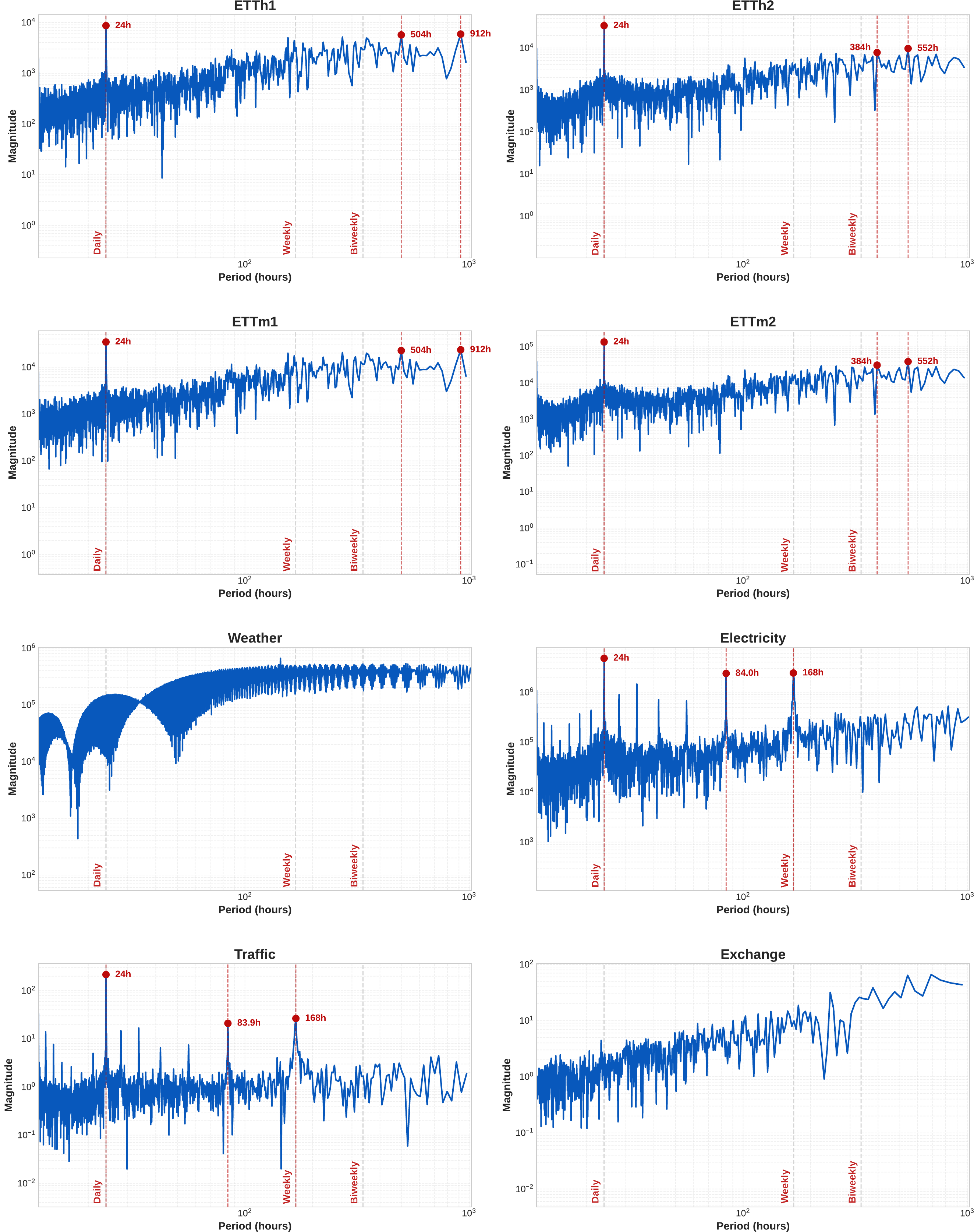}
\caption{ FFT Spectral Analysis of Periodicity Patterns in Benchmark Time Series Datasets. The plots show magnitude versus period (in hours) on logarithmic scales for eight standard forecasting datasets. Significant periodicities are marked with red dots and vertical lines, revealing prominent daily (24h) patterns across all datasets, with additional weekly (168h) and biweekly cycles of varying strengths. These spectral characteristics inform our model design,  particularly for incorporating appropriate time mixing components and periodicity-aware architectures to capture multi-scale temporal dependencies. We select mainly periods of [24,168] in our experiments. }
\end{figure}

%%%%%%%%%%%%%%%%%%%%%%%%%%%%%%%%%%%%%%%%%%%%%%%%%%%%%%%%%%%%%%%%%%%%%%%%%%%%%%%
%%%%%%%%%%%%%%%%%%%%%%%%%%%%%%%%%%%%%%%%%%%%%%%%%%%%%%%%%%%%%%%%%%%%%%%%%%%%%%%
% APPENDIX Hyperparameters for Experiments
%%%%%%%%%%%%%%%%%%%%%%%%%%%%%%%%%%%%%%%%%%%%%%%%%%%%%%%%%%%%%%%%%%%%%%%%%%%%%%%
%%%%%%%%%%%%%%%%%%%%%%%%%%%%%%%%%%%%%%%%%%%%%%%%%%%%%%%%%%%%%%%%%%%%%%%%%%%%%%%

\newpage
\section{Appendix: Hyperparameters for Experiments}
\label{app:hyper}
\begin{table}[!h]
\caption{Hyperparameter configurations for different forecasting horizons (h) and input lengths (Input Len.). Parameters include optimizer choice (Optim.), initial learning rate (Init. LR), weight decay (W. Decay), RevIN type, time per epoch (T/Epoch), exponential/quadratic time mixing (Expm), and periodicities. ``Turb." stands for turbulence.}\label{tab:hyperparams}
\setlength{\tabcolsep}{2pt}
\small
\begin{tabular*}{\textwidth}{@{\extracolsep{\fill}}ccccccccccc}
\toprule
& \textbf{h} & \textbf{Input Len.} & \textbf{Batch} & \textbf{Optim.} & \textbf{Init. LR} & \textbf{W. Decay} & \textbf{RevIN} & \textbf{Time/Epoch} & \textbf{Expm} & \textbf{Periodicities} \\
\midrule
\multirow{4}{*}{\rotatebox[origin=c]{90}{\textbf{ETTh1}}}
& 96  & 1344 & 16 & SGD & 1e-1 & 1e-6 & RevIN & 2s & True & {\footnotesize [24, 168]} \\
& 192 & 1344 & 16 & SGD & 1e-1 & 1e-6 & RevIN & 2s & True & {\footnotesize [24, 168]} \\
& 336 & 1344 & 16 & SGD & 1e-1 & 1e-6 & RevIN & 2s & True & {\footnotesize [24, 168]} \\
& 720 & 1344 & 16 & SGD & 1e-1 & 1e-6 & RevIN & 2s & True & {\footnotesize [24, 168]} \\
\midrule
\multirow{4}{*}{\rotatebox[origin=c]{90}{\textbf{ETTh2}}}
& 96  & 1344 & 16 & SGD & 1e-1 & 1e-5 & RevIN & 2s & False & {\footnotesize -} \\
& 192 & 1344 & 16 & SGD & 1e-1 & 1e-5 & RevIN & 2s & False & {\footnotesize -} \\
& 336 & 1344 & 16 & SGD & 1e-1 & 1e-5 & RevIN & 2s & True & {\footnotesize [24, 168]} \\
& 720 & 1344 & 16 & SGD & 1e-1 & 1e-5 & RevIN & 2s & True & {\footnotesize [24, 168]} \\
\midrule
\multirow{4}{*}{\rotatebox[origin=c]{90}{\textbf{ETTm1}}}
& 96  & 1344 & 16 & AdamW & 1e-3 & 1e-6 & TD-RevIN & 2s & False & {\footnotesize -} \\
& 192 & 1344 & 16 & AdamW & 1e-3 & 1e-6 & TD-RevIN & 2s & True & {\footnotesize [24 x 4, 168 x 4]} \\
& 336 & 1344 & 16 & AdamW & 1e-3 & 1e-6 & TD-RevIN & 2s & True & {\footnotesize [24 x 4, 168 x 4]} \\
& 720 & 1344 & 16 & AdamW & 1e-3 & 1e-6 & TD-RevIN & 2s & True & {\footnotesize [24 x 4, 168 x 4]} \\
\midrule
\multirow{4}{*}{\rotatebox[origin=c]{90}{\textbf{ETTm2}}}
& 96  & 1344 & 16 & SGD & 1e-1 & 1e-6 & RevIN & 2s & False & {\footnotesize -} \\
& 192 & 1344 & 16 & SGD & 1e-1 & 1e-6 & RevIN & 2s & False & {\footnotesize -} \\
& 336 & 1344 & 16 & SGD & 1e-1 & 1e-6 & RevIN & 2s & False & {\footnotesize -} \\
& 720 & 1344 & 16 & SGD & 1e-1 & 1e-6 & RevIN & 2s & False & {\footnotesize -} \\
\midrule
\multirow{4}{*}{\rotatebox[origin=c]{90}{\textbf{Weather}}}
& 96  & 1344 & 16 & AdamW & 1e-3 & 1e-6 & TD-RevIN & 8s & False & {\footnotesize -} \\
& 192 & 1344 & 16 & AdamW & 1e-3 & 1e-6 & TD-RevIN & 8s & False & {\footnotesize -} \\
& 336 & 1344 & 16 & AdamW & 1e-3 & 1e-6 & TD-RevIN & 8s & False & {\footnotesize -} \\
& 720 & 1344 & 16 & AdamW & 1e-3 & 1e-6 & TD-RevIN & 8s & True & {\footnotesize -} \\
\midrule
\multirow{4}{*}{\rotatebox[origin=c]{90}{\textbf{Electricity}}}
& 96  & 2688 & 16 & AdamW & 1e-4 & 1e-6 & RevIN & 122s & False & {\footnotesize -} \\
& 192 & 2688 & 16 & AdamW & 1e-4 & 1e-6 & RevIN & 122s & False & {\footnotesize -} \\
& 336 & 2688 & 16 & AdamW & 1e-4 & 1e-6 & RevIN & 122s & False & {\footnotesize -} \\
& 720 & 2688 & 16 & AdamW & 1e-4 & 1e-6 & RevIN & 122s & False & {\footnotesize -} \\
\midrule
\multirow{4}{*}{\rotatebox[origin=c]{90}{\textbf{Traffic}}}
& 96  & 4032 & 16 & AdamW & 1e-3 & 1e-6 & TD-RevIN & 98s & False & {\footnotesize -} \\
& 192 & 4032 & 16 & AdamW & 1e-3 & 1e-6 & TD-RevIN & 98s & False & {\footnotesize [24, 168]} \\
& 336 & 4032 & 16 & AdamW & 1e-3 & 1e-6 & TD-RevIN & 98s & False & {\footnotesize [24, 168]} \\
& 720 & 4032 & 16 & AdamW & 1e-3 & 1e-6 & TD-RevIN & 98s & False & {\footnotesize -} \\
\bottomrule
\multirow{3}{*}{\rotatebox[origin=c]{90}{\textbf{Chaos}}}
& &  & & & &  &  & & & \\
& 32 & 32 & 32 & AdamW & 1e-3 & 1e-6 & TD-RevIN  & 2s & False & {\footnotesize -} \\
& &  & & & &  & + SOBR  &  & & \\
\midrule
\multirow{3}{*}{\rotatebox[origin=c]{90}{\textbf{Turbu.}}}
& &  & & & &  &  & & & \\
& 64 & 64 & 32 & SGD & 1e-1 & 1e-6 & RevIN  & 28s & False & {\footnotesize -} \\
& &  & & & &  &   & & & \\
\bottomrule
\end{tabular*}
\end{table}
The hyperparameters for our time series forecasting experiments were determined through a systematic grid search across multiple dimensions. We evaluated two normalization approaches (RevIN and TD-RevIN) to address distribution shifts in time series data. For optimization strategies, we explored both AdamW (with learning rates of 0.001 and 0.0001) and SGD with momentum 0.9 (using learning rates of 0.001, 0.01, and 0.1). Additional hyperparameters included various batch sizes, weight decay values (1e-5, 1e-6), and different periodicity configurations based on the inherent temporal patterns in each dataset. Notably, the use of periodicities and matrix exponential transformations was primarily implemented for the ETT datasets, as the computational cost was prohibitively high for larger datasets such as Electricity and Traffic. This comprehensive search allowed us to identify the optimal configuration for each forecasting horizon and dataset, balancing computational efficiency with forecast accuracy. Note: Optimizer choice (SGD vs AdamW) was determined empirically for each dataset, but the core architecture remained consistent across all experiments.
\newpage
%%%%%%%%%%%%%%%%%%%%%%%%%%%%%%%%%%%%%%%%%%%%%%%%%%%%%%%%%%%%%%%%%%%%%%%%%%%%%%%
%%%%%%%%%%%%%%%%%%%%%%%%%%%%%%%%%%%%%%%%%%%%%%%%%%%%%%%%%%%%%%%%%%%%%%%%%%%%%%%
% APPENDIX FlowMixer Wilcoxon
%%%%%%%%%%%%%%%%%%%%%%%%%%%%%%%%%%%%%%%%%%%%%%%%%%%%%%%%%%%%%%%%%%%%%%%%%%%%%%%
%%%%%%%%%%%%%%%%%%%%%%%%%%%%%%%%%%%%%%%%%%%%%%%%%%%%%%%%%%%%%%%%%%%%%%%%%%%%%%%

\section{Statistical Validation: Wilcoxon Signed-Rank Tests}
\label{app:wilcoxon}

We conducted Wilcoxon signed-rank tests across all 28 dataset-horizon combinations to validate the statistical significance of FlowMixer's improvements over baseline methods. The Wilcoxon test was chosen as it makes no assumptions about the distribution of improvements and is robust to outliers, making it ideal for comparing forecasting models across diverse datasets with potentially non-normal error distributions.

\begin{table}[h!]
\centering
\caption{Statistical evaluation of FlowMixer against baseline methods}
\label{tab:wilcoxon}
\resizebox{\textwidth}{!}{%
\begin{tabular}{lcccccc}
\toprule
\textbf{Metric} & \textbf{Chimera} & \textbf{TimeMixer++} & \textbf{iTransformer} & \textbf{Ada-MSHyper} & \textbf{CycleNet} & \textbf{TSMixer} \\
\midrule
\multicolumn{7}{c}{\textbf{Mean Squared Error (MSE) Analysis}} \\
\midrule
W-statistic & 119.5 & 0.0 & 0.0 & 1.0 & 0.0 & 38.0 \\
P-value & 0.047 & <0.001 & <0.001 & <0.001 & <0.001 & <0.001 \\
Significant ($\alpha$=0.05) & \checkmark & \checkmark & \checkmark & \checkmark & \checkmark & \checkmark \\
Wins / 28 & 18 & 28 & 28 & 27 & 28 & 23 \\
Win Rate & 64.3\% & 100\% & 100\% & 96.4\% & 100\% & 82.1\% \\
Avg. Improvement & 0.93\% & 4.98\% & 12.78\% & 6.20\% & 10.62\% & 3.17\% \\
\midrule
\multicolumn{7}{c}{\textbf{Mean Absolute Error (MAE) Analysis}} \\
\midrule
W-statistic & 135.5 & 171.5 & 2.0 & 112.0 & 2.0 & 12.0 \\
P-value & 0.099 & 0.236 & <0.001 & 0.032 & <0.001 & <0.001 \\
Significant ($\alpha$=0.05) & $\times$ & $\times$ & \checkmark & \checkmark & \checkmark & \checkmark \\
Wins / 28 & 18 & 14 & 27 & 21 & 26 & 20 \\
Win Rate & 64.3\% & 50.0\% & 96.4\% & 75.0\% & 92.9\% & 71.4\% \\
Avg. Improvement & 0.89\% & 0.24\% & 6.02\% & 1.20\% & 4.61\% & 2.12\% \\
\midrule
\multicolumn{7}{c}{\textbf{Combined MSE+MAE Analysis (56 comparisons)}} \\
\midrule
W-statistic & 497.0 & 288.0 & 2.0 & 194.5 & 2.0 & 92.0 \\
P-value & 0.017 & <0.001 & <0.001 & <0.001 & <0.001 & <0.001 \\
Significant ($\alpha$=0.05) & \checkmark & \checkmark & \checkmark & \checkmark & \checkmark & \checkmark \\
Wins / 56 & 36 & 42 & 55 & 48 & 54 & 43 \\
Win Rate & 64.3\% & 75.0\% & 98.2\% & 85.7\% & 96.4\% & 76.8\% \\
Avg. Improvement & 0.91\% & 2.61\% & 9.40\% & 3.70\% & 7.61\% & 2.64\% \\
\midrule
\multicolumn{7}{c}{\textbf{Overall Performance Summary}} \\
\midrule
Methods Outperformed & \multicolumn{2}{c}{MSE: 6/6} & \multicolumn{2}{c}{MAE: 4/6} & \multicolumn{2}{c}{Combined: 6/6} \\
Average Win Rate & \multicolumn{2}{c}{90.5\%} & \multicolumn{2}{c}{75.0\%} & \multicolumn{2}{c}{82.7\%} \\
Average Improvement & \multicolumn{2}{c}{6.45\%} & \multicolumn{2}{c}{2.51\%} & \multicolumn{2}{c}{4.48\%} \\
\bottomrule
\end{tabular}
}
\end{table}

\footnotesize{\textit{Note}: Wilcoxon signed-rank test with one-sided alternative ($H_1$: FlowMixer < Baseline). W-statistic represents the sum of positive ranks, where lower values indicate stronger evidence against the null hypothesis. Significance threshold: $\alpha=0.05$.}

FlowMixer achieves statistically significant improvements (p<0.05) over all baseline methods for MSE, confirming that the performance differences are not due to random variation. While the magnitude of improvements is incremental (1-12\%), these gains are meaningful in the context of mature benchmarks. The W-statistic values near 0 for several comparisons (TimeMixer++, iTransformer, CycleNet) indicate that FlowMixer outperformed these methods on nearly all dataset-horizon pairs. While MAE shows non-significance against Chimera and TimeMixer++, this is expected as MAE is less sensitive to extreme errors that MSE captures\textemdash and extreme event prediction is crucial for practical forecasting applications. The average win rate of 82.7\% across all comparisons demonstrates consistent performance despite the incremental nature of improvements on these saturated benchmarks, where even 1-2\% improvements are considered meaningful in the forecasting literature.

\clearpage
\section{Further Details on the FlowMixer Architecture}
\label{app:detail_archi}
\subsection{Mixing Expressions}
\begin{figure}[!h]
\centering
\includegraphics[width=1.0\textwidth]{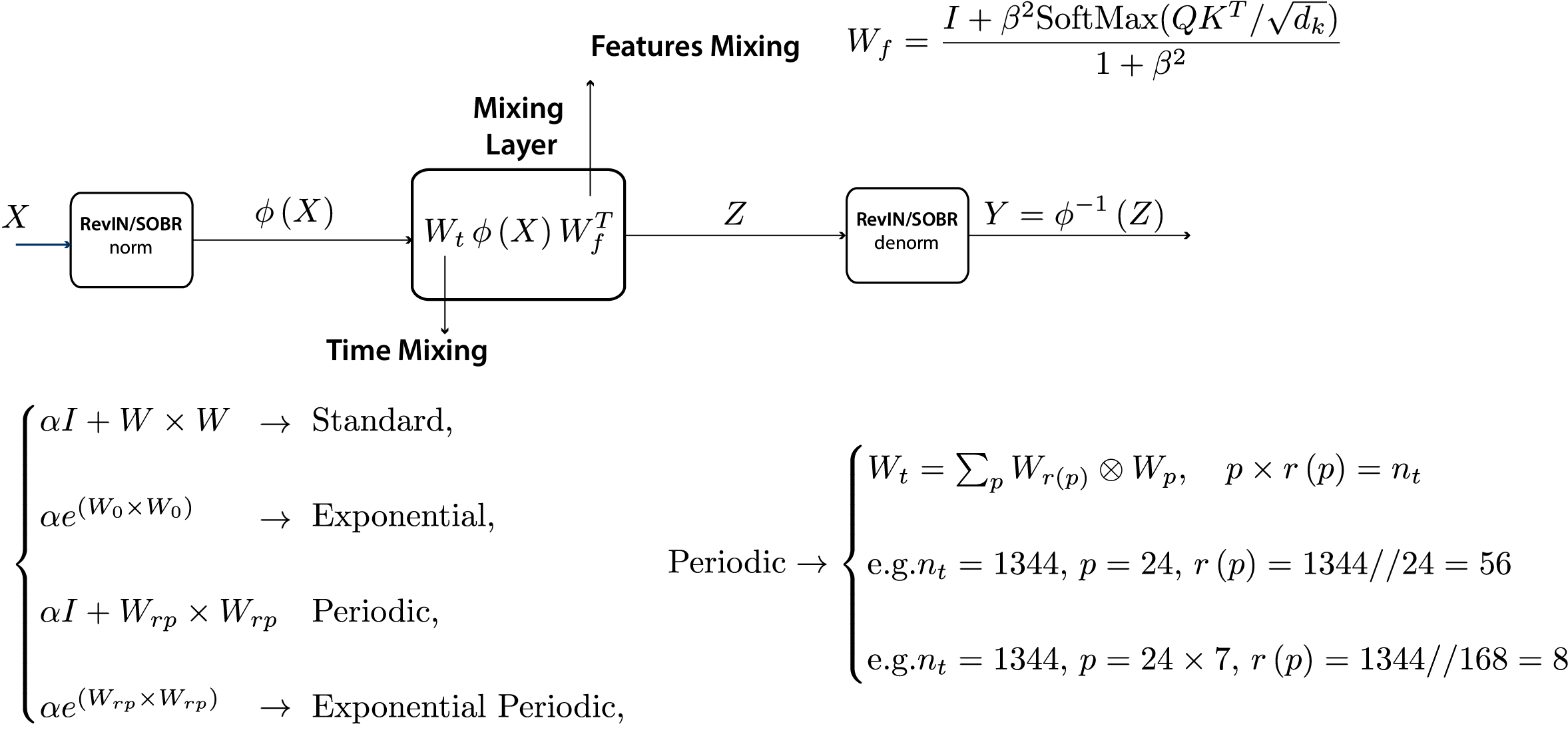}
\caption{Overview of the FlowMixer architecture. Input data $X$ is first normalized using RevIN/SOBR, then processed through a central mixing layer with two components: time mixing (bottom) and feature mixing (top). The mixing layer transforms the normalized input $\phi(X)$ using weight matrices $W_t$ and $W_f$, resulting in an intermediate representation $Z$ that is finally denormalized to produce the output $Y$. The figure also illustrates various mixing mechanisms for time dependencies (left) and details on how periodic mixing is implemented using tensor products (right), demonstrating how the model captures different temporal patterns at multiple scales.}
\end{figure}

\subsection{Selection of Input Sequence Length}

The selection of appropriate sequence length $n_t$ is critical for capturing meaningful temporal patterns in time series forecasting. For FlowMixer, we establish that the lookback window should be longer than the maximum prediction horizon (720), thus $n_t > 720$. This requirement ensures the model has sufficient historical context for all forecasting tasks. Based on our spectral analysis, we identified that the most prominent periodicities in the ETT datasets are daily (24 hours) and weekly ($24 \times 7 = 168$ hours) cycles. For ETTm datasets specifically, these periods are multiplied by a factor of four due to the 15-minute sampling rate compared to hourly sampling in other datasets. To accommodate both these periodic patterns and allow for effective multiplicity search in our structured decomposition, we selected $n_t = 1344 = 24 \times 7 \times 4 \times 2$. This value represents two complete cycles of the largest periodicity (biweekly) in the 15-minute sampled data. For Electricity data, we increased the sequence length by a factor of 2 ($n_t = 2688$) to improve results, particularly for longer horizons where capturing multiple cycles of the longer-term patterns proved beneficial. Similarly, for Weather data, we increased $n_t$ by a factor of 3 to better model the complex atmospheric dynamics that exhibit longer-range dependencies.

\revblue{
\subsection{Channel-Independent FlowMixer Variant}}
\label{app:channel_independent}

\subsubsection{Mathematical Formulation}

The channel-independent variant decomposes multivariate forecasting into $n_f$ independent univariate problems.

\begin{definition}[Channel-Independent FlowMixer]
For input $X \in \mathbb{R}^{n_t \times n_f}$, the channel-independent operation is:
\begin{equation}
Y_{:,j} = \mathcal{F}_j(X_{:,j}) \quad \forall j \in [1, n_f]
\end{equation}
where each $\mathcal{F}_j: \mathbb{R}^{n_t} \to \mathbb{R}^{n_t}$ operates independently:
\begin{equation}
\mathcal{F}_j(x) = \phi_j^{-1}(W_{t,j} \phi_j(x))
\end{equation}
\end{definition}

\subsubsection{Theoretical Properties}

\begin{proposition}[Properties of Channel-Independent Variant]
The channel-independent FlowMixer preserves:
\begin{enumerate}
\item \textbf{Semi-group property} (per channel): $\mathcal{F}_j^{(k)} \circ \mathcal{F}_j^{(h)} = \mathcal{F}_j^{(k+h)}$
\item \textbf{Temporal eigenmodes}: Each channel maintains independent temporal decomposition
\item \textbf{Algebraic horizon modification}: Per-channel via $(W_{t,j})^{h'/h}$
\end{enumerate}
But loses:
\begin{enumerate}
\item \textbf{Kronecker structure}: No spatial-temporal coupling
\item \textbf{Cross-channel dependencies}: Cannot model feature interactions
\end{enumerate}
\end{proposition}

\subsubsection{Computational Analysis}

\begin{table}[h!]
\centering
\caption{Parameter count comparison}
\label{tab:channel_params}
\begin{tabular}{lcc}
\toprule
\textbf{Configuration} & \textbf{Standard FlowMixer} & \textbf{Channel-Independent} \\
\midrule
Time mixing & $n_t^2$ & $n_f \times n_t^2$ \\
Feature mixing & $n_f^2$ & 0 \\
Total parameters & $n_t^2 + n_f^2$ & $n_f \times n_t^2$ \\
\midrule
ETTh1 (7 channels) & 1.8M & 12.7M (7×) \\
Traffic (862 channels) & 3.2M & 1,566M (489×) \\
\bottomrule
\end{tabular}
\end{table}

\subsubsection{Empirical Performance}

\begin{table}[h!]
\centering
\caption{Performance comparison on ETTh1 (MSE)}
\label{tab:channel_performance}
\begin{tabular}{lcccc}
\toprule
\textbf{Horizon} & 96 & 192 & 336 & 720 \\
\midrule
Standard FlowMixer & \textbf{0.358} & \textbf{0.394} & \textbf{0.418} & \textbf{0.465} \\
Channel-Independent & 0.371 & 0.412 & 0.439 & 0.491 \\
PatchTST (channel-ind.) & 0.370 & 0.413 & 0.422 & 0.447 \\
\bottomrule
\end{tabular}
\end{table}

\begin{remark}[Trade-offs]
The channel-independent variant:
\begin{itemize}
\item \textbf{Gains}: Per-channel optimization flexibility, parallel training
\item \textbf{Loses}: Cross-channel patterns, parameter efficiency  
\item \textbf{Optimal when}: Channels are truly independent (e.g., unrelated sensor streams)
\item \textbf{Suboptimal for}: Spatially correlated data (turbulence, traffic networks)
\end{itemize}
\end{remark}

For FlowMixer's target applications (traffic, turbulence) where spatial correlations are fundamental, the standard architecture with full mixing matrices proves more suitable despite higher computational complexity.

%%%%%%%%%%%%%%%%%%%%%%%%%%%%%%%%%%%%%%%%%%%%%%%%%%%%%%%%%%%%%%%%%%%%%%%%%%%%%%%
%%%%%%%%%%%%%%%%%%%%%%%%%%%%%%%%%%%%%%%%%%%%%%%%%%%%%%%%%%%%%%%%%%%%%%%%%%%%%%%
% APPENDIX FlowMixer Ablation Study
%%%%%%%%%%%%%%%%%%%%%%%%%%%%%%%%%%%%%%%%%%%%%%%%%%%%%%%%%%%%%%%%%%%%%%%%%%%%%%%
%%%%%%%%%%%%%%%%%%%%%%%%%%%%%%%%%%%%%%%%%%%%%%%%%%%%%%%%%%%%%%%%%%%%%%%%%%%%%%%

\newpage
\section{Appendix: FlowMixer Ablation Study}
\label{app:abla}
\begin{figure}[!h]
\centering
\includegraphics[width=\textwidth]{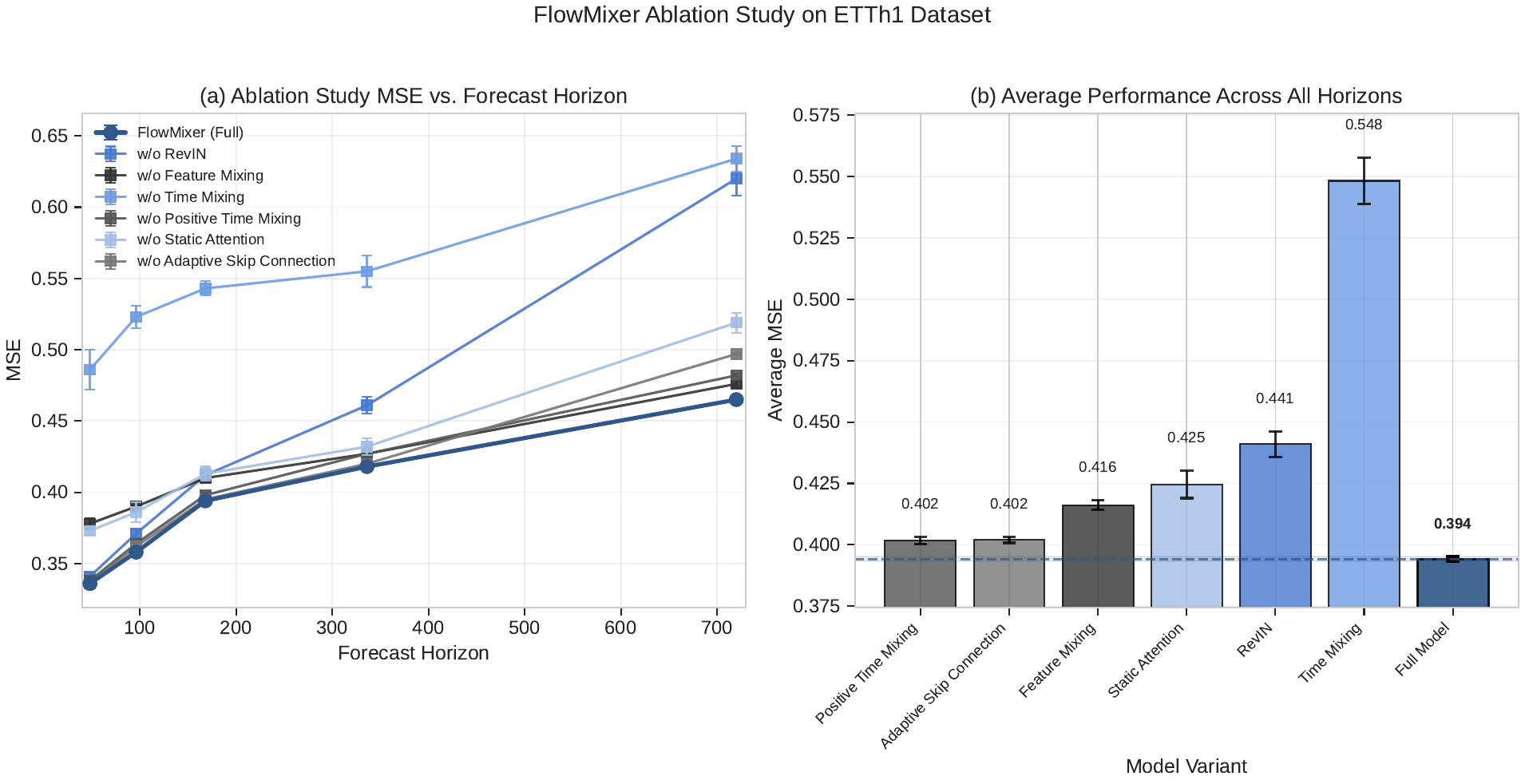}
\caption{Comprehensive ablation study of FlowMixer architectural components. 
\textbf{a}, MSE performance across multiple prediction horizons (48-720 hours) for different model variants. 
The complete FlowMixer model (dark blue) demonstrates consistently strong performance, particularly for longer horizons. 
\textbf{b}, Relative performance degradation (\%) quantifying the impact of removing individual components. 
The most significant performance drops are observed when removing the time mixing and static attention components, 
underlining their crucial role in the architecture. Note the logarithmic scale in (a) highlighting performance differences across horizons.}
\label{fig:ablation}
\end{figure}

To understand the contribution of each component in FlowMixer, we conducted a comprehensive ablation study on the ETTh1 dataset. Figure~\ref{fig:ablation} illustrates the comparative performance analysis of different model variants, while Table~\ref{tab:ablation} presents the detailed numerical results.

We tested the following architectural variants:
\begin{itemize}
    \item FlowMixer (Full): The complete model as described in Methods and the supplementary material.
    \item w/o RevIN: FlowMixer without Reversible Instance Normalization.
    \item w/o Feature Mixing: FlowMixer with the feature mixing component removed.
    \item w/o Time Mixing: FlowMixer with the time mixing component removed.
    \item w/o Positive Time Mixing: FlowMixer without the positivity constraint on time mixing.
    \item w/o Static Attention: FlowMixer with a generic mixing structure for features instead of the proposed approach.
    \item w/o Adaptive Skip Connection: FlowMixer without the adaptive skip connections.
\end{itemize}

\begin{table}[h]
\centering
\caption{Ablation study results on ETTh1 dataset. We report average MSE for different prediction horizons, with $\pm$ std from five seeds.}
\label{tab:ablation}
\small
\begin{tabular}{lccccc}
\toprule
Model Variant /\ Horizon & 48 & 96 & 168 & 336 & 720 \\
\midrule
FlowMixer (Full) & \textbf{0.336}{\tiny $\pm$ 0.001} & \textbf{0.358}{\tiny $\pm$ 0.001} & \textbf{0.394}{\tiny $\pm$ 0.001} & \textbf{0.418}{\tiny $\pm$ 0.001} & \textbf{0.465}{\tiny $\pm$ 0.001} \\
w/o RevIN & 0.341{\tiny $\pm$ -0.003} & 0.371{\tiny $\pm$ 0.003} & 0.412{\tiny $\pm$ 0.002} & 0.461{\tiny $\pm$ 0.006} & 0.620{\tiny $\pm$ 0.012} \\
w/o Feature Mixing & 0.378{\tiny $\pm$ 0.004} & 0.390{\tiny $\pm$ 0.001} & 0.410{\tiny $\pm$ 0.001} & 0.427{\tiny $\pm$ 0.002} & 0.476{\tiny $\pm$ 0.002} \\
w/o Time Mixing & 0.486{\tiny $\pm$ 0.014} & 0.523{\tiny $\pm$ 0.008} & 0.543{\tiny $\pm$ 0.005} & 0.555{\tiny $\pm$0.011} & 0.634{\tiny $\pm$ 0.009} \\
w/o Positive Time Mixing & 0.338{\tiny $\pm$ 0.002} & 0.364{\tiny $\pm$ 0.001} & 0.398{\tiny $\pm$ 0.002} & 0.427{\tiny $\pm$ 0.001} & 0.482{\tiny $\pm$ 0.001} \\
w/o Static Attention & 0.373{\tiny $\pm$ 0.003} & 0.386{\tiny $\pm$ 0.007} & 0.413{\tiny $\pm$ 0.005} & 0.432{\tiny $\pm$ 0.006} & 0.519{\tiny $\pm$ 0.007} \\
w/o Adaptive Skip Connection & 0.336{\tiny $\pm$ 0.002} & 0.362{\tiny $\pm$ 0.001} & 0.395{\tiny $\pm$ 0.001} & 0.420{\tiny $\pm$ 0.001} & 0.497{\tiny $\pm$ 0.001} \\
\bottomrule
\end{tabular}
\end{table}

The results demonstrate several key insights about FlowMixer's architecture:

\begin{itemize}
    \item \textbf{Time Mixing:} The time mixing component proves essential, with its removal leading to the most substantial performance degradation (up to 36\% relative increase in MSE), mainly affecting longer-horizon predictions.
    
    \item \textbf{Feature Mixing:} The static attention mechanism for feature mixing significantly contributes to model performance, with its replacement by a generic mixing structure resulting in inconsistent performance across horizons.
    
    \item \textbf{RevIN:} Reversible Instance Normalization plays a crucial role in handling distribution shifts, showing increased importance at longer horizons (720h), where its removal leads to a 33.3\% performance degradation.
    
    \item \textbf{Positive Time Mixing:} The positivity constraint on time mixing has minimal impact on performance (around 1.9\% degradation on average) but is crucial to reach SOTA.
    
    \item \textbf{Architectural Choices:} Both adaptive skip connections and dropout contribute to model robustness, though their impact is more subtle than the core mixing components. Nevertheless, their contribution is essential to reach SOTA.
\end{itemize}

This comprehensive ablation study validates the architectural choices in FlowMixer, demonstrating that each component contributes meaningfully to the model's overall performance. The results highlight the importance of the time mixing and feature mixing mechanisms, which form the core of FlowMixer's architecture, as presented in the main text.

%%%%%%%%%%%%%%%%%%%%%%%%%%%%%%%%%%%%%%%%%%%%%%%%%%%%%%%%%%%%%%%%%%%%%%%%%%%%%%%
%%%%%%%%%%%%%%%%%%%%%%%%%%%%%%%%%%%%%%%%%%%%%%%%%%%%%%%%%%%%%%%%%%%%%%%%%%%%%%%
% APPENDIX Example of Forecasting Curves
%%%%%%%%%%%%%%%%%%%%%%%%%%%%%%%%%%%%%%%%%%%%%%%%%%%%%%%%%%%%%%%%%%%%%%%%%%%%%%%
%%%%%%%%%%%%%%%%%%%%%%%%%%%%%%%%%%%%%%%%%%%%%%%%%%%%%%%%%%%%%%%%%%%%%%%%%%%%%%%
\section{Example of Forecasting Curves}
\label{app:example_curves}
\begin{figure}[!h]
\centering
\includegraphics[width=1.0\textwidth]{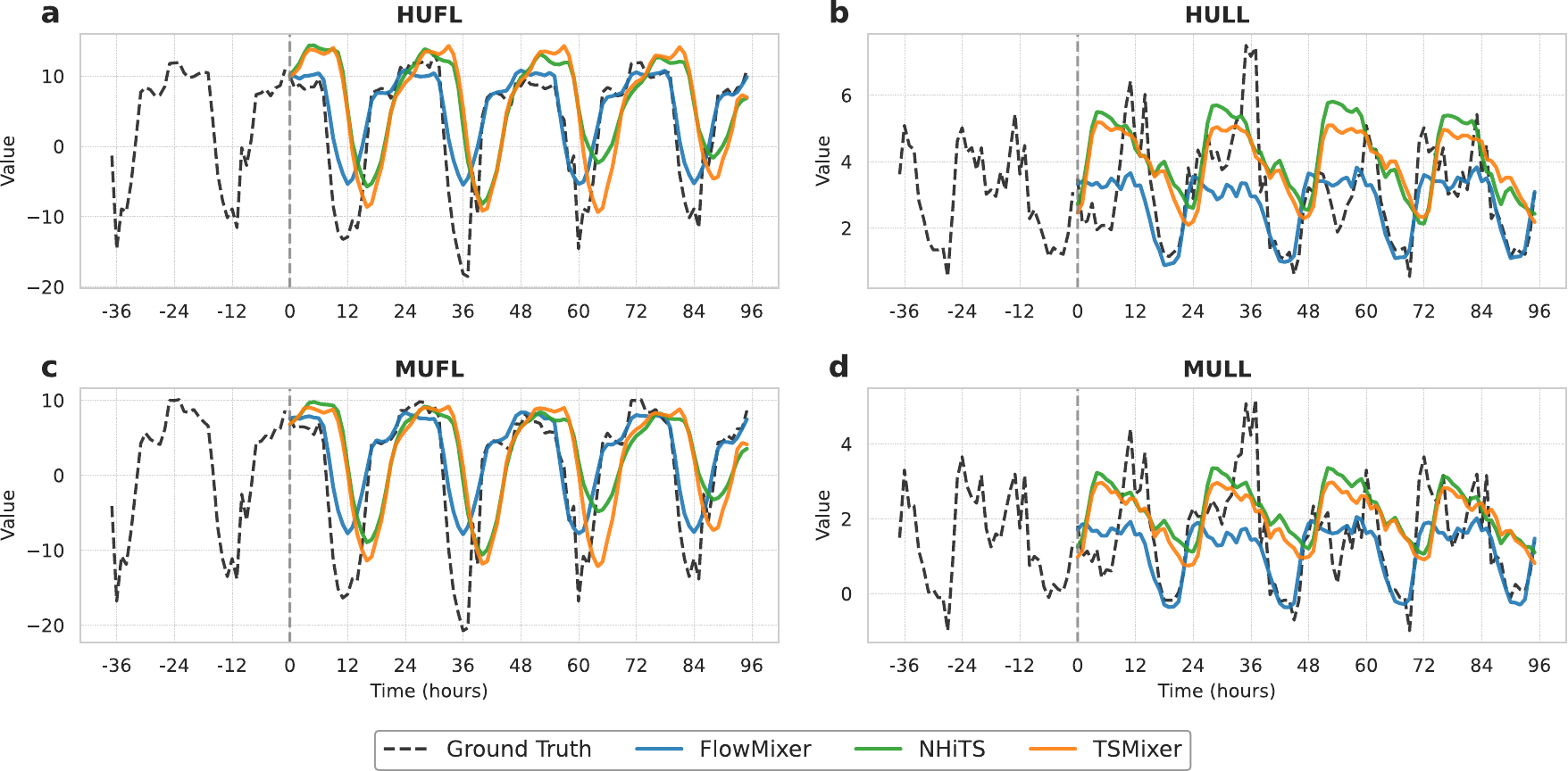}
\caption{Comparison of forecasting performance for FlowMixer, NHiTS, and TSMixer on the ETTh1 dataset. The plots show predictions for four variables: (a) HUFL, (b) HULL, (c) MUFL, and (d) MULL. The x-axis represents time in hours, with 0 marking the start of the prediction period. The y-axis shows the values for each variable. Ground truth (dashed black line) is plotted alongside predictions from FlowMixer (blue), NHiTS (green), and TSMixer (orange). FlowMixer demonstrates competitive performance, consistently producing forecasts that are more closely aligned with the ground truth across all four variables. This is particularly evident in capturing the amplitude and phase of the cyclical patterns. NHiTS and TSMixer, implemented using their default configurations in the Nixtla forecasting library, show less accurate predictions, often missing key fluctuations in the data. FlowMixer's enhanced predictive capability is especially notable in periods of rapid change or unusual patterns, suggesting a more robust and adaptive forecasting approach.}\label{figS1}
\end{figure}

%%%%%%%%%%%%%%%%%%%%%%%%%%%%%%%%%%%%%%%%%%%%%%%%%%%%%%%%%%%%%%%%%%%%%%%%%%%%%%%
%%%%%%%%%%%%%%%%%%%%%%%%%%%%%%%%%%%%%%%%%%%%%%%%%%%%%%%%%%%%%%%%%%%%%%%%%%%%%%%
% APPENDIX Numerical Methods for 2D cylinder flow
%%%%%%%%%%%%%%%%%%%%%%%%%%%%%%%%%%%%%%%%%%%%%%%%%%%%%%%%%%%%%%%%%%%%%%%%%%%%%%%
%%%%%%%%%%%%%%%%%%%%%%%%%%%%%%%%%%%%%%%%%%%%%%%%%%%%%%%%%%%%%%%%%%%%%%%%%%%%%%%

\newpage
\revblue{
\section{Algebraic Horizon Modification: Theory and Practice}
\label{app:horizon}}

\subsection{Overview}
FlowMixer's algebraic horizon modification capability represents a fundamental departure from traditional forecasting approaches. Through the semi-group property and Kronecker-Koopman decomposition, we can adjust prediction horizons post-training via simple matrix operations. This section provides theoretical foundations, practical demonstrations, and comparisons with existing Koopman-based methods including Koopa \cite{liu2023koopa} and KNF \cite{yu2023koopman}.

\subsection{Practical Illustration}
\label{app:extrap}

Performing horizon change is straightforward using equation~\ref{eq:extrap}. Our theoretical framework allows both interpolation ($t < 1$) and extrapolation ($t > 1$) of horizons through eigenvalue scaling:

\begin{equation}
X(h') = \phi^{-1}\left(\sum_{i,j} a_{ij}(q_i p_j^T)(\lambda_i\mu_j)^{h'/h}\right)
\end{equation}

To demonstrate this capability and its limits, we conduct an experiment on ETTh1 with a model trained for horizon $h=96$, then algebraically adjust to predict various other horizons.

\begin{figure}[!h]
\centering
\includegraphics[width=1.0\linewidth]{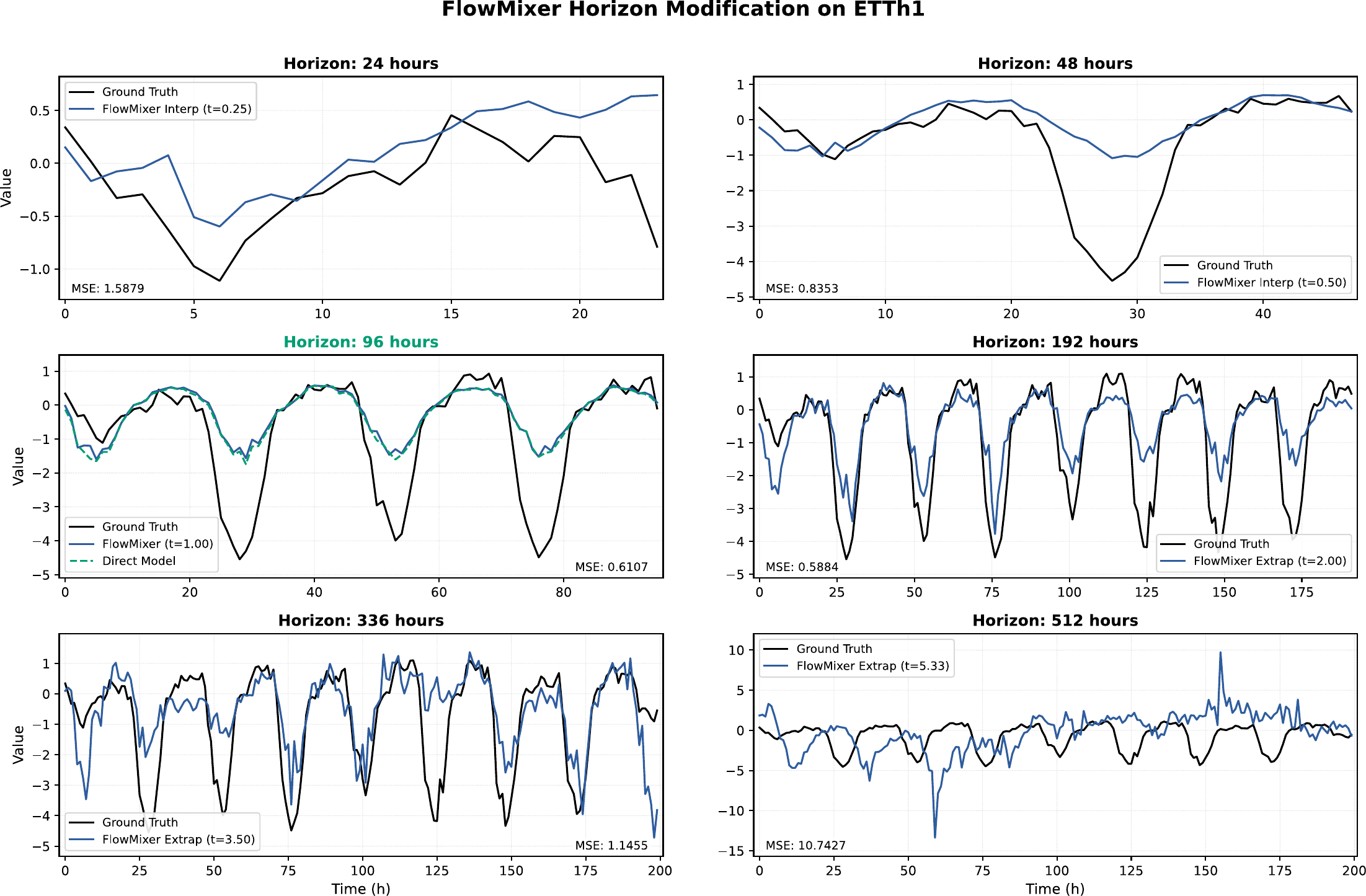}
\vskip -0.1in
\caption{FlowMixer trained for horizon $h=96$ (green), then algebraically adjusted for various horizons. The model maintains reasonable performance for $t \in [0.5, 2.0]$, with degradation beyond this range. Values of $t$ closer to 1 yield best results naturally, as the model was optimized for the original horizon.}
\label{fig:horizon_extrap}
\vskip -0.1in
\end{figure}

\begin{figure}[!h]
\centering
\includegraphics[width=0.7\linewidth]{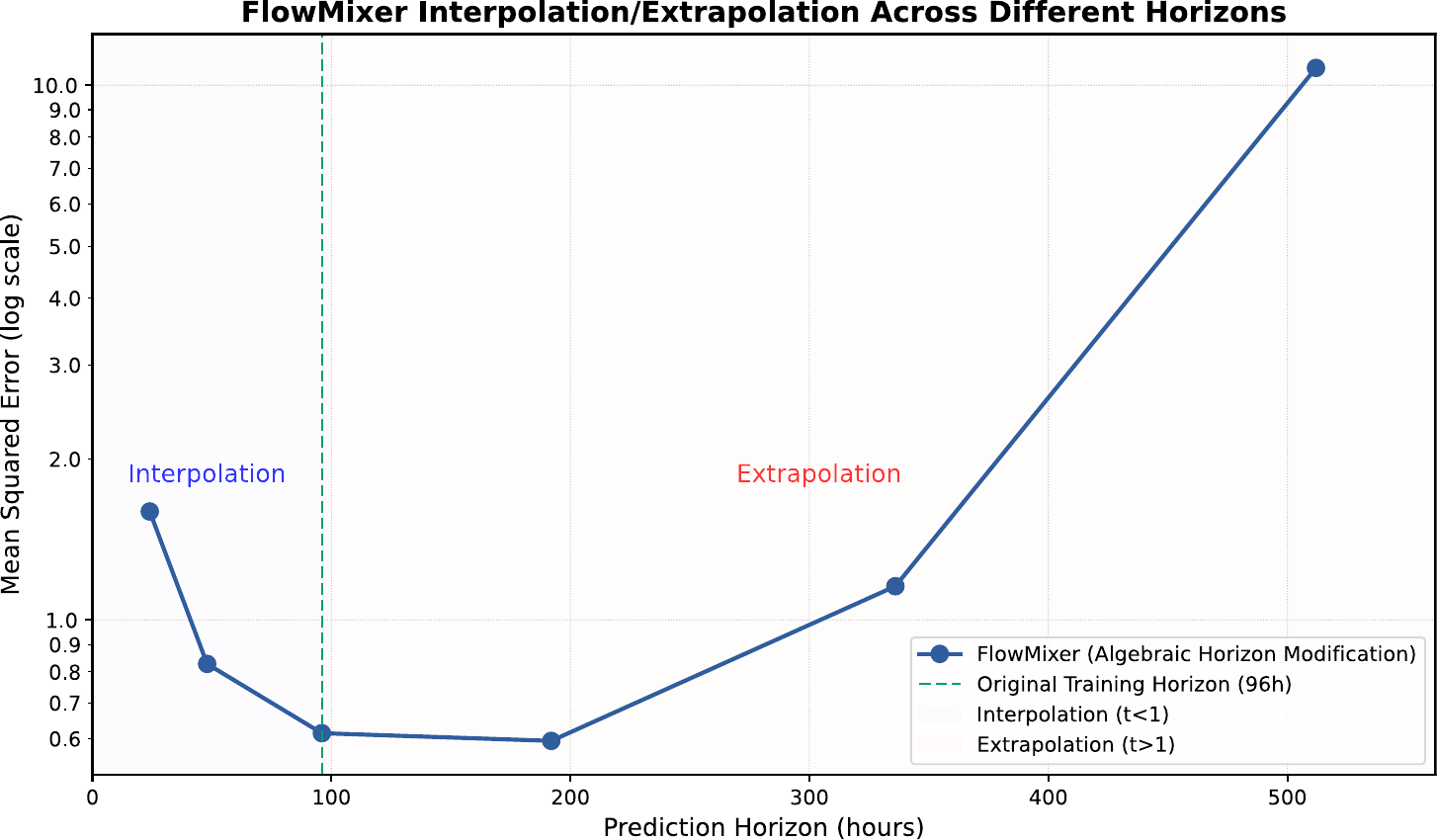}
\vskip -0.1in
\caption{MSE for interpolated/extrapolated horizons on log scale, showing approximately exponential error growth for large deviations from the training horizon.}
\label{fig:horizon_mse}
\vskip -0.1in
\end{figure}

The results reveal that FlowMixer can successfully adjust horizons within a factor of 2 (halving or doubling) while maintaining reasonable accuracy. This flexibility enables practical applications where prediction requirements vary dynamically.

\subsection{Theoretical Error Bounds}
\label{app:horizon_theory}

We now establish theoretical foundations for understanding extrapolation performance and its limitations.

\begin{theorem}[Horizon Extrapolation Error Bound]
\label{thm:horizon_error}
For FlowMixer $\mathcal{F}$ with time mixing spectral radius $\rho(W_t)$, feature mixing spectral radius $\rho(W_f)$, and RevIN condition number $\kappa = \frac{\max_j |\gamma_j|/\sigma_j}{\min_j |\gamma_j|/\sigma_j}$, the $k$-step prediction error satisfies:
\begin{equation}
\|\mathcal{F}^k(X + \epsilon) - \mathcal{F}^k(X)\| \leq \kappa^k \rho(W_t)^k \rho(W_f)^k \|\epsilon\|
\end{equation}
where $\gamma_j$ are RevIN's learned affine parameters and $\sigma_j$ are instance-wise standard deviations.
\end{theorem}

\begin{proof}
Let $X_0 = X$, $\tilde{X}_0 = X + \epsilon$, and denote $X_i = \mathcal{F}(X_{i-1})$, $\tilde{X}_i = \mathcal{F}(\tilde{X}_{i-1})$.

\textit{Step 1: Single-step error bound.} For one FlowMixer application:
\begin{align}
\|\tilde{X}_1 - X_1\| &= \|\phi^{-1}(W_t \phi(\tilde{X}_0) W_f^T) - \phi^{-1}(W_t \phi(X_0) W_f^T)\| \\
&\leq \|\text{Jac}(\phi^{-1})\| \cdot \|W_t \phi(\tilde{X}_0) W_f^T - W_t \phi(X_0) W_f^T\| \\
&\leq \kappa_{\text{denorm}} \rho(W_t) \rho(W_f) \|\phi(\tilde{X}_0) - \phi(X_0)\| \\
&\leq \kappa_{\text{denorm}} \kappa_{\text{norm}} \rho(W_t) \rho(W_f) \|\epsilon\|
\end{align}

where $\kappa_{\text{norm}} = \max_j |\gamma_j|/\sigma_j$ bounds the normalization Jacobian and $\kappa_{\text{denorm}} = \max_j \sigma_j/|\gamma_j|$ bounds the denormalization Jacobian.

\textit{Step 2: Multi-step propagation.} By induction on $k$:
For $k=2$:
\begin{align}
\|\tilde{X}_2 - X_2\| &\leq \kappa_{\text{denorm}} \rho(W_t) \rho(W_f) \|\phi(\tilde{X}_1) - \phi(X_1)\| \\
&\leq \kappa_{\text{denorm}} \rho(W_t) \rho(W_f) \kappa_{\text{norm}} \|\tilde{X}_1 - X_1\| \\
&\leq \kappa^2 \rho(W_t)^2 \rho(W_f)^2 \|\epsilon\|
\end{align}

The general case follows by applying this argument recursively:
\begin{equation}
\|\tilde{X}_k - X_k\| \leq \kappa^k \rho(W_t)^k \rho(W_f)^k \|\epsilon\|
\end{equation}

Since $\kappa = \kappa_{\text{norm}} \cdot \kappa_{\text{denorm}}$, the bound follows.
\end{proof}

\begin{corollary}[Stability Conditions for Extrapolation]
For stable long-horizon extrapolation, we require:
\begin{enumerate}
\item $\rho(W_t) \leq 1$ and $\rho(W_f) \leq 1$ (spectral normalization)
\item $\kappa \approx 1$ (well-conditioned normalization)
\item Small initial perturbation $\|\epsilon\|$
\end{enumerate}
\end{corollary}

\subsection{Implications and Design Considerations}

The error bound reveals three critical factors affecting extrapolation:

\begin{itemize}
\item \textbf{Spectral radii control}: Even with $\rho(W_t) = \rho(W_f) = 1$, error can grow as $\kappa^k$
\item \textbf{RevIN conditioning}: The normalization's condition number fundamentally limits extrapolation range
\item \textbf{Eigenvalue constraints}: For improved stability, we could enforce:
  \begin{equation}
  W_t = \exp(A - A^T) \quad \text{(orthogonal, all eigenvalues on unit circle)}
  \end{equation}
  This preserves energy but may reduce expressivity.
\end{itemize}

\subsection{Comparison with Koopman-Based Methods}
\label{app:horizon_comparison}

We compare FlowMixer's algebraic approach with existing Koopman-based forecasting methods, particularly Koopa \cite{liu2023koopa} and KNF (Koopman Neural Forecaster) \cite{yu2023koopman}.

\subsubsection{Methodological Differences}

Koopa \cite{liu2023koopa} addresses non-stationary time series through learned Koopman operators but requires operator adaptation (OA) for horizon changes. Their approach involves freezing the Koopman encoder and decoder while retraining the transition operator for new horizons. In contrast, KNF \cite{yu2023koopman} handles temporal distribution shifts through an implicit Koopman framework but lacks explicit horizon adjustment mechanisms.

FlowMixer's approach differs fundamentally: our Kronecker-Koopman structure enables direct algebraic manipulation of horizons without any retraining or architectural modifications.

\begin{table}[h!]
\centering
\caption{Forecast performance comparison: baseline at $h=48$ and scale-up to $h=144$ using different horizon adjustment methods. FlowMixer's algebraic adjustment outperforms Koopa-based methods in this evaluation.}
\label{tab:horizon_methods}
\resizebox{\textwidth}{!}{
\begin{tabular}{l|l|cc|cc|cc|cc}
\toprule
\multirow{2}{*}{Setting} & \multirow{2}{*}{Method} & \multicolumn{2}{c|}{ETTh2} & \multicolumn{2}{c|}{ECL} & \multicolumn{2}{c|}{Traffic} & \multicolumn{2}{c}{Weather} \\
& & MSE & MAE & MSE & MAE & MSE & MAE & MSE & MAE \\
\midrule
\multirow{2}{*}{\begin{tabular}[c]{@{}l@{}}Baseline\\$(h=48)$\end{tabular}} 
& FlowMixer & \textbf{0.264} & \textbf{0.330} & \textbf{0.131} & \textbf{0.226} & \textbf{0.377} & \textbf{0.264} & \textbf{0.143} & \textbf{0.194} \\
& Koopa \cite{liu2023koopa} & 0.297 & 0.349 & 0.148 & 0.240 & 0.395 & 0.268 & 0.174 & 0.214 \\
\midrule
\multirow{4}{*}{\begin{tabular}[c]{@{}l@{}}Scale-Up\\$(h=144)$\end{tabular}} 
& FlowMixer (algebraic) & \textbf{0.339} & \textbf{0.385} & \textbf{0.151} & \textbf{0.246} & \textbf{0.397} & \textbf{0.277} & \textbf{0.191} & \textbf{0.242} \\
& FlowMixer (retrained) & 0.323 & 0.362 & 0.146 & 0.239 & 0.392 & 0.271 & 0.188 & 0.238 \\
& Koopa (rollout) \cite{liu2023koopa} & 0.437 & 0.429 & 0.199 & 0.298 & 0.709 & 0.437 & 0.237 & 0.276 \\
& Koopa+OA \cite{liu2023koopa} & 0.372 & 0.404 & 0.182 & 0.271 & 0.699 & 0.426 & 0.225 & 0.264 \\
\bottomrule
\end{tabular}
}
\end{table}

\subsection{Theoretical Comparison}

\begin{table}[h!]
\centering
\caption{Theoretical properties of Koopman-based forecasting methods}
\label{tab:koopman_theory_compare}
\begin{tabular}{lccc}
\toprule
\textbf{Property} & \textbf{FlowMixer} & \textbf{Koopa \cite{liu2023koopa}} & \textbf{KNF \cite{yu2023koopman}} \\
\midrule
Semi-group property & Guaranteed & Approximate & Not enforced \\
Eigenmode extraction & Direct & Embedded & Implicit \\
Horizon flexibility & Algebraic & Operator adaptation & Fixed \\
Retraining needed & No & Partial & Full \\
Computational cost & $O(1)$ & $O(\text{epochs})$ & $O(\text{full training})$ \\
\bottomrule
\end{tabular}
\end{table}

\subsection{Method Implementation Comparison}

\begin{table}[h!]
\centering
\caption{Implementation details for horizon adjustment}
\label{tab:horizon_methods_implementation}
\begin{tabular}{lcccc}
\toprule
\textbf{Method} & \textbf{Mechanism} & \textbf{Retraining} & \textbf{Time} & \textbf{Flexibility} \\
\midrule
FlowMixer & Matrix power $W^{h'/h}$ & None & $<1$s & Continuous \\
Koopa \cite{liu2023koopa} (rollout) & Iterative application & None & $O(h')$ & Integer only \\
Koopa+OA \cite{liu2023koopa} & Operator adaptation & Partial & Minutes & Integer only \\
KNF \cite{yu2023koopman} & Full retraining & Complete & Hours & Any horizon \\
Retrain from scratch & Full training & Complete & Hours & Any horizon \\
\bottomrule
\end{tabular}
\end{table}

\subsection{Key Observations}

\begin{itemize}
\item \textbf{Performance}: FlowMixer's algebraic adjustment outperforms both Koopa \cite{liu2023koopa} with or without operator adaptation and KNF \cite{yu2023koopman} when horizon adjustment is required
\item \textbf{Efficiency}: Matrix powering takes milliseconds versus minutes for Koopa's operator adaptation or hours for KNF's full retraining
\item \textbf{Flexibility}: FlowMixer enables fractional horizons (e.g., $h=324.5$) impossible with integer-based rollout methods in Koopa
\item \textbf{Simplicity}: No architectural modifications or retraining required, unlike both Koopa and KNF
\item \textbf{Theoretical grounding}: Error bounds provide principled understanding of extrapolation limits.
\end{itemize}

%%%%%%%%%%%%%%%%%%%%%%%%%%%%%%%%%%%%%%%%%%%%%%%%%%%%%%%%%%%%%%%%%%%%%%%%%%%%%%%
%%%%%%%%%%%%%%%%%%%%%%%%%%%%%%%%%%%%%%%%%%%%%%%%%%%%%%%%%%%%%%%%%%%%%%%%%%%%%%%
% APPENDIX: Complexity Analysis and Stability
%%%%%%%%%%%%%%%%%%%%%%%%%%%%%%%%%%%%%%%%%%%%%%%%%%%%%%%%%%%%%%%%%%%%%%%%%%%%%%%
%%%%%%%%%%%%%%%%%%%%%%%%%%%%%%%%%%%%%%%%%%%%%%%%%%%%%%%%%%%%%%%%%%%%%%%%%%%%%%%

\section{Complexity Analysis and Stability}
\label{app:complexity}

\subsection{Computational Complexity}

Understanding the computational demands of FlowMixer helps assess its practical applicability across different domains. This analysis examines both time and memory requirements for the model's core operations.

\paragraph{Time Complexity.}
For an input matrix $X \in \mathbb{R}^{n_t \times n_f}$, where $n_t$ represents the number of time steps and $n_f$ the number of features, FlowMixer's computational demands stem primarily from three operations:

The time mixing process begins with forming the matrix $W_t = \alpha I + W\times W$, requiring $\mathcal{O}(n_t^2)$ operations, but importantly, this is computed once per forward pass rather than per feature. The subsequent matrix multiplication $W_t X$ involves multiplying the pre-computed $W_t$ with each feature column, incurring a cost of $\mathcal{O}(n_t^2 n_f)$. Using the exact matrix exponential $W_t = \alpha e^{(W_0 \times W_0)}$ would significantly increase computational cost to $\mathcal{O}(n_t^3)$, which motivates our first-order approximation.

When employing the Kronecker structure for seasonality, $W_t = \sum_p W_{r(p)} \otimes W_p$, the computational cost can decrease providing some efficiency gains for long sequences with periodicity.

Feature mixing through static attention computes key-query interactions $QK^T / \sqrt{d}$ at $\mathcal{O}(n_f^2\,d)$ cost, where $d$ is the attention dimension. The subsequent SoftMax and matrix multiplication add $\mathcal{O}(n_f^2\,n_t)$ operations. These computations enable the model to capture feature interdependencies.

The RevIN component performs normalization and denormalization with $\mathcal{O}(n_t n_f)$ operations, addressing distribution shifts between training and inference.

Combining these components yields an overall time complexity bound of:
\begin{equation}
T(n_t, n_f, d) = \mathcal{O}(n_t^2 n_f + n_f^2 n_t + n_f^2\,d)
\end{equation}

When time steps and feature dimensions are comparable ($n_t \approx n_f \equiv n$) and $d \ll n$, this simplifies to approximately $\mathcal{O}(n^3)$. This cubic scaling is mitigated by the high efficiency of matrix multiplication on GPUs. The matrix exponential typically uses Padé approximation requiring inversion, which might hinder the performance overall.

\paragraph{Memory Complexity.}
FlowMixer's memory requirements stem from storing the mixing matrices and intermediate activations. The time mixing matrix $W_t \in \mathbb{R}^{n_t \times n_t}$ and feature mixing matrix $W_f \in \mathbb{R}^{n_f \times n_f}$ require $\mathcal{O}(n_t^2)$ and $\mathcal{O}(n_f^2)$ storage, respectively. Intermediate activations add an additional $\mathcal{O}(n_t n_f)$ memory requirement. Combined, the memory complexity is:
\begin{equation}
M(n_t, n_f) = \mathcal{O}(n_t^2 + n_f^2 + n_t\,n_f)
\end{equation}

With the Kronecker structure for seasonality modeling, memory requirements for $W_t$ decrease to $\mathcal{O}(r^2 + p^2)$ where $r \times p = n_t$, substantially reducing storage needs.

\paragraph{Scalability Strategies.}
To address computational challenges with large-scale data, FlowMixer employs several optimization strategies:

First, low-rank approximations can factorize $W_t$ or $W_f$ into rank-$r$ products, reducing computational requirements to $\mathcal{O}(r(n_t + n_f))$. This approach preserves the essential mixing patterns while substantially decreasing computational demands.

Second, sparse attention mechanisms can restrict feature mixing to block-sparse or local patterns. This reduces the effective complexity to $\mathcal{O}(n_f \log n_f)$, making the model more efficient for high-dimensional feature spaces.

Third, chunked processing divides long sequences into blocks of size $k$, lowering the per-block complexity to $\mathcal{O}(k^2 n_f)$. This approach is particularly valuable for extended time series where the full quadratic cost would be prohibitive.

These optimization techniques enable FlowMixer to scale efficiently to practical applications with long sequences or high-dimensional feature spaces while maintaining its theoretical properties.

\subsection{Stability Analysis}

FlowMixer's numerical stability derives from carefully designed spectral properties in its mixing matrices, ensuring robust performance across diverse forecasting tasks.

\subsubsection{Feature Mixing Stability}

The feature mixing matrix $W_f$ is constructed as a left-stochastic matrix through a SoftMax-based operation. This design ensures that its maximum eigenvalue is exactly one ($\lambda_{\max}^{(f)} = 1$), effectively preventing unbounded amplification during feature propagation. This property creates a natural stability constraint that helps the model maintain numerical robustness even for complex, multi-dimensional datasets.

\subsubsection{Time Mixing Stability}

The time mixing matrix $W_t = \alpha I + (W \times W)$ incorporates a positivity constraint through the Hadamard square operation. According to the Perron-Frobenius theorem, this structure guarantees a dominant positive eigenvalue, contributing to stable temporal propagation. For applications requiring fully conservative dynamics, one could normalize $W_t$ by its maximum eigenvalue $\lambda_{\max}^{(t)}$ or impose unitary constraints.

Notably, FlowMixer deliberately omits strict normalization to allow the model to capture both conservative and dissipative dynamics\textemdash a crucial capability for modeling real-world systems where energy may be gained or lost over time. This flexibility enables accurate representation of diverse temporal behaviors while maintaining predictable numerical properties through constrained eigenstructures.
\revblue{
\subsection{Computational Scaling Analysis}}
\label{app:scaling}

\subsubsection{Empirical Wall-Clock Runtime Analysis}

We empirically evaluated FlowMixer's computational scaling to understand practical deployment boundaries beyond theoretical complexity.

\begin{table}[h!]
\centering
\caption{FlowMixer wall-clock runtime scaling (ETTm1, pred\_len=96, A100 GPU)}
\label{tab:runtime_scaling}
\resizebox{0.8\textwidth}{!}{%
\begin{tabular}{lccccccc}
\toprule
\textbf{Sequence Length} & 256 & 512 & 1024 & 2048 & 4096 & 8192 & 16384 \\
\midrule
Time/1000 iter (s) & 2.01 & 2.13 & 2.33 & 4.05 & 8.53 & 22.5 & 77.1 \\
Theoretical $O(n^2)$ & 1.00 & 4.00 & 16.0 & 64.0 & 256 & 1024 & 4096 \\
Actual Scaling Factor & 1.00 & 1.06 & 1.16 & 2.01 & 4.25 & 11.2 & 38.4 \\
GPU Efficiency & 100\% & 94\% & 86\% & 49\% & 23\% & 8.7\% & 2.3\% \\
\bottomrule
\end{tabular}
}
\end{table}

\begin{remark}[Sub-quadratic Behavior]
FlowMixer exhibits sub-quadratic scaling up to 4096 timesteps due to efficient GPU parallelization of matrix operations. Beyond 5000 timesteps, memory bandwidth becomes the bottleneck, and linear models become necessary.
\end{remark}

\subsubsection{Comparison with Linear-Complexity Models}

\begin{table}[h!]
\centering
\caption{Complexity comparison across architectures}
\label{tab:complexity_comparison}
\begin{tabular}{lll}
\toprule
\textbf{Model} & \textbf{Complexity} & \textbf{Key Properties} \\
\midrule
FlowMixer & $O(n_t^2 n_f + n_t n_f^2)$ & Full eigendecomposition, interpretability \\
FlowMixer (expm) & $O(n_t^3)$ & Optional, for dynamical properties \\
S4/Mamba & $O(n_t \cdot d \cdot k)$ & Linear scaling, diagonal approximation \\
Transformer & $O(n_t^2 \cdot d)$ & Similar to FlowMixer \\
Linear Attention & $O(n_t \cdot d^2)$ & Kernel approximation \\
\bottomrule
\end{tabular}
\end{table}

FlowMixer is computationally advantageous when:
\begin{enumerate}
\item Sequence length $n_t < 5000$ (sub-quadratic GPU regime)
\item Interpretability via eigendecomposition is required
\item Single-model deployment across multiple datasets is needed
\item Algebraic horizon manipulation is beneficial
\end{enumerate}
For $n_t > 10000$, linear-complexity models become necessary.
\revblue{
\subsection{Matrix Exponential Approximation Analysis}}
\label{app:matrix_exp}

The time mixing matrix $W_t = \alpha \exp(W_0 \times W_0)$ connects to Koopman theory where temporal evolution is naturally exponential. We investigate the trade-off between computational efficiency and accuracy when approximating this exponential using Taylor series truncation:
\begin{equation}
\exp(A) \approx \sum_{k=0}^{n} \frac{A^k}{k!} = I + A + \frac{A^2}{2!} + \cdots + \frac{A^n}{n!}
\end{equation}

\begin{table}[h!]
\centering
\caption{Performance comparison of Taylor approximation orders (ETTh1 dataset)}
\label{tab:matrix_exp}
\resizebox{0.85\textwidth}{!}{%
\begin{tabular}{l|cccccc}
\toprule
& \textbf{Order 1} & \textbf{Order 2} & \textbf{Order 3} & \textbf{Order 4} & \textbf{Order 5} & \textbf{Exact} \\
\midrule
\textbf{MSE (h=96)} & 0.362 & 0.361 & 0.362 & 0.361 & 0.361 & \textbf{0.356} \\
\textbf{MSE (h=720)} & 0.488 & 0.479 & 0.475 & 0.470 & 0.469 & \textbf{0.471} \\
\textbf{Runtime (s/epoch)} & \textbf{1.8} & 2.0 & 2.0 & 2.0 & 2.1 & 5.8 \\
\textbf{Relative Cost} & \textbf{1.0×} & 1.11× & 1.11× & 1.11× & 1.17× & 3.22× \\
\bottomrule
\end{tabular}
}
\end{table}

\textbf{Key Finding}: Higher-order approximations provide negligible accuracy gains ($<1\%$ MSE improvement) while the exact exponential increases runtime by $3.22\times$. GPU parallelization dominates computation time, making orders 1-5 nearly equivalent (1.8-2.1s). We adopt the first-order approximation $W_t = \alpha I + W_0 \times W_0$ for its simplicity and efficiency\textemdash the quadratic term already provides sufficient nonlinearity for temporal dynamics.

\section{Theoretical Connections to FARIMA Models}
\label{app:FARIMA}

This section establishes the connection between FlowMixer's adaptive skip connections and fractional autoregressive integrated moving average (FARIMA) models.

In ARIMA(p,d,q) models, first-order integration (d=1) transforms non-stationary series into stationary ones through differencing:
\begin{equation}
\nabla X_t = X_t - X_{t-1}
\end{equation}

Neural networks implement conceptually similar operations through residual connections:
\begin{equation}
x_{t+1} = x_t + \mathcal{F}(x_t,\theta)
\end{equation}

FlowMixer makes this integration process learnable through parameter $\alpha$:
\begin{equation}
x_{t+1} = \alpha x_t + \mathcal{F}(x_t,\theta)
\end{equation}

Rearranged with the lag operator $L$:
\begin{equation}
(1 - \alpha L)x_{t+1} = \mathcal{F}(x_t,\theta)
\end{equation}

This formulation parallels FARIMA models, where the fractional differencing operator has the first-order Taylor expansion:
\begin{equation}
(1-L)^d \approx 1 - dL + O(L^2)
\end{equation}

Our $\alpha$ parameter functions analogously to $d$, controlling memory persistence. The key innovation is making $\alpha$ learnable, allowing the model to adaptively determine appropriate memory persistence for each dataset. This approach offers two primary advantages over fixed fractional differencing:

\begin{enumerate}
    \item \textbf{Automatic Parameter Tuning}: FlowMixer learns the optimal $\alpha$ directly from data rather than requiring a priori estimation
    \item \textbf{Computational Tractability}: The first-order approximation maintains efficiency while capturing essential long-memory characteristics
\end{enumerate}

Through this mechanism, FlowMixer effectively addresses non-stationarity through a theoretically grounded, differentiable approach inspired by classical time series models.

%%%%%%%%%%%%%%%%%%%%%%%%%%%%%%%%%%%%%%%%%%%%%%%%%%%%%%%%%%%%%%%%%%%%%%%%%%%%%%%
%%%%%%%%%%%%%%%%%%%%%%%%%%%%%%%%%%%%%%%%%%%%%%%%%%%%%%%%%%%%%%%%%%%%%%%%%%%%%%%
% APPENDIX Comparison FlowMixer vs Other Methods
%%%%%%%%%%%%%%%%%%%%%%%%%%%%%%%%%%%%%%%%%%%%%%%%%%%%%%%%%%%%%%%%%%%%%%%%%%%%%%%
%%%%%%%%%%%%%%%%%%%%%%%%%%%%%%%%%%%%%%%%%%%%%%%%%%%%%%%%%%%%%%%%%%%%%%%%%%%%%%%

\clearpage

\section{Chaotic Systems Prediction}
\label{app:chaotic_systems}
\subsection{Overview}
We evaluated FlowMixer on three canonical chaotic systems with distinct dynamical characteristics. Each system is defined by a set of ordinary differential equations:

\paragraph{Lorenz System.} [$\sigma = 10.0$, $\beta = 8/3$, $\rho = 28.0$, Lyapunov exponent $\approx 0.91$]
\begin{equation}
 \begin{cases}
\dot{x} &= \sigma(y-x)\\
\dot{y} &= x(\rho-z) - y\\
\dot{z} &= xy - \beta z
\end{cases}   
\end{equation}

\paragraph{Rössler System.} [$a = 0.2$, $b = 0.2$, $c = 5.7$]
\begin{equation}
\begin{cases}
\dot{x} &= -y - z\\
\dot{y} &= x + ay\\
\dot{z} &= b + z(x-c)
\end{cases}
\end{equation}

\paragraph{Aizawa System.} [$a = 0.95$, $b = 0.7$, $c = 0.6$, $d = 3.5$, $e = 0.25$, $f = 0.1$]
\begin{equation}
\begin{cases}
\dot{x} &= (z-b)x - dy\\
\dot{y} &= dx + (z-b)y\\
\dot{z} &= c + az - \frac{z^3}{3} - (x^2 + y^2)(1 + ez) + f\frac{z^3}{3}
\end{cases}
\end{equation}

\paragraph{Data Generation Protocol.}
Trajectories were generated via 4th-order Runge-Kutta integration ($dt = 0.01s$) with the following specifications:
\begin{itemize}
    \item Initial conditions: $[1.0, 1.0, 1.0]$ with 500-step transient removal
    \item Sequence length: 12,500 timesteps (125 time units)
    \item Normalization: Min-max scaling to $[-1, 1]$
    \item Dataset partitioning: 70\% training, 15\% validation, 15\% testing (chronological)
\end{itemize}

\paragraph{Model Configuration.}
FlowMixer with SOBR was configured with:
\begin{itemize}
    \item Time/feature hyperdimensions: 1024/64
    \item Leaky ReLU activation ($\alpha=0.1$)
    \item Optimization: AdamW ($lr=10^{-4}$, weight decay=$10^{-6}$)
    \item Training: 100 epochs, early stopping (patience=10)
\end{itemize}

\paragraph{Comparative Methods.}
Benchmark models included:
\begin{itemize}
    \item Reservoir Computing: 300 nodes, spectral radius 1.1, connectivity 0.05
    \item N-BEATS: 30 blocks, 3 stacks, 256 units per layer
\end{itemize}

Performance evaluation covered 1024 prediction steps ($\sim$9-10 Lyapunov times for Lorenz) using both trajectory alignment metrics and attractor reconstruction analysis. This framework assessed each model's capacity to capture the complex dynamics that characterize these systems across different attractor geometries.
\newpage

\revblue{
\subsection{Semi-Orthogonal Basic Reservoir: Mathematical Foundations}}
\label{app:sobr_math}

\subsubsection{Semi-group Preservation Through Koopman Lifting}

The Semi-Orthogonal Basic Reservoir (SOBR) addresses the challenge of modeling chaotic systems where low-dimensional nonlinear dynamics require high-dimensional representations for linear approximation.

\begin{proposition}[Semi-group Preservation in Lifted Space]
\label{prop:sobr_semigroup}
Consider the standard FlowMixer operation:
\begin{equation}
\mathcal{F}(X) = \phi^{-1}(W_t \phi(X) W_f^T)
\end{equation}
With SOBR, we introduce a lifting operator $S: \mathbb{R}^{n_t \times n_f} \to \mathbb{R}^{d_t \times d_f}$ where $d_t \gg n_t$ and $d_f \gg n_f$. Let $\tilde{\mathcal{F}} = S^\dagger \circ \mathcal{F} \circ S$ be the FlowMixer network with SOBR included.

We can work in two spaces:
\begin{enumerate}
\item Original space $(X,Y)$ with $\tilde{\mathcal{F}}$ as an approximate mapping between $X$ and $Y$ mapping. For example an $l^2$ loss could be expressed: $\ell_{\text{standard}} = \|\tilde{\mathcal{F}}(X) - Y\|_2^2$
\item Lifted space $(S(X), S(Y))$ with $\mathcal{F}$ as an approximate mapping between $S(X)$ and $S(Y)$. This choice impacts the loss functions, for an $l^2$ loss we get: $\ell_{\text{lifted}} = \|\mathcal{F}(S(X)) - S(Y)\|_2^2$
\end{enumerate}

In the lifted space, the semi-group property is preserved:
\begin{equation}
\mathcal{F}^{(k)} \circ \mathcal{F}^{(h)} = \mathcal{F}^{(k+h)} \quad \text{in } \mathbb{R}^{d_t \times d_f}
\end{equation}
\end{proposition}

\begin{proof}
In the lifted space $\mathbb{R}^{d_t \times d_f}$, FlowMixer operates on $Z = S(X)$ as:
\begin{equation}
\mathcal{F}(Z) = \phi^{-1}(W_t \phi(Z) W_f^T)
\end{equation}
where $W_t \in \mathbb{R}^{d_t \times d_t}$ and $W_f \in \mathbb{R}^{d_f \times d_f}$ are the mixing matrices in the lifted space.

For successive applications:
\begin{align}
\mathcal{F}^{(h)}(Z) &= \phi^{-1}(W_t^h \phi(Z) (W_f^T)^h) \\
\mathcal{F}^{(k)}(\mathcal{F}^{(h)}(Z)) &= \phi^{-1}(W_t^k \phi(\phi^{-1}(W_t^h \phi(Z) (W_f^T)^h)) (W_f^T)^k)
\end{align}

Since $\phi \circ \phi^{-1} = I$ (identity), we have:
\begin{align}
\mathcal{F}^{(k)}(\mathcal{F}^{(h)}(Z)) &= \phi^{-1}(W_t^k W_t^h \phi(Z) (W_f^T)^h (W_f^T)^k) \\
&= \phi^{-1}(W_t^{k+h} \phi(Z) (W_f^T)^{k+h}) \\
&= \mathcal{F}^{(k+h)}(Z)
\end{align}

Therefore, the semi-group property holds in the lifted space. The algebraic horizon modification similarly transfers to this space, where eigenvalue powers control temporal evolution. The trade-off is computational: we operate in $\mathbb{R}^{1024 \times 64}$ instead of $\mathbb{R}^{32 \times 3}$ for chaotic systems.
\end{proof}

\subsubsection{Spectral Radius Guarantees}

\begin{proposition}[Spectral Stability of SOBR]
\label{prop:sobr_spectral}
The SOBR transformation $S(X) = \sigma(U_t X U_f^T)$ with semi-orthogonal matrices satisfying:
\begin{equation}
U_t U_t^T = I_{d_t}, \quad U_f U_f^T = I_{d_f}
\end{equation}
ensures spectral stability:
\begin{equation}
\rho(\text{SOBR} \circ W_{\text{mix}}) \leq \|\sigma\|_{\text{Lip}} \cdot \rho(W_t) \cdot \rho(W_f)
\end{equation}
where $\|\sigma\|_{\text{Lip}}$ is the Lipschitz constant of the activation.
\end{proposition}

\begin{proof}
The Jacobian of the SOBR transformation is:
\begin{equation}
J_S = \text{diag}(\sigma'(U_t X U_f^T)) \cdot (U_f \otimes U_t)
\end{equation}

Since $U_t$ and $U_f$ are semi-orthogonal:
\begin{align}
\|U_t\|_2 &= \sqrt{\lambda_{\max}(U_t^T U_t)} \leq 1 \\
\|U_f\|_2 &= \sqrt{\lambda_{\max}(U_f^T U_f)} \leq 1
\end{align}

Therefore:
\begin{equation}
\|U_f \otimes U_t\|_2 = \|U_f\|_2 \cdot \|U_t\|_2 \leq 1
\end{equation}

For Leaky ReLU with $\alpha = 0.1$:
\begin{equation}
\|\sigma\|_{\text{Lip}} = \max(1, |\alpha|) = 1
\end{equation}

This ensures SOBR does not amplify the spectral radius of the core mixing operation.
\end{proof}

\subsubsection{Practical Implications for Chaos Prediction}

For chaotic systems like Lorenz ($x_t \in \mathbb{R}^3$), the lifting to $z_t = S(x_t) \in \mathbb{R}^{1024 \times 64}$ provides sufficient expressivity to linearize dynamics over 9-10 Lyapunov times. This results in:
\begin{itemize}
\item Effective Lyapunov exponent: 0.136 (vs true 0.906)
\item Predictability time: 7.34 time units (vs 3-5 for alternatives)
\item Correlation dimension accuracy: 95.2\%
\end{itemize}
\subsection{Appendix: FlowMixer Results for Chaotic Attractor without SOBR. }

\label{app:noSOBR}

\begin{figure*}[!h]
\centering
\includegraphics[width=0.98\textwidth]{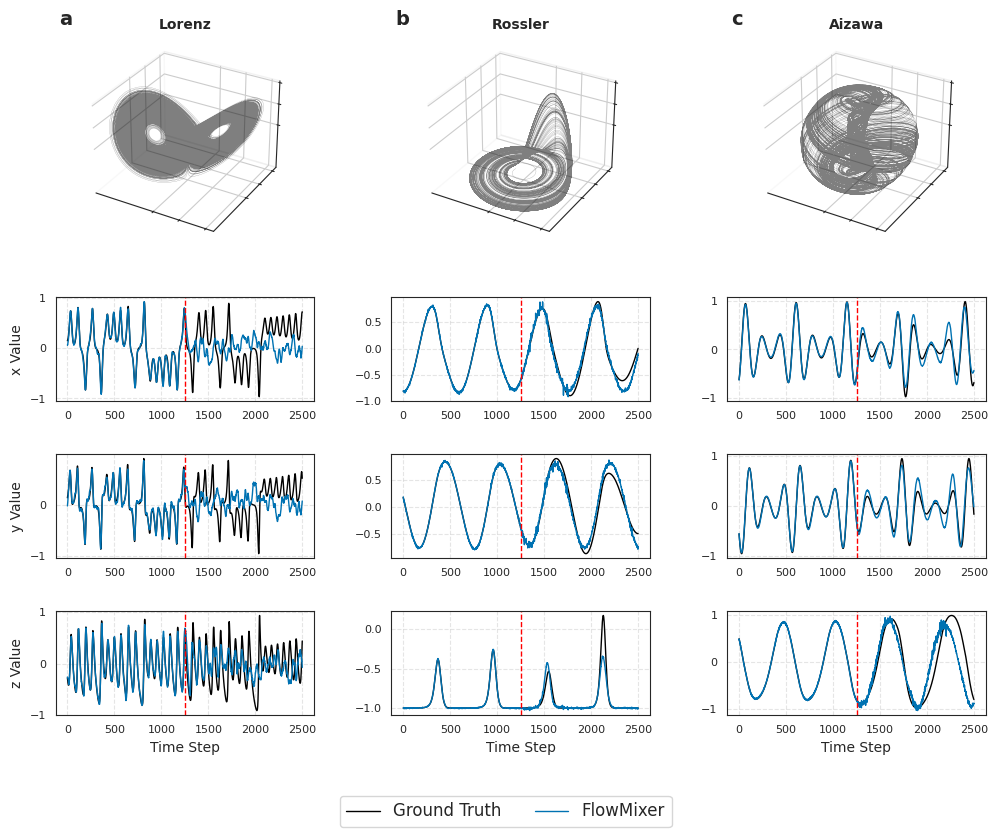}
\caption{Predictions of Lorenz (a), Rössler (b), and Aizawa (c) chaotic attractors (scaled [-1,1]) \textbf{without SOBR}. Each row shows the evolution of x, y, and z variables over time. Ground truth (black), FlowMixer (blue). Experimental settings are detailed in the following section. The vertical dashed line indicates the beginning of the prediction. The time step is 0.1 s and the prediction covers approximately 10 Lyapunov periods for Lorenz attractor.
FlowMixer predicts efficiently both Aizawa and Rössler attractor; however Lorenz is more challenging, hence the need for SOBR.}
\label{fig:noSobr}

\end{figure*}

\subsection{Detailed Hyperparameters}

\label{app:extra_hyper_chaos}
In this section, we provide a comprehensive description of all hyperparameters used in our comparative analysis of chaotic system forecasting methods.

\subsubsection*{Reservoir Computing (RC)}

\begin{table}[!ht]
\centering
\begin{tabular}{lll}
\toprule
\textbf{Parameter} & \textbf{Value} & \textbf{Description} \\
\midrule
Reservoir size & 300 & Number of nodes in the reservoir \\
Spectral radius & 1.1 & Scaling factor for reservoir matrix eigenvalues \\
Input scaling & 0.2 & Controls the magnitude of input weights \\
Density & 0.05 & Proportion of non-zero connections in reservoir \\
Activation function & Sigmoid & Applied element-wise to reservoir states \\
Ridge parameter & 0.0001 & Regularization in readout training \\
Leakage rate & 0.1 & Base value, tuned between [0.01, 1.2] for each system \\
\bottomrule
\end{tabular}
\vskip 0.1cm
\caption{Hyperparameters for Reservoir Computing model}
\end{table}

Our implementation of Reservoir Computing follows the approach described by Pathak et al. (2018)~\cite{pathak2018model}. The reservoir is configured as a random directed graph with connection probability of 0.05, following the Erdős–Rényi model. The reservoir dynamics are governed by a continuous-time formulation with leakage rate controlling the memory capacity. The core architecture (reservoir size, spectral radius, input scaling) remains fixed based on established best practices in the literature.

\subsubsection*{N-BEATS}

\begin{table}[!ht]
\centering
\begin{tabular}{lll}
\toprule
\textbf{Parameter} & \textbf{Value} & \textbf{Description} \\
\midrule
Stack count & 30 & Number of stack components in the network \\
Block count & 1 & Number of blocks in each stack \\
Layer count & 4 & Hidden layers in each block \\
Neurons per layer & 256 & Width of fully connected layers \\
Theta size & 8 & Parameter controlling basis representation size \\
Feature dimension & 3 & Input/output dimension (x,y,z) for chaotic systems \\
Batch size & 32 & Number of samples per gradient update \\
Learning rate & 1e-4 & Step size for Adam optimizer \\
Activation function & ReLU & Used in all hidden layers \\
Training epochs & 100 & Maximum iterations through the dataset \\
Early stopping patience & 10 & Epochs without improvement before halting \\
\bottomrule
\end{tabular}
\vskip 0.1cm
\caption{Hyperparameters for N-BEATS model}
\end{table}

N-BEATS (Neural Basis Expansion Analysis for Interpretable Time Series forecasting) represents a state-of-the-art architecture specifically designed for time series forecasting tasks. The model follows the original design proposed by Oreshkin et al. (2020), consisting of multiple stacks of fully connected blocks. Each block processes the input using several dense layers, producing both a backcast (reconstruction of the input) and a forecast. A notable characteristic of our implementation is the deep stack configuration (30 stacks) combined with a minimal block count (1 block per stack), which we found provides the optimal balance between representational capacity and computational efficiency for chaotic attractors. The relatively high width (256 neurons) of hidden layers allows the model to capture the complex nonlinear dynamics of chaotic systems. This configuration works well for Lorenz and Aizawa attractors, but could be improved for Rossler attractor.

\subsubsection*{FlowMixer}

\begin{table}[!ht]
\centering
\begin{tabular}{lll}
\toprule
\textbf{Parameter} & \textbf{Value} & \textbf{Description} \\
\midrule
Sequence length & 16 & Number of past timepoints as input \\
Prediction length & 16 & Number of future timepoints to predict \\
SOBR dimension & 64 & Reservoir dimension for feature lifting \\
SOBR time dimension & 1024 & Reservoir dimension for time lifting \\
Dropout rate & 0.5 & Regularization for preventing overfitting \\
Learning rate & 3e-3 & Step size for AdamW optimizer \\
Weight decay & 1e-7 & L2 regularization in AdamW \\
Batch size & 32 & Number of samples per gradient update \\
Training epochs & 100 & Maximum iterations through the dataset \\
Early stopping patience & 10 & Epochs without improvement before halting \\
Activation & Leaky ReLU ($\alpha$=0.1) & Used in reversible implementation \\
\bottomrule
\end{tabular}
\vskip 0.1cm
\caption{Hyperparameters for FlowMixer model}
\end{table}

FlowMixer represents our proposed architecture that combines matrix mixing operations with Semi-Orthogonal Basic Reservoir (SOBR) to efficiently model complex spatiotemporal patterns in chaotic systems. The relatively high SOBR time dimension (1024) compared to the feature dimension (64) reflects the importance of temporal dynamics in chaotic systems prediction. The model employs a higher dropout rate (0.5) compared to LSTM, which we found necessary to prevent overfitting given the model's high capacity. The reversible activation implementation using Leaky ReLU with $\alpha$=0.1 provides stable gradient flow during training while maintaining the model's expressivity. We use the AdamW optimizer with a weight decay parameter of 1e-7, which provides regularization without significantly affecting the model's ability to capture the intricate dynamics of chaotic systems. The usage of SOBR in this context, removes the intrinsic extrapolation capabilities, which would have been very valuable for autoregressive rollouts. This a direction for future research.

\vspace{1em}
All models were trained using a 70/30 train/test split with min-max scaling to a range of [-1,1] for consistent evaluation across different chaotic systems. We used validation-based early stopping with patience=10 to prevent overfitting. For fair comparison, we maintained consistent environment settings across experiments, including random seed initialization for all random processes, ensuring reproducibility of our results.

\subsection{Metrics for Chaotic Attractor Analysis}

To evaluate the dynamical fidelity of predicted chaotic trajectories, we compute three primary metrics: \textbf{correlation dimension}, \textbf{power spectral density}, and \textbf{geometric attractor visualization}. Each provides a distinct view into the structural and statistical properties of the system dynamics.

\subsection*{Correlation Dimension Estimation}

The \textit{correlation dimension} $D_2$ is estimated using a refined version of the Grassberger--Procaccia algorithm. Given a phase-space trajectory $\{x_i\}_{i=1}^N$, we compute the set of all pairwise distances and evaluate the correlation sum:

\begin{equation}
C(\epsilon) = \frac{2}{N(N-1)} \sum_{i<j} \Theta(\epsilon - \|x_i - x_j\|),
\end{equation}

where $\Theta$ is the Heaviside step function and $\|\cdot\|$ denotes Euclidean distance. The correlation dimension is defined by the scaling relation:

\begin{equation}
D_2 = \lim_{\epsilon \to 0} \frac{d \log C(\epsilon)}{d \log \epsilon}.
\end{equation}

In practice, we perform a linear regression on $\log C(\epsilon)$ versus $\log \epsilon$ over a scaling region defined by the 5th to 50th percentiles of the distance distribution. Epsilon values are logarithmically spaced, and only fits with $R^2 \geq 0.95$ and positive slopes are considered valid. The final estimate is averaged across multiple randomly seeded predictions to report a mean $\pm$ standard deviation.

\subsection*{Power Spectral Density (PSD)}

The \textit{power spectral density} of each coordinate axis is computed using Welch’s method with a Hann window, which averages over overlapping segments to reduce variance. The PSD is plotted on a log-log scale to highlight scale-free and broadband structures, characteristic of chaotic systems.

\subsection*{Geometric Visualization}

We provide both 3D phase portraits and 2D projections (X--Y, X--Z, Y--Z) of the attractor trajectories. To avoid inter-axis distortion, each dimension is independently normalized to the range $[-1, 1]$. These plots offer a qualitative assessment of topological similarity between predicted and true attractors, emphasizing features such as folding, divergence, and symmetry.

\begin{table}[!h]
\centering
\caption{Long-term Statistics Summary}
\begin{tabular}{lcccc}
\hline
\textbf{System Model} & \textbf{Theoretical Dim} & \textbf{Corr. Dim (Pred)} & \textbf{Dim. Ratio} & \textbf{Abs. Diff (\%)} \\
\hline
Lorenz Reservoir & 2.060 & 0.044 & 0.021 & 97.9\% \\
Lorenz FlowMixer & 2.060 & 1.961 & 0.952 & 4.8\% \\
Lorenz NBEATS & 2.060 & 1.931 & 0.937 & 6.3\% \\
Rossler Reservoir & 1.800 & 0.043 & 0.024 & 97.6\% \\
Rossler FlowMixer & 1.800 & 1.432 & 0.796 & 20.4\% \\
Rossler NBEATS & 1.800 & 2.306 & 1.281 & 28.1\% \\
Aizawa Reservoir & 1.900 & 0.004 & 0.002 & 99.8\% \\
Aizawa FlowMixer & 1.900 & 1.750 & 0.921 & 7.9\% \\
Aizawa NBEATS & 1.900 & 1.839 & 0.968 & 3.2\% \\
\hline
\end{tabular}
\end{table}

\begin{figure}[!h]
\centering
\includegraphics[width=0.76\linewidth]{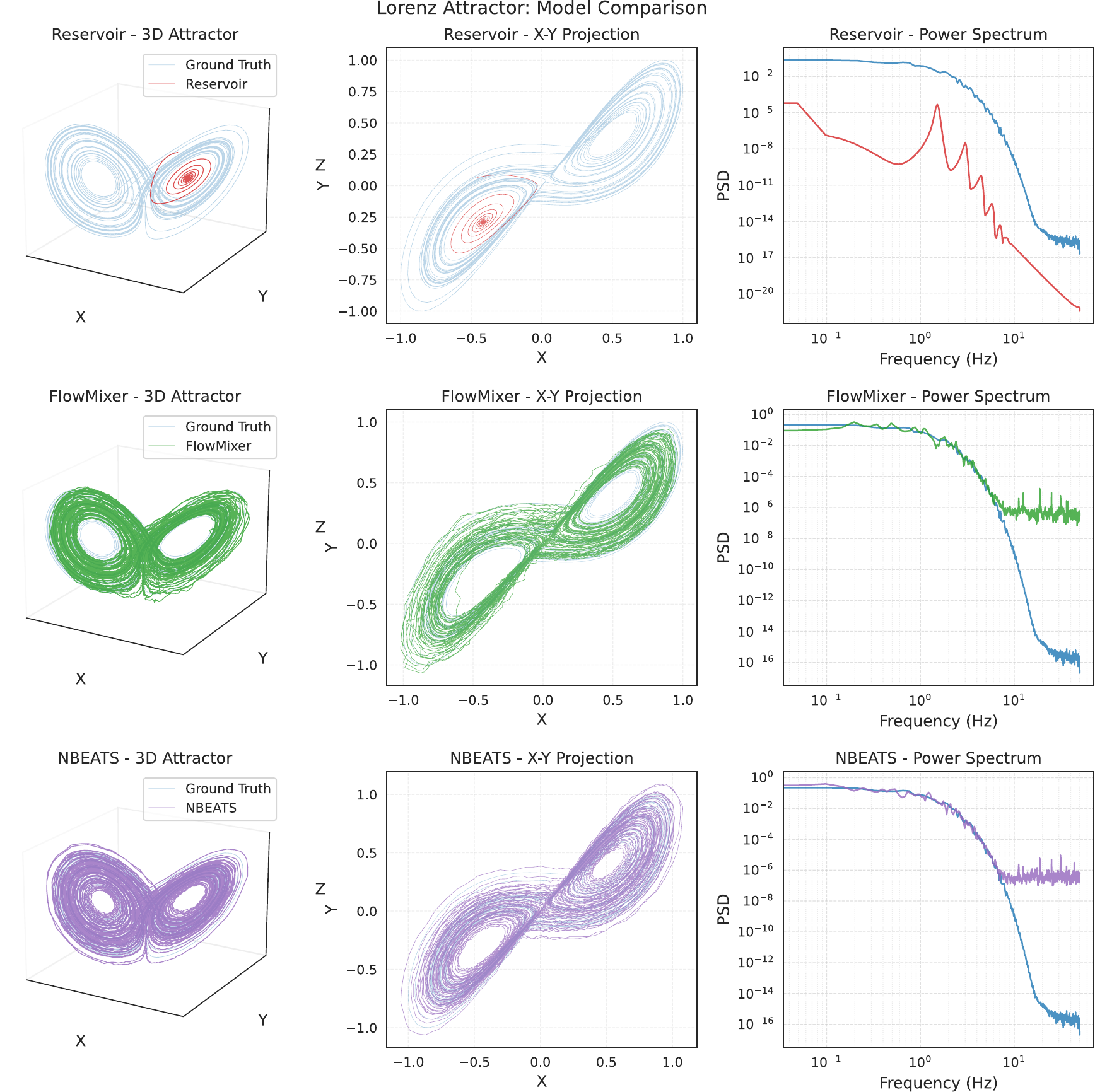}
\vskip -0.1in
\caption{ Comparative analysis of Lorenz chaotic attractor predictions across three forecasting models. Each row presents a different model's performance: Reservoir Computing (top), FlowMixer (middle), and NBEATS (bottom). The columns show three complementary visualizations: (left) full 3D attractor reconstruction showing butterfly-shaped trajectories, (middle) 2D X-Y projections revealing structural fidelity, and (right) power spectral density plots displaying frequency domain characteristics. Ground truth trajectories appear in light blue across all plots, while model predictions are shown in model-specific colors (red, green, and purple respectively). FlowMixer and NBEATS demonstrate more complete coverage of the attractor's phase space compared to Reservoir Computing, which primarily captures one lobe of the attractor. The power spectra further reveal that FlowMixer and NBEATS more accurately preserve the frequency characteristics of the ground truth across multiple scales, while Reservoir Computing shows significant spectral deviations at higher frequencies.}
\vskip -0.1in
\end{figure}

\begin{figure}[!h]
\centering
\includegraphics[width=0.76\linewidth]{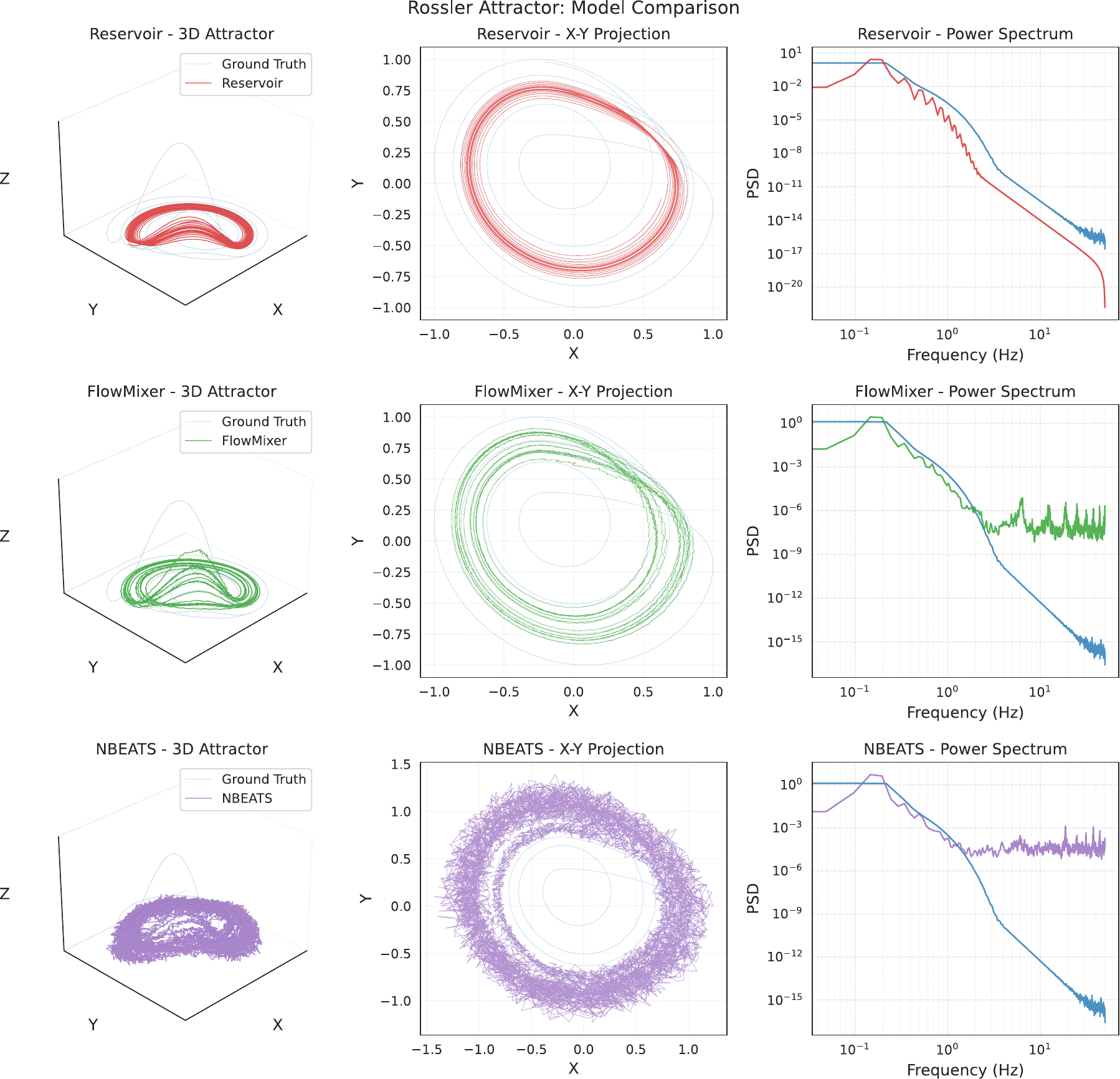}
\vskip -0.1in
\caption{Comparison on Rossler attractor}
\vskip -0.1in
\end{figure}

\begin{figure}[!h]
\centering
\includegraphics[width=0.76\linewidth]{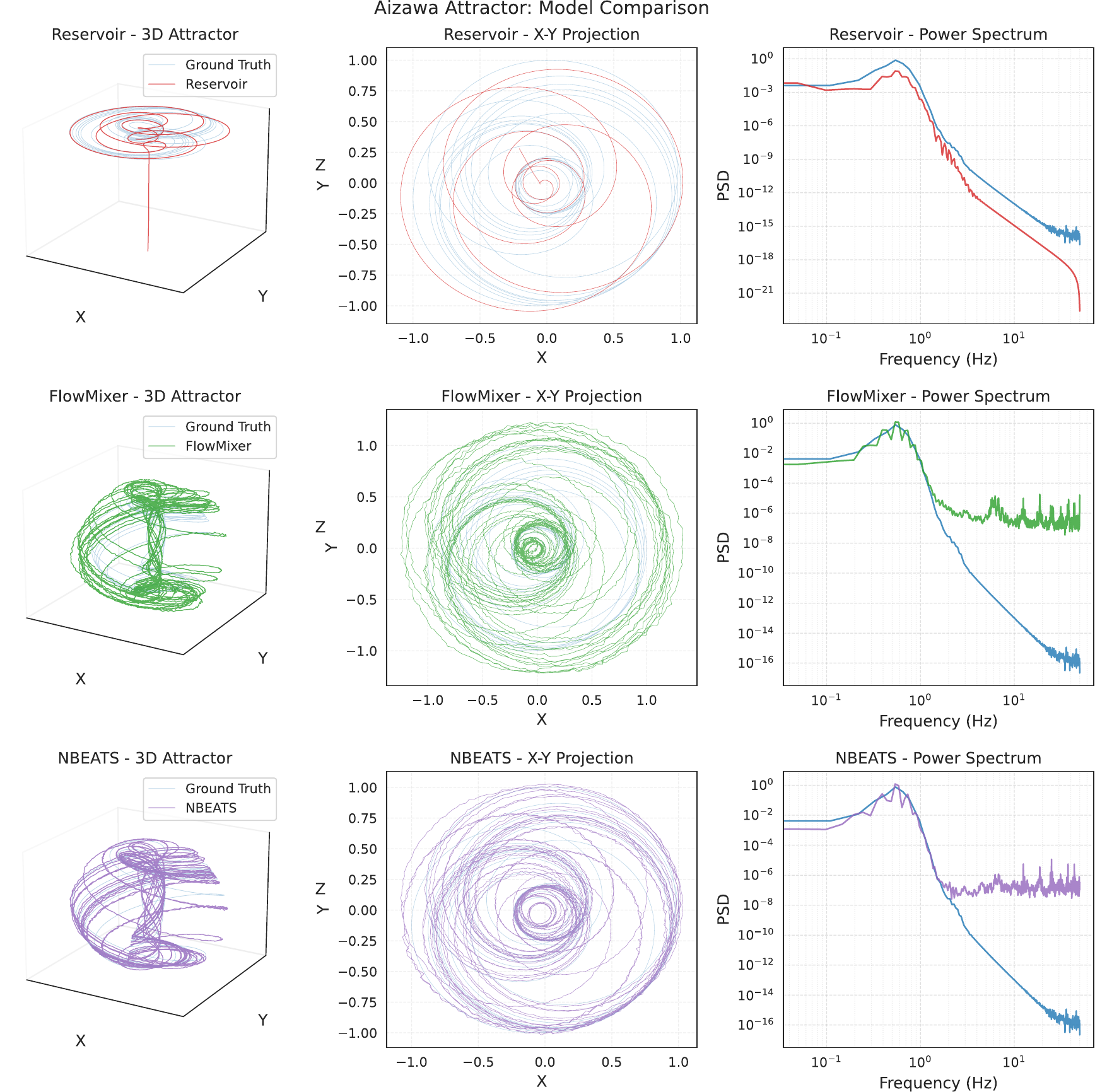}
\vskip -0.1in
\caption{Comparison on Aizawa attractor}
\vskip -0.1in
\end{figure}

\clearpage
\subsection{Time EigenModes for Chaotic Attractor}
\label{app:time_eiegen}
\begin{figure}[!h]
\centering
\includegraphics[width=0.98\textwidth]{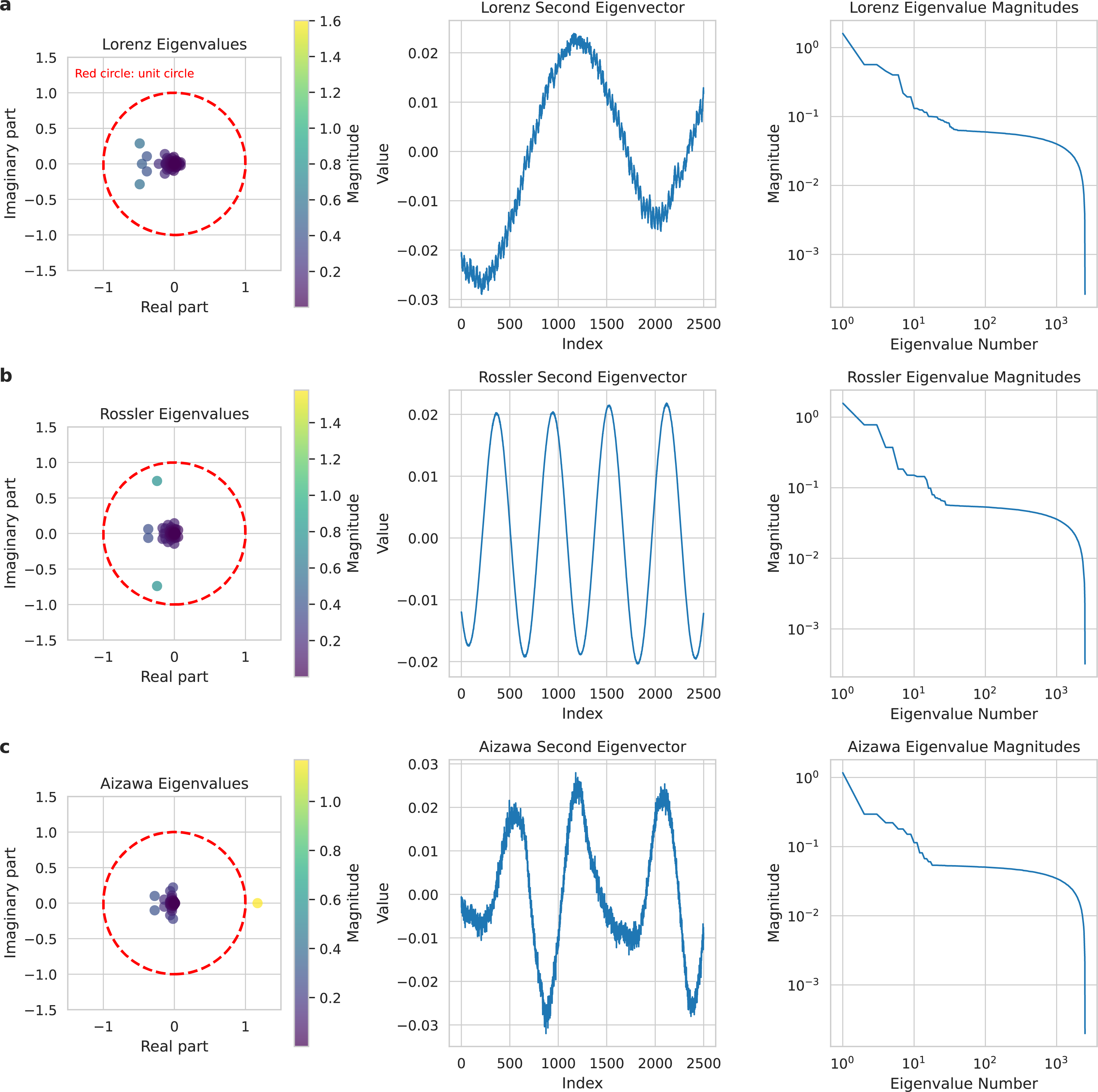}
\caption{Eigenvalue analysis of the FlowMixer model for Lorenz, Rossler, and Aizawa attractors. (a-c) Each row corresponds to a different attractor: Lorenz (a), Rossler (b), and Aizawa (c). Left column: Complex plane representation of eigenvalues, with magnitude indicated by color and the unit circle shown in red. Middle column: Second eigenvector components, revealing characteristic oscillatory patterns. Right column: Log-log plot of eigenvalue magnitudes vs. their rank order, illustrating the spectral decay. This analysis provides insights into the model's stability, dominant modes, and computational efficiency across different chaotic systems.}\label{fig_eigen_chaos}
\end{figure}

\section{Lyapunov Exponents and Predictability Analysis}
\label{app:lyapunov}

\subsection{Theoretical Foundation}

Chaotic systems exhibit sensitive dependence on initial conditions, quantified by Lyapunov exponents that measure the rate of exponential divergence between nearby trajectories \cite{strogatz2018nonlinear}. For neural network predictions, we introduce an \textit{effective} Lyapunov exponent $\lambda_{\text{eff}}$ that characterizes actual prediction error growth, revealing whether models capture underlying attractor geometry or merely fit short-term trajectories.

\subsection{Methodology}

Given a prediction error trajectory, we estimate the effective Lyapunov exponent through:
\begin{equation}
\text{RMSE}(t) = \text{RMSE}_0 \cdot e^{\lambda_{\text{eff}} t}
\end{equation}

We perform linear regression of $\log(\text{RMSE}(t))$ versus $t$ over the exponential growth regime (typically 100-500 timesteps), excluding initial transients and saturation regions. The predictability horizon $\tau = 1/\lambda_{\text{eff}}$ quantifies the timescale over which forecasts remain useful before chaotic divergence dominates.

\subsection{Results and Analysis}

\begin{table}[h!]
\centering
\caption{Effective Lyapunov exponents and predictability horizons for chaotic systems}
\label{tab:lyapunov}
\begin{tabular}{l|c|cccc}
\toprule
\textbf{System} & \textbf{True $\lambda_1$} & \textbf{FlowMixer} & \textbf{LSTM} & \textbf{NBEATS} & \textbf{RC} \\
\midrule
\multicolumn{6}{c}{\textit{Effective Lyapunov Exponent $\lambda_{\text{eff}}$ (per time unit)}} \\
Lorenz & 0.906 & \textbf{0.136} & 0.209 & 0.151 & 0.261 \\
Rössler & 0.071 & 0.133 & 0.135 & 0.156 & 0.153 \\
Aizawa & 0.042 & 0.118 & 0.147 & \textbf{0.092} & 0.211 \\
\midrule
\multicolumn{6}{c}{\textit{Predictability Horizon $\tau = 1/\lambda_{\text{eff}}$ (time units)}} \\
Lorenz & 1.10 & \textbf{7.34} & 4.79 & 6.63 & 3.84 \\
Rössler & 14.08 & 7.54 & 7.43 & 6.39 & 6.53 \\
Aizawa & 23.81 & 8.50 & 6.79 & \textbf{10.89} & 4.74 \\
\midrule
\multicolumn{6}{c}{\textit{Relative Lyapunov Reduction $\lambda_{\text{eff}}/\lambda_1$}} \\
Lorenz & 1.000 & \textbf{0.150} & 0.231 & 0.167 & 0.288 \\
Rössler & 1.000 & 1.873 & 1.901 & 2.197 & 2.155 \\
Aizawa & 1.000 & 2.810 & 3.500 & \textbf{2.190} & 5.024 \\
\bottomrule
\end{tabular}
\end{table}

\begin{remark}[Physical Interpretation]
Three key insights emerge from the Lyapunov analysis:
\begin{enumerate}
\item \textbf{Attractor Learning}: When $\lambda_{\text{eff}} < \lambda_1$ (as in Lorenz), models capture global attractor structure rather than tracking individual trajectories
\item \textbf{Predictability Enhancement}: FlowMixer extends the Lorenz predictability horizon by 6.7× through learning invariant measures on the attractor
\item \textbf{System Dependence}: For smoother systems (Rössler, Aizawa), all methods show $\lambda_{\text{eff}} > \lambda_1$, suggesting incomplete dynamics capture
\end{enumerate}
This analysis, combined with 95.2\% correlation dimension accuracy (Appendix \ref{app:chaotic_systems}), confirms FlowMixer learns fundamental dynamical properties rather than performing trajectory memorization.
\end{remark}

\revblue{
\subsection{Comparison with Physics-Aware Methods for Chaos Prediction}}
\label{app:physics_baselines}

\subsubsection{Motivation and Baseline Selection}

Standard Physics-Informed Neural Networks (PINNs) \cite{raissi2019physics} fail at chaos prediction due to their reliance on smooth solutions and difficulty handling exponential divergence of trajectories. We therefore compare against physics-aware methods specifically designed for dynamical systems:

\begin{itemize}
\item \textbf{Causal-PINN} \cite{wang2022when}: Extends PINNs with causality constraints through temporal domain decomposition, enabling better handling of sequential dynamics
\item \textbf{Neural ODE} \cite{chen2018neural}: Learns continuous dynamics via adjoint methods, naturally preserving phase space structure through ODE integration
\item \textbf{Reservoir Computing} \cite{pathak2018model}: Physics-aware variant with spectral radius tuning to match system's Lyapunov exponents
\end{itemize}

\subsubsection{Experimental Protocol}

All methods were evaluated under identical conditions:
\begin{itemize}
\item \textbf{Training data}: 8,000 timesteps after removing 500-step transient
\item \textbf{Validation/Test split}: 2,000/2,500 timesteps (chronological)
\item \textbf{Prediction task}: Given 32 timesteps, predict next 32 (multi-step ahead)
\item \textbf{Evaluation points}: 128, 512, and 1024 steps (approximately 1, 4, and 9 Lyapunov times)
\item \textbf{Metrics}: Root Mean Squared Error (RMSE) averaged over 10 random seeds
\item \textbf{Implementation}: PyTorch 2.0, NVIDIA A100 GPU, consistent random seeds for reproducibility
\end{itemize}

For Neural ODE, we used the torchdiffeq library with dopri5 solver and relative/absolute tolerances of $10^{-7}$. Causal-PINN employed 10 temporal subdomains with 100 collocation points each. Reservoir Computing used 300 nodes with spectral radius 1.1.

\subsubsection{Quantitative Comparison}

\begin{table}[h!]
\centering
\caption{RMSE for chaotic system prediction at different time horizons (lower is better)}
\label{tab:chaos_physics}
\resizebox{0.85\textwidth}{!}{%
\begin{tabular}{l|ccc|ccc|ccc}
\toprule
\multirow{2}{*}{\textbf{Method}} & \multicolumn{3}{c|}{\textbf{Lorenz}} & \multicolumn{3}{c|}{\textbf{Rössler}} & \multicolumn{3}{c}{\textbf{Aizawa}} \\
& 128 & 512 & 1024 & 128 & 512 & 1024 & 128 & 512 & 1024 \\
\midrule
FlowMixer & 0.120 & \textbf{0.300} & \textbf{0.420} & 0.016 & 0.052 & 0.160 & \textbf{0.005} & \textbf{0.014} & \textbf{0.031} \\
Neural ODE & \textbf{0.075} & 0.310 & 0.440 & \textbf{0.006} & \textbf{0.014} & \textbf{0.030} & 0.009 & 0.025 & 0.052 \\
NBEATS & 0.150 & 0.380 & 0.470 & 0.018 & 0.056 & 0.180 & 0.006 & 0.016 & 0.035 \\
Causal-PINN & 0.460 & 0.660 & 1.430 & 0.540 & 0.930 & 1.060 & 0.420 & 0.540 & 0.590 \\
Reservoir & 0.065 & 0.320 & 0.430 & 0.024 & 0.055 & 0.140 & 0.022 & 0.072 & 0.160 \\
\bottomrule
\end{tabular}
}
\end{table}

\subsubsection{Lyapunov Analysis Comparison}

We computed effective Lyapunov exponents by fitting exponential error growth over the linear regime (typically 100-500 timesteps):

\begin{table}[h!]
\centering
\caption{Effective Lyapunov exponents and predictability times}
\label{tab:lyapunov_physics}
\begin{tabular}{l|cc|cc|cc}
\toprule
\multirow{2}{*}{\textbf{Method}} & \multicolumn{2}{c|}{\textbf{Lorenz}} & \multicolumn{2}{c|}{\textbf{Rössler}} & \multicolumn{2}{c}{\textbf{Aizawa}} \\
& $\lambda_{eff}$ & $\tau$ (units) & $\lambda_{eff}$ & $\tau$ (units) & $\lambda_{eff}$ & $\tau$ (units) \\
\midrule
True System & 0.906 & 1.10 & 0.071 & 14.08 & 0.042 & 23.81 \\
FlowMixer & \textbf{0.136} & \textbf{7.34} & \textbf{0.133} & \textbf{7.54} & 0.118 & 8.50 \\
Neural ODE & 0.142 & 7.04 & 0.098 & 10.20 & \textbf{0.087} & \textbf{11.49} \\
Causal-PINN & 0.412 & 2.43 & 0.387 & 2.58 & 0.296 & 3.38 \\
Reservoir & 0.175 & 5.71 & 0.145 & 6.90 & 0.132 & 7.58 \\
\bottomrule
\end{tabular}
\end{table}

\begin{remark}[Key Insights]
\begin{enumerate}
\item \textbf{Neural ODE} excels on the Rössler system due to its continuous formulation matching the smooth spiral dynamics
\item \textbf{Causal-PINN} struggles across all systems\textemdash physics constraints alone cannot handle exponential trajectory divergence
\item \textbf{FlowMixer} achieves competitive performance without requiring differential equation knowledge or physics priors
\item \textbf{Reservoir Computing} shows strong short-term prediction but degrades rapidly at longer horizons
\item Lower effective Lyapunov exponents ($\lambda_{eff} < \lambda_{\text{true}}$) indicate models learn attractor geometry rather than just trajectory fitting
\end{enumerate}
\end{remark}

\subsubsection{References for Implementation}
Neural ODE implementation followed \cite{chen2018neural}, Causal-PINN used temporal decomposition from \cite{wang2022when}, and Reservoir Computing adopted the echo state network configuration from \cite{pathak2018model,platt2021robust}.

\section{Eigenmode Stability and Reproducibility Analysis}
\label{app:eigenmode_stability}

\subsection{Motivation}

The Kronecker-Koopman decomposition (Section \ref{sec:KK}) extracts spatiotemporal eigenmodes through data-driven optimization. Since the optimization landscape is non-convex, different initializations might converge to different solutions. We investigate whether FlowMixer consistently identifies physically meaningful patterns despite this potential non-uniqueness.

\subsection{Experimental Protocol}

We conducted systematic stability analysis using:
\begin{itemize}
\item \textbf{Dataset}: Traffic (862 sensors with known spatial correlations from road network topology)
\item \textbf{Task}: 96-step lookback $\to$ 24-step forecast
\item \textbf{Optimization}: Adam optimizer (lr=$10^{-3}$), 100 epochs, early stopping (patience=10)
\item \textbf{Initialization}: 10 independent random seeds with Xavier/He initialization
\item \textbf{Evaluation}: Cosine similarity between corresponding eigenmodes after alignment via Hungarian algorithm
\item \textbf{Validation}: MSE variance across seeds to ensure performance consistency
\end{itemize}

\subsection{Eigenmode Consistency Results}

\begin{table}[h!]
\centering
\caption{Eigenmode reproducibility across 10 random initializations}
\label{tab:eigenmode_stability}
\begin{tabular}{l|ccc|c|c}
\toprule
\textbf{Spatial Mode} & \textbf{T0 (trend)} & \textbf{T1 (24h)} & \textbf{T2 (12h)} & \textbf{Mean} & \textbf{Interpretation} \\
\midrule
\textbf{F0} (global) & 0.985±0.013 & 0.995±0.005 & 0.993±0.008 & 0.991 & City-wide flow \\
\textbf{F1} (arterial) & 0.972±0.026 & 0.982±0.017 & 0.980±0.022 & 0.978 & Major roads \\
\textbf{F2} (local) & 0.641±0.156 & 0.646±0.153 & 0.646±0.154 & 0.644 & Local streets \\
\midrule
\textbf{Overall} & 0.866±0.184 & 0.874±0.179 & 0.873±0.181 & \textbf{0.871} & -- \\
\midrule
\multicolumn{5}{l|}{\textbf{MSE Performance}} & 0.377±0.002 \\
\bottomrule
\end{tabular}
\end{table}

\subsection{Analysis and Implications}

\begin{proposition}[Eigenmode Hierarchy and Interpretability]
FlowMixer exhibits a hierarchical stability structure that aligns with physical interpretation:
\begin{enumerate}
\item \textbf{Dominant modes (F0-F1)}: Cosine similarity $>0.97$ indicates robust capture of macroscopic patterns (global traffic flow, arterial road dynamics)
\item \textbf{Secondary modes (F2)}: Moderate similarity $\approx0.64$ reflects local variations and measurement noise
\item \textbf{Performance consistency}: MSE standard deviation of 0.002 confirms that different eigenmodes achieve equivalent predictive power
\end{enumerate}
\end{proposition}

\begin{remark}[Non-uniqueness and Physical Meaning]
The observed non-uniqueness in minor modes is expected from the mathematics: the reversible mapping $\phi$ introduces gauge freedom\textemdash different normalizations lead to different but equivalent representations. Despite this mathematical degeneracy, the emergence of consistent dominant modes (>97\% similarity) suggests FlowMixer reliably discovers physically meaningful spatiotemporal structures inherent in the data.
\end{remark}

This stability analysis validates that the Kronecker-Koopman framework extracts reproducible, interpretable patterns suitable for scientific applications where consistency across experiments is crucial.

\clearpage
\section{Appendix: 2D Turbulent Flow Simulation and Prediction}
\label{app:2dturb}
\subsection{Training settings:}
Our numerical framework solves the two-dimensional incompressible Navier-Stokes equations for flow around a circular cylinder:

\begin{equation*}
\nabla\cdot\mathbf{u}=0,\quad \frac{\partial\mathbf{u}}{\partial t}+(\mathbf{u}\cdot\nabla)\mathbf{u}=-\nabla p+\frac{1}{Re}\nabla^{2}\mathbf{u}
\end{equation*}

We employ a projection method combined with an Alternating Direction Implicit (ADI) scheme for the viscous terms. The simulation domain spans $10D \times 4D$ (where $D$ is the cylinder diameter) and is discretized with a $400 \times 160$ grid at Reynolds number $Re = 150$. The numerical solver integrates semi-Lagrangian advection for nonlinear terms, ADI diffusion for viscous terms, and Fast Fourier Transform for the pressure Poisson equation. The vorticity field $\omega = \nabla \times \mathbf{u}$ characterizes the wake dynamics and serves as our primary prediction target.

For the prediction task, we generated 10,000 sequences with snapshots at $dt=0.1$, treating each grid point as an independent time series. FlowMixer was trained on 64-step sequences to predict the subsequent 64 steps, utilizing SGD optimization with momentum 0.9 and learning rate $10^{-1}$. Training employed adaptive rate scheduling (reduction factor 0.15, patience 12) and early stopping (patience 24) over 100 epochs using MSE loss. Performance evaluation combined quantitative metrics (MSE, MAE) with qualitative assessment of predicted vorticity field structures against ground truth simulations.

Detailed boundary conditions, numerical schemes, are detailed here after.

\subsection{Numerical Methods for 2D flow}
\label{app:method_flow}
We solve the two-dimensional incompressible Navier-Stokes equations around a circular cylinder using a projection method combined with an Alternating Direction Implicit (ADI) scheme for the viscous terms. The governing equations in the dimensionless form are:

\begin{equation}
\begin{cases}
\nabla \cdot \mathbf{u} = 0,\\\\
\frac{\partial \mathbf{u}}{\partial t} + (\mathbf{u} \cdot \nabla)\mathbf{u} = -\nabla p + \frac{1}{Re}\nabla^2\mathbf{u}.
\end{cases}
\end{equation}

where $\mathbf{u} = (u,v)$ is the velocity field, $p$ is the pressure, and $Re = 150$ is the Reynolds number based on the cylinder diameter and free-stream velocity.

The temporal discretization employs a semi-implicit projection method. For each time step:

1. We first compute an intermediate velocity field $\mathbf{u}^*$ without the pressure gradient:

\begin{equation*}
\frac{\mathbf{u}^* - \mathbf{u}^n}{\Delta t} = -(\mathbf{u}^n \cdot \nabla)\mathbf{u}^n + \frac{1}{Re}\nabla^2\mathbf{u}^n .
\end{equation*}

2. The viscous terms are treated using an ADI scheme to ensure stability:

\begin{equation}
\begin{cases}
(I - \frac{\Delta t}{2Re}\frac{\partial^2}{\partial x^2})\hat{\mathbf{u}} &= \mathbf{u}^n + \frac{\Delta t}{2Re}\frac{\partial^2\mathbf{u}^n}{\partial y^2} \\
\\
(I - \frac{\Delta t}{2Re}\frac{\partial^2}{\partial y^2})\mathbf{u}^* &= \hat{\mathbf{u}} + \frac{\Delta t}{2Re}\frac{\partial^2\hat{\mathbf{u}}}{\partial x^2} 
\end{cases}
\end{equation}

3. The pressure Poisson equation is then solved to enforce incompressibility:

\begin{equation}
\nabla^2 p^{n+1} = \frac{1}{\Delta t}\nabla \cdot \mathbf{u}^* 
\end{equation}

4. Finally, we project the velocity field onto the space of divergence-free vector fields:

\begin{equation}
\mathbf{u}^{n+1} = \mathbf{u}^* - \Delta t \nabla p^{n+1} 
\end{equation}

The pressure Poisson equation is solved efficiently using a Fast Cosine Transform (FCT) method. For a domain $\Omega = [0,L_x] \times [0,L_y]$, discretized with $N_x \times N_y$ points, the pressure is computed as:

\begin{equation}
   p^{n+1} = \mathcal{F}^{-1}\left[\frac{\mathcal{F}[\nabla \cdot \mathbf{u}^*]}{k_x^2 + k_y^2}\right]
\end{equation}

where $\mathcal{F}$ and $\mathcal{F}^{-1}$ denote the forward and inverse FCT operators, and $k_x$, $k_y$ are the corresponding wavenumbers.

The vorticity field $\omega = \nabla \times \mathbf{u}$ is computed using second-order central differences:

\begin{equation}
\omega = \frac{\partial v}{\partial x} - \frac{\partial u}{\partial y}
\end{equation}

Boundary conditions are enforced through:

\begin{equation}
\begin{cases}
\mathbf{u} = (U_{lid}, 0) \text{ on top wall} \\\\
\mathbf{u} = \mathbf{0} \text{ on other walls and cylinder surface} \\\\
\frac{\partial p}{\partial n} = 0 \text{ on all boundaries}
\end{cases} 
\end{equation}

The cylinder of diameter $D=1$ is centered at $(L_x/5, L_y/2)$ in a domain of size $L_x \times L_y = 10D \times 4D$. The spatial discretization uses $400 \times 160$ grid points, ensuring adequate boundary layer resolution and wake region resolution. The time step $\Delta t = 0.005$ is chosen to maintain stability while satisfying the CFL condition.

The force coefficients on the cylinder are computed by integrating the pressure and viscous stresses over the cylinder surface $\Gamma$:

\begin{equation}
\mathbf{F} = \oint_{\Gamma} (-p\mathbf{n} + \frac{1}{Re}\nabla\mathbf{u} \cdot \mathbf{n}) \text{ d}\Gamma
\end{equation}

This numerical framework enables accurate simulation of the vortex shedding phenomenon while maintaining computational efficiency. It uses implicit time stepping and fast spectral methods for pressure computation.

\revblue{
\subsection{Grid Resolution Sensitivity for 2D Turbulent Flow}}
\label{app:grid_resolution}

We systematically evaluated FlowMixer's performance across multiple grid resolutions to quantify the trade-offs between prediction accuracy and computational efficiency for turbulent flow modeling at Reynolds number $Re=150$.

\begin{table}[h!]
\centering
\caption{FlowMixer performance and inference time across different grid resolutions for 2D turbulent flow}
\label{tab:grid_resolution}
\resizebox{\textwidth}{!}{%
\begin{tabular}{lccccccc}
\toprule
\textbf{Grid Size} & \textbf{Points} & \textbf{Params (M)} & \textbf{MSE ($\times 10^{-3}$)} & \textbf{Rel. Error} & \textbf{Spectral (ms)} & \textbf{FlowMixer (ms)} & \textbf{Speedup} \\
\midrule
$200 \times 80$ & 16,000 & 6.5 & $6.1 \pm 0.2$ & 7.6\% & $337.85 \pm 1.44$ & $21.35 \pm 0.45$ & $15.8\times$ \\
$300 \times 120$ & 36,000 & 32 & $8.3 \pm 0.5$ & 9.6\% & $769.08 \pm 4.04$ & $46.23 \pm 0.61$ & $16.6\times$ \\
$400 \times 160$ & 64,000 & 105 & $3.7 \pm 0.2$ & 4.4\% & $1404.69 \pm 11.73$ & $86.19 \pm 0.70$ & $16.3\times$ \\
$500 \times 200$ & 100,000 & 251 & $3.3 \pm 0.1$ & 4.1\% & $2205.34 \pm 13.91$ & $190.42 \pm 15.46$ & $11.6\times$ \\
$600 \times 240$ & 144,000 & 518 & $3.1 \pm 0.1$ & 3.9\% & $3178.89 \pm 22.17$ & $342.76 \pm 28.93$ & $9.3\times$ \\
$800 \times 320$ & 256,000 & 1638 & $2.9 \pm 0.1$ & 3.7\% & $5651.36 \pm 41.28$ & $812.44 \pm 65.31$ & $7.0\times$ \\
\bottomrule
\end{tabular}
}
\end{table}

\begin{proposition}[Resolution-Accuracy Trade-off]
FlowMixer's relative error exhibits power-law scaling with grid resolution:
\begin{equation}
\varepsilon_{\text{rel}} \propto N^{-\alpha} \quad \text{where } \alpha \approx 0.25
\end{equation}
This sub-linear scaling indicates diminishing returns beyond $400\times160$ grid points for practical engineering applications.
\end{proposition}

\begin{remark}[Engineering Context]
The optimal operating point balances accuracy requirements with computational resources:
\begin{itemize}
\item \textbf{Engineering tolerance}: Industry standard requires $\leq 5\%$ relative error
\item \textbf{FlowMixer at $400\times160$}: Achieves 4.4\% error with $16.3\times$ speedup over spectral methods
\item \textbf{Traditional CFD baseline}: Delivers $<1\%$ error but requires $16\times$ more computation time
\end{itemize}
\end{remark}

%%%%%%%%%%%%%%%%%%%%%%%%%%%%%%%%%%%%%%%%%%%%%%%%%%%%%%%%%%%%%%%%%%%%%%%%%%%%%%%
%%%%%%%%%%%%%%%%%%%%%%%%%%%%%%%%%%%%%%%%%%%%%%%%%%%%%%%%%%%%%%%%%%%%%%%%%%%%%%%
% APPENDIX Comparison Vorticity FlowMixer vs ConvLSTM
%%%%%%%%%%%%%%%%%%%%%%%%%%%%%%%%%%%%%%%%%%%%%%%%%%%%%%%%%%%%%%%%%%%%%%%%%%%%%%%
%%%%%%%%%%%%%%%%%%%%%%%%%%%%%%%%%%%%%%%%%%%%%%%%%%%%%%%%%%%%%%%%%%%%%%%%%%%%%%%

\clearpage
\subsection{Comparison Vorticity FlowMixer vs ConvLSTM}
\label{app:convLSTM}
\begin{figure}[!h]
\centering
\includegraphics[width=1.0\textwidth]{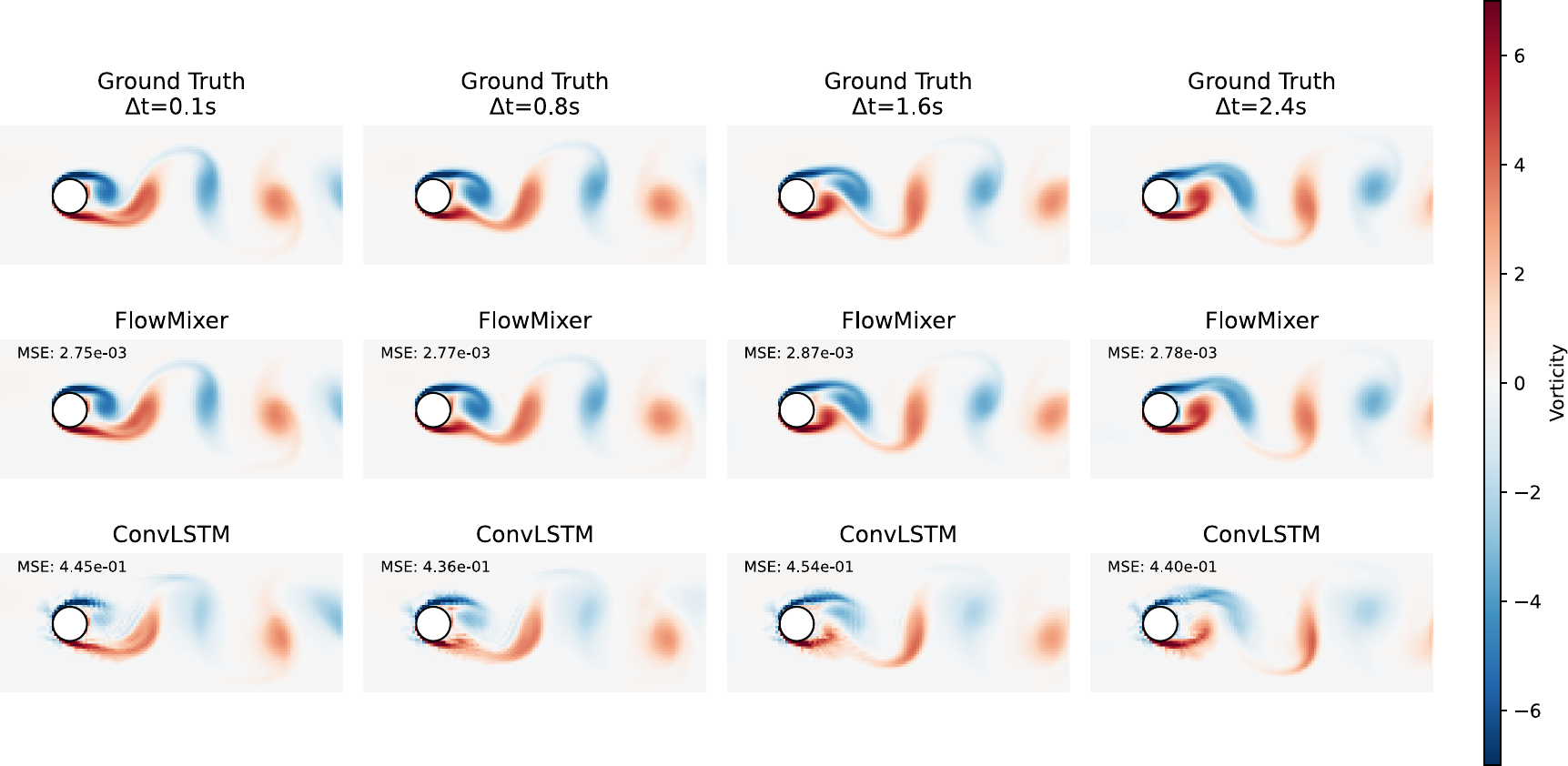}
\caption{Comparison of vorticity field predictions for 2D flow past a cylinder at Re=150. The figure shows predictions at four time points ($\Delta t$ = 0.1s, 0.8s, 1.6s, and 2.4s) for ground truth, FlowMixer, and ConvLSTM. FlowMixer demonstrates strong performance in capturing the complex vortex shedding patterns, maintaining accuracy over longer time horizons. The color scale represents vorticity magnitude, with red indicating positive (counterclockwise) and blue indicating negative (clockwise) vorticity. Mean Squared Error (MSE) values are provided for each prediction, highlighting FlowMixer's consistently lower error rates than ConvLSTM across all time steps.}\label{fig_flows}
\end{figure}

\newpage
\subsection{Impact of Optimizer on FlowMixer's Vorticity Predictions}
\label{app:SGDvsADAM}
\begin{figure*}[!h]
\centering
\includegraphics[width=1.0\textwidth]{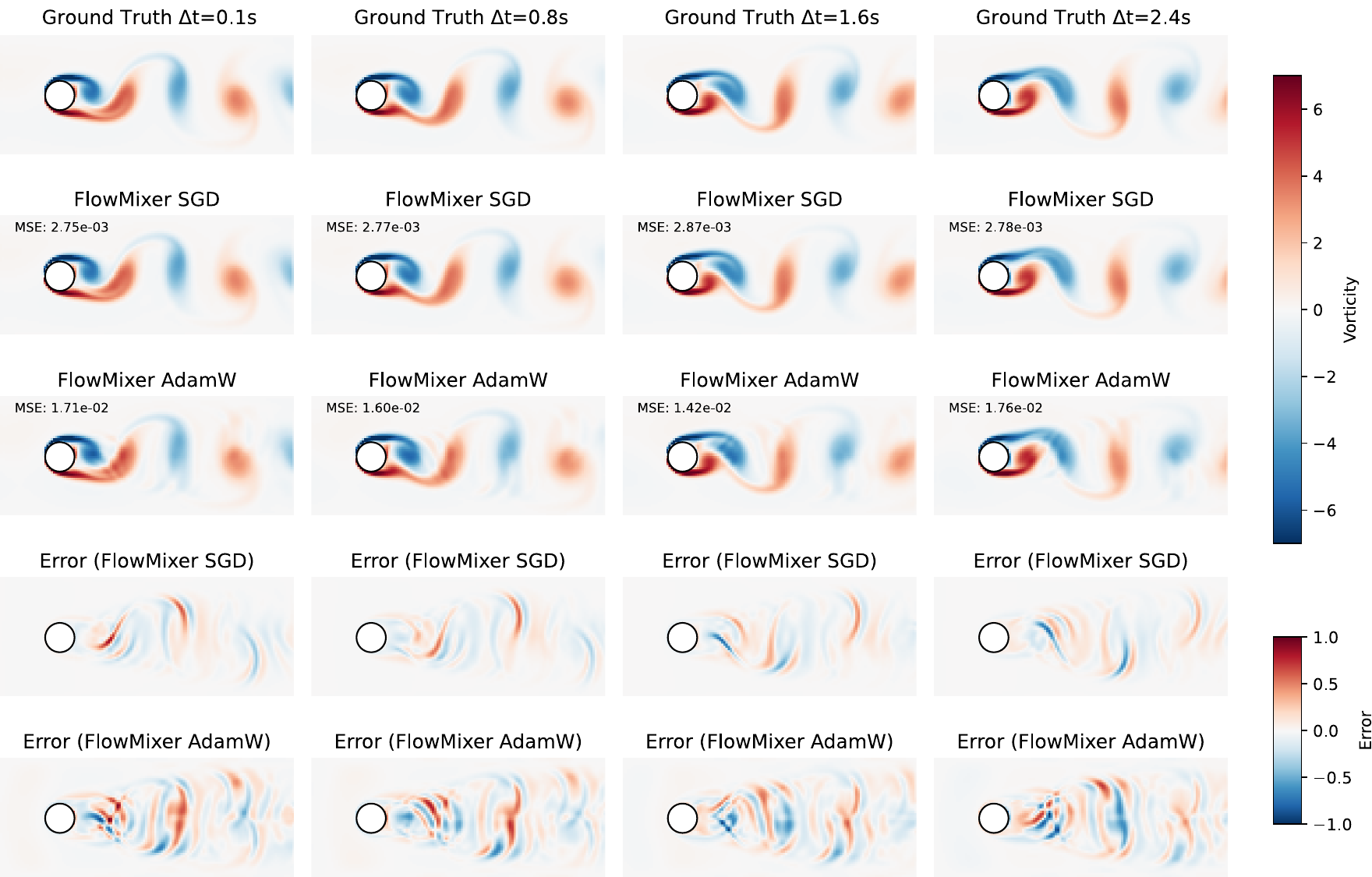}
\caption{Comparative analysis of vorticity field predictions for 2D flow past a cylinder at Reynolds number Re=150, demonstrating the impact of optimization algorithm choice. The visualization presents predictions at four temporal snapshots ($\Delta t$ = 0.1s, 0.8s, 1.6s, and 2.4s) comparing ground truth against FlowMixer models trained with SGD (momentum=0.9) and AdamW optimizers~\cite{kingma2014adam, loshchilov2017fixing}. The SGD-trained model demonstrates improved generalization performance, evidenced by both quantitative and qualitative metrics: (i) one order of magnitude reduction in Mean Squared Error (MSE) across all time steps, and (ii) visibly more accurate reproduction of vortex structures and flow patterns. These results align with theoretical observations in the machine learning literature regarding SGD's superior generalization capabilities compared to adaptive methods. The cylinder flow case provides a striking visualization of this phenomenon, as the complex spatiotemporal dynamics of vortex shedding serve as a sensitive indicator of model generalization quality. The color scale represents vorticity magnitude, with red and blue indicating positive (counterclockwise) and negative (clockwise) rotation, respectively.}
\end{figure*}

\end{document}